\documentclass[journal]{IEEEtran}
\usepackage{amsmath,times,algorithm,algorithmic,color,colortbl}
\usepackage{cite}
\usepackage{amssymb}
\usepackage{setspace}

\ifCLASSINFOpdf
\usepackage[pdftex]{graphicx}
\else
\usepackage{graphicx}
\fi

\begin{document}
%
\title{Multi-Target Tracking and Occlusion Handling with Learned Variational Bayesian Clusters and a Social Force Model}
\author{Ata-ur-Rehman,
        Syed~Mohsen~Naqvi,~\IEEEmembership{Senior~Member,~IEEE,}\\
        Lyudmila~Mihaylova,~\IEEEmembership{Senior~Member,~IEEE}~and~Jonathon~Chambers,~\IEEEmembership{Fellow,~IEEE}

\thanks{Ata-ur-Rehman and L. Mihaylova are with the Department of Automatic Control and Systems Engineering, University of Sheffield, United Kingdom, (email: a.ur-rehman@sheffield.ac.uk; L.S.Mihaylova@sheffield.ac.uk) }
\thanks{ S. M. Naqvi and J. A. Chambers are with the
School of Electrical and Electronic Engineering, Newcastle University, United Kingdom. (e-mail: Mohsen.Naqvi@newcastle.ac.uk; Jonathon.Chambers@newcastle.ac.uk)}}

\maketitle
\begin{abstract}
This paper considers the problem of multiple human target tracking
in a sequence of video data. A solution is proposed which is able to
deal with the challenges of a varying number of targets,
interactions and when every target gives rise to multiple
measurements. The developed novel algorithm comprises variational
Bayesian clustering combined with a social force model, integrated
within a particle filter with an enhanced prediction step. It
performs measurement-to-target association by automatically
detecting the measurement relevance. The performance of the
developed algorithm is evaluated over several sequences from
publicly available data sets: AV16.3, CAVIAR and PETS2006, which
demonstrates that the proposed algorithm successfully initializes
and tracks a variable number of targets in the presence of complex
occlusions. A comparison with state-of-the-art techniques due to
Khan et al., Laet et al. and Czyz et al. shows improved tracking
performance.
\end{abstract}


\IEEEpeerreviewmaketitle

\section{Introduction}
\label{intro}

\subsection{Motivation}
\IEEEPARstart{T}{his} paper presents a robust multiple tracking
algorithm which can be used to track a variable number of people
moving in a room or enclosed environment. Hence, the estimation of
the unknown number of targets and their states is considered in each
video frame. Multi-target tracking has many applications, such as
surveillance, intelligent transportation, behavior analysis and
human computer interfaces \cite{zia,B.Ristic,Mohsin-j}. It is a
challenging problem and a wealth of research has been undertaken to
provide more efficient solutions \cite{zia,tinne,nn1}.
 Two of the major challenges associated with visual tracking of multiple targets are: 1) mitigating occlusions and 2) handling a variable number of targets. Therefore, the focus of this paper is to present a technique to robustly handle these challenges.

Overcoming occlusions is a difficult task because during an
occlusion it is not possible to receive measurements originating
from the occluded objects. Significant research effort has been made
to address this problem, e.g. \cite{zia,tinne,Mul-cam2,3d-1}.

Many 2-D tracking systems rely on appearance models of people. For
instance~\cite{feature-2} applies the kernel density approach
while~\cite{Mila_m-cues} is based on color histograms, gradients and
texture models to track human objects. These techniques use template
matching which is not robust to occlusions because the occluded
parts of targets cannot be matched. They perform an exhaustive
search for the desired target in the whole video frame which
obviously requires much processing time. These techniques are also
 sensitive to illumination changes.

Occlusions can be handled with the help of efficient association of
available data to the targets. Most of the existing multiple target
tracking algorithms assume that the targets generate one measurement
at a time \cite{zia,single-meas1}. In the case of human tracking
from video and many other tracking applications, targets may
generate more than one measurement \cite{tinne,Mohsin-j2}. To
exploit multiple measurements, most algorithms add a preprocessing
step which involves extracting features from the measurements
\cite{Mila_m-cues}. This preprocessing step solves the problem to
some extent but it results in information loss due to the
dimensionality reduction and hence generally degrades the tracking
results. Another problem with many of these algorithms is that they
adopt the hard assignment technique \cite{Mila_m-cues,color} wherein
likelihood models are employed to calculate the probability of the
measurements.

Recently, in \cite{tinne} an approach relying on clustering and a
joint probabilistic data association filter (JPDAF) was proposed to
overcome occlusions. Rather than extracting features from the
measurements, this approach groups the measurements into clusters
and then assigns clusters to respective targets. This approach is
attractive in the sense that it prevents information loss but it
fails to provide a robust solution to mitigate the occlusion
problem. This is because it only utilizes the location information
of targets and hence the identity of individual targets is not
maintained over the tracking period. As a result, the tracker may
confuse two targets in close interaction, eventually causing a
tracker switching problem. Other advances of such filtering
approaches to multiple target tracking include multiple detection
JPDAF (MD-JPDAF) \cite{MD-JPDAF} and interacting multiple model
JPDAF (IMM-JPDAF) \cite{IMM-JPDAF}.

Interaction models have been proposed in the literature to mitigate the occlusion problem \cite{zia,Godsill-Group1}.
The interaction model presented in \cite{zia} exploits a Markov random field approach to penalize the predicted state of particles which may cause occlusions. This approach works well for tracking multiple targets where two or more of them do not occupy the same space, but it does not address tracking failures caused by inter-target occlusions during target crossovers.

\subsection{Dynamic Motion Models}

Dynamic models can be divided broadly into macroscopic and
microscopic models \cite{sfm-book}. Macroscopic models focus on the
dynamics of a collection of targets. Microscopic models deal with
the dynamics of individual targets by taking into account the
behavior of every single target and how they react to the movement
of other targets and static obstacles.

A representative of microscopic models is the social force
model~\cite{SFM}. The social force model can be used to predict the
motion of every individual target, i.e. the driving forces which
guide the target towards a certain goal and the repulsive forces
from other targets and static obstacles. A modified social force
model is used in~\cite{cavallaro} and in \cite{Pellegrini} for
modeling the movements of people in non-observed areas.

Many multi-target tracking algorithms in the literature only
consider the case when the number of targets is known and fixed
\cite{Mohsin-j,Godsill-Group1,color,Godsill}. In \cite{zia} a
reversible jump Markov chain Monte Carlo (RJMCMC) sampling technique
is described but in the experimental results a very strong
assumption is made that the targets (ants) are restricted to enter
or leave from a very small region (nest site) in a video frame.
Random finite set (RFS) theory techniques have been proposed for
tracking multiple targets, e.g.~\cite{Mahler,rfs}, in particular the target states 
 are considered as an RFS.

An appealing approach where the number of targets is estimated is
proposed in \cite{tinne} and is based on the evaluated number of
clusters. Non-rigid bodies such as humans can however produce multiple
clusters per target. Therefore, the number of clusters does not
always remain equal to the number of targets. Hence, calculating the
number of targets on the basis of the number of clusters can be
inaccurate in the case of human target tracking.

\subsection{Main Contributions}
The main contributions of this work are:
\begin{itemize}
\item[1.] An algorithm that provides a solution to multiple target tracking and complex inter-target interactions and
occlusions. Dealing with occlusions is based on social forces
between targets.
\item[2.] An improved data association technique which clusters the measurements and then uses their locations and features for accurate target  identification.
\item[3.] A new technique based on the estimated positions of targets, size and location of the clusters, geometry of the monitored area, along with a death and birth concept to handle robustly the variable number of targets.
\end{itemize}
The framework proposed in \cite{tinne} uses the JPDAF together with
the variational Bayesian clustering technique for data association
and a simple dynamic model for state transition. Our proposed
technique differs from that of \cite{tinne} in three main aspects:
1) in the dynamic model which is based on social forces between
targets, 2) a new features based data association technique, and 3)
a new technique for estimation of number of targets which is based
on the information from clusters, position of existing targets and
geometry of monitored area location.

The remainder of the paper is organized as follows: Section \ref{sec2} describes the proposed algorithm, Section \ref{sec2d} presents the proposed data association algorithm, Section \ref{clus} explains the clustering process and Section \ref{vnt} describes the proposed technique for estimating the number of targets.  Extensive experimental validation is given in Section \ref{sec3}. Section \ref{sec-dis} contains discussion and finally, conclusions are presented in Section \ref{sec4}.

\section{The Proposed Algorithm}
\label{sec2}
\subsection{Bayesian estimation }
\label{dmm} The goal of the multi-target tracking process is to
track the state of $N$ unknown targets. The state of each target is
represented as ${\bf x}^i_k, \, (i=1,\ldots, N)$, which is a column
vector containing the position and velocity information
of the $i^{th}$ target at time $k$. The joint state of all the
targets is constructed as the concatenation of the individual target
states $\mathbf{x}_k=[({\bf x}^1_k)^T, \ldots, ({\bf x}^i_k)^T,
\ldots, ({\bf x}^N_k)^T ]^T$, where $(\cdot)^T$ denotes the
transpose operator. Measurements at time $k$ are represented as
$\mathbf{ {y}}_k=[({\bf  {y}}^1_k)^T,\ldots,({\bf
  {y}}^j_k)^T,\ldots,({\bf   {y}}^L_k)^T ]^T$, where ${\bf
 {y}}^j_k$ is described in Section \ref{mes_mod}.

Within the Bayesian framework, the tracking problem consists of
estimating the belief of the state vector $\mathbf{x}_k$ given the
measurement vector $\mathbf{  {y}}_{1:k}$. The objective is to
sequentially estimate the posterior probability distribution
$p(\mathbf{x}_k|\mathbf{ y}_{1:k})$ at every time step
$k$. The posterior state distribution $p(\mathbf{x}_k|\mathbf{
y}_{1:k})$ can be estimated in two steps. First, the
prediction step~\cite{Arulam} is
\begin {equation}\label{eq3-2}
p( {\bf x}_{k}| {\bf   {y}}_{1:k-1})= \int p( {\bf x}_{k}| {\bf
x}_{k-1})p( {\bf x}_{k-1}| {\bf   {y}}_{1:k-1})d{\bf x}_{k-1},
\end{equation}
where $p( {\bf x}_{k}| {\bf x}_{k-1})$ is the state transition
probability. After the arrival of the latest measurements, the update step becomes
\begin{equation}\label{eq3-14}
p({\bf x}_{k}|{\bf   {y}}_{1:k})= \frac{p({\bf   {y}}_{k}|{\bf
x}_{k})p({\bf x}_{k}|{\bf   {y}}_{1:k-1})}{p({\bf   {y}}_{k}|{\bf
{y}}_{1:k-1})},
\end{equation}
where $p({\bf   {y}}_{k}|{\bf x}_{k})$ is the measurement likelihood function.

The two step tracking process yields a closed form expression only
for linear and Gaussian dynamic models \cite{single-meas1}. Particle
filtering \cite{Arulam} is a popular solution for suboptimal
estimation of the posterior distribution $p(\mathbf{x}_k|\mathbf{
{y}}_{1:k})$, especially in the nonlinear and
non-Gaussian case. The posterior state distribution is estimated by
a set of random samples ${\bf x}_{k}^{s}$, $s=1,\ldots,N_s$ and their associated weights ${w}^{s}_{k}$ at time $k$
\begin {equation}\label{eq52}
p({\bf x}_{k}|{\bf   {y}}_{1:k}) \approx
\sum_{s=1}^{N_{s}}{w}^{s}_{k}\delta ({\bf x}_{k}-{\bf x}_{k}^{s}),
\end{equation}
where $N_{s}$ is the total number of particles and
$\delta (\cdot)$ denotes a multivariate Dirac delta
function.

We apply particle filtering within the JPDAF framework to estimate
the states of targets. The JPDAF recursively updates the marginal
posterior distribution $p({\bf x}_{k}^i|{\bf
{y}}_{1:k})$ for each target. In our work,
given the state vector ${\bf x}^i_{k-1}$ of target $i$, the next set
of particles at time step $k$ is predicted by the social force
dynamic model (equations (\ref{fm2})-(\ref{fm5})) which is described later in Section \ref{fm}. The state transition model
is represented by equation (\ref{fm5}). The $N_s$ particles and
their corresponding weights can approximate the marginal posterior at time step $k$.

To assign weights to each sample we follow the likelihood model along with the data association framework which is explained in Section \ref{sec2d}.

\subsection{Measurement Model}
\label{mes_mod} The algorithm considers $L$ number of pixels in a
single video frame captured by one camera as input measurements.
From the sequentially arriving video frames, the silhouette of the
moving targets (foreground) is obtained by background subtraction~\cite{K.Kim}.
At time $k$, the measurement vector is: $\mathbf{y}_k=[({\bf
y}^1_k)^T,\ldots,({\bf y}^j_k)^T,\ldots,({\bf y}^L_k)^T ]^T$. The
state $\mathbf{x}_k$ at time $k$ is estimated by using the
measurement vector
 \begin{equation}
{\bf {y}}_k = h_k(\mathbf{x}_{k}, {\bf e}_{m,k}),
\end{equation}
where ${\bf e}_{m,k}$ is the measurement noise vector. After the
background subtraction~\cite{K.Kim}, the measurements at time $k$
are: $\tilde{\mathbf{y}}_k=[(\tilde{{\bf
y}}^1_k)^T,\ldots,(\tilde{{\bf y}}^j_k)^T,\ldots,(\tilde{{\bf
y}}^M_k)^T ]^T$,
which are assumed to correspond to the moving targets (please see
Section \ref{sec3} for details). The measurement
function $h_k(\cdot)$ is represented by some features of the video frames,
e.g. the color histogram. The measurement vector
$\tilde{\mathbf{y}}_k$ contains only a set of foreground pixels.
After background subtraction the number of pixels with non-zero
intensity values is reduced to $M$, where $M<<L$. The
number of foreground pixels $M$ is variable over time.

The background subtracted silhouette regions are clustered with the
variational Bayesian algorithm described in Section \ref{clus}.
During the clustering process each data point (pixel) ${\tilde{\bf
{y}}}^j_k$ contains only the coordinates of the pixel. During the
data association stage we use the red, green, blue (RGB) color
information contained in each pixel to extract color features of a
cluster.

\subsection{The Social Force Model}
\label{fm} The system dynamics model is defined by the following
equation
\begin{equation}
\mathbf{x}_k = f_k(\mathbf{x}_{k-1}, {\boldsymbol \xi}_{k-1}),
\end{equation}
where ${\boldsymbol \xi}_{k-1}$ is the system noise vector. Under
the Markovian assumption our state transition model for the $i^{th}$
target is represented as $p({\bf x}^i_k|{\bf
x}^i_{k-1},\mathcal{F}_i)$, where $\mathcal{F}_i$ is the social
force being applied on target $i$.

The motion of every target can be influenced by a number of factors,
e.g. by the presence of other targets and static obstacles. The
interactions between them are modeled here by means of a social
force model~\cite{SFM}, the key idea of which is to calculate
attraction and repulsion forces which represent the social behavior
of humans. The so-called ``social forces'' depend on the distances
between the targets.

Following Newtonian dynamics, a change of states stems from the
existence of exterior forces. Given a target $i$, the total number of
targets that influence target $i$ at time $k$, is $N_i$. The overall
force ${\bf \mathcal{F} }_i$ applied on target $i$ is the sum of the
forces exerted by all the neighboring targets, ${\bf \mathcal{F}}_i
= \sum_{j\in N_i} {\bf \mathcal{F}}_j$.
Most of the existing force based models \cite{SFM} consider only
repulsive forces between targets to avoid collisions. However, in
reality there can be many other types of social forces between
targets due to different social behavior of targets, for instance
attraction, spreading and following \cite{sfm-yan}.

Suppose a target $i$ at time step $k$ has $N_i$ neighbors,
therefore, it would have $N_i$ links with other targets $l = 1
\ldots N_i$. Only those targets are considered as neighbors of
target $i$ which are within a threshold distance $\hat{d}$ from $i$.
We consider that a social behavior over link $l$ is given by
$\varphi_l \in\{1 \ldots \Gamma\}$ where $\Gamma$ is the total number of
social behaviors. An interaction mode vector $\boldsymbol{\varphi}_i~=~[\varphi_1, \varphi_2, \ldots, \varphi_{N_i}]$ for target $i$ is a
vector comprising the different social behaviors over all the links.
The total number of interaction modes is $S = \Gamma^{N_i}$.

A force due to one interaction mode is calculated as a sum of forces over all the social links

\begin{equation}\label{fm2}
\mathcal{F}(\boldsymbol{\varphi}_i) = \sum_{l = 1}^{N_i} \phi^l,
\end{equation}
where $\mathcal{F}(\boldsymbol{\varphi}_i)$ is the force due to social mode $\boldsymbol{\varphi}_i$ and $\phi^l$ is the force over the link $l$.
%
%
In our work we consider three social forces: repulsion, attraction
and non-interaction. These broadly encompass all possible human
motions.

People naturally tend to avoid collisions between each other. This
is reflected by \emph{repulsion} forces. \emph{Attraction} accounts
for the behaviors when people approach each other for the intention
of meeting, this behavior is usually ignored in the existing force
based models. The \emph{non-interactions} mode represents the
behavior of independent motion of every person.

The repulsive force applied by target $j$ on target $i$ over the link
$l$ is calculated as \cite{sfm-yan}
\begin{equation}\label{fm4}
\phi_{-}^l= f_r \exp\biggr{(}\frac{r^l_{ij}-d^l_{ij}}{b}\biggr{)}{\bf u}_{ji},
\end{equation}
when  $d^l_{ij}$ is less than an empirically found threshold $\hat{d}$, defined as $3m$ in Table \ref{par} in Section \ref{trackn-res}; where $b$ which is set equal to $\hat{d}$ is the boundary in which a target $j$ has its influence of force on target $i$. The Euclidean distance between targets $i$ and $j$ over link $l$ is defined as $d^l_{ij}$, $r^l_{ij} = r^l_i + r^l_j$
is the sum of radii of influence of targets $i$ and $j$ over link $l$ and $f_r$ is the magnitude of the repulsive force. The unit
vector from target $j$ to target $i$ is represented as ${\bf
u}_{ji}$.

When targets approach each other, repulsive forces act, to
avoid collisions. The repulsive force therefore increases as described
by equation (\ref{fm4}). Similarly, when targets move
apart the repulsion decreases because it is less likely for a collision
to occur, and therefore they tend to come closer to each other.
Attraction is a phenomenon opposite to repulsion. When targets move
apart, they tend to attract each other for the intension of meeting
and when they move closer, the repulsion increases which means they
are less attracted.  This attraction phenomenon is described by
equation (\ref{fm3}). However, if targets are at a distance more
than a threshold, it can be assumed they do not influence each other.

An attractive force applied by target $j$ on target $i$ over the
link $l$ is calculated as \cite{sfm-yan}
\begin{equation}\label{fm3}
\phi_{+}^l = f_a exp\biggr{(}-\frac{(r_{ij}^l-d_{ij}^l)}{b}\biggr{)}{\bf u}_{ij},
\end{equation}
when $d_{ij}^l$ is less than an empirically found threshold $\hat{d}$, defined as $3m$ in Table \ref{par} in Section \ref{trackn-res}, where $f_a$ is the magnitude of the attractive force and force due to
non-interactions is considered to be zero.


To represent the state ${\bf x}_{k}$ of targets at every time step $k$, a pixel coordinate
system is used. Since, our state transition equation (\ref{fm5}) below is based on
the social force model, to predict the new
particles ${\bf x}_{k}^{s}$, position and velocity values contained within the state vector ${\bf x}_{k-1}$ and
the particles at time step $k-1$ are converted to the image ground coordinate
system (a process known as homography \cite{Hartley}). After predicting the new set
of $N_s$ particles for state ${\bf x}_{k}$ these particles are
converted back to pixel coordinates. For conversion between the two
coordinate systems the four point homography technique is
implemented, as used in~\cite{jpdaf-alternate}. The position and velocity at time step $k$ in pixel coordinates are represented as ${\bf p}_k$ and ${\bf v}_k$, respectively, whereas, the position and velocity in ground coordinates systems are represented as $\tilde{\bf p}_k$ and $\tilde{\bf v}_k$ respectively.
At every time step a new state is predicted with respect to interaction mode $\boldsymbol{\varphi}_i$ according to the following model \cite{sfm-yan}
\begin{equation}\label{fm5}
\biggr{[}\begin{matrix} \tilde{\bf p}_k(\boldsymbol{\varphi}_i)\\ \tilde{\bf v}_k(\boldsymbol{\varphi}_i) \end{matrix}\biggr{]} = \biggr{[}\begin{matrix} \tilde{\bf p}_{k-1}(\boldsymbol{\varphi}_i)+\tilde{\bf v}_{k-1}\Delta t+ \frac{1}{2}\frac {{\bf \mathcal{F}}(\boldsymbol{\varphi}_i)}{m}\Delta t^2\\
\tilde{\bf v}_{k-1}(\boldsymbol{\varphi}_i) + \frac {\mathcal{F}(\boldsymbol{\varphi}_i)}{m}\Delta t \end{matrix}\biggr{]} + {\boldsymbol \xi}_{k},
\end{equation}
where $\Delta t$ is the time interval between two frames, $m$ represents mass of the target and ${\boldsymbol \xi}_{k}$ is the system noise vector.

The social force model gives more accurate predictions
than the commonly used constant velocity model~\cite{FortmanandBarshalom} and helps tracking
objects in the presence of close interaction. This cannot be done
with a simple constant velocity motion
model.

The implementation of the force model is further explained in {\bf
Algorithm~1} at the end of the next section.

\section{Data Association and Likelihood}
\label{sec2d} To deal with the data association problem we represent
a measurement to target $(M \rightarrow T)$ association hypothesis
by the parameter $\boldsymbol{\psi}_k$, whereas
$\boldsymbol{\Psi}_{k}$ represents the set of all hypotheses. At
this stage we assume that one target can generate one measurement.
However, later in this section we relax this assumption and develop
a data association technique for associating multiple measurements
(clusters of measurements) to one target. If we assume that
measurements are independent of each other then the likelihood model
conditioned on the state ${\bf x}_{k}$ of targets and measurement to
target association hypothesis $\boldsymbol{\psi}_k$ can be
represented as in~\cite{single-meas1} (equation (11))
\begin{equation}\label{eq38}
\hspace{-0.3cm} p({\bf {y}}_{k}|{\bf x}_k,\boldsymbol{\psi}_k) =
\prod_{{j'}\in\mathcal{\boldsymbol{J}}_o} p_c({\bf
{y}}_k^{j'})\prod_{j\in\mathcal{\boldsymbol{J}}}p_t({\bf {y}}_k^j|{\bf
x}_k^{i,j}),
\end{equation}
where $\mathcal{\boldsymbol{J}}_o$ and $\mathcal{\boldsymbol{J}}$
are, respectively, measurement indices sets, corresponding to the
clutter measurements and measurements from the targets to be
tracked, for convenance of notation the time dependence is not shown on these quantities.
The terms $p_c(\cdot)$ and $p_t(\cdot)$ are characterizing,
respectively, the clutter likelihood and the measurement likelihood
association to each target $i$ whose state is represented as ${\bf
x}_k^{i,j}$. If we assume that the clutter likelihood $p_c(\cdot)$
is uniform over the volume $\gamma$ of the measurement space and
${f_k}$ represents the total number of clutter measurements, then
the probability $p({\bf {y}}_{k}|{\bf x}_k,\boldsymbol{\psi}_k)$ can
be represented as

\begin{equation}\label{eq38-b}
\hspace{-0.3cm} p({\bf {y}}_{k}|{\bf x}_k,\boldsymbol{\psi}_k) =
\gamma^{f_k}\prod_{j\in\mathcal{\boldsymbol{J}}}p_t({\bf
{y}}_k^j|{\bf x}_k^{i,j}).
\end{equation}
The form of $p_t({\bf {y}}_k^j|{\bf x}_k^{i,j})$ is given in
Section~\ref{ztot}.

\subsection{The JPDAF Framework}
The standard JPDAF algorithm estimates the marginal distribution of
each target by following the prediction and update process described
in equations (\ref{eq3-2}) and (\ref{eq3-14}). In the development below,
we exploit the framework from~\cite{single-meas1}. The prediction step for
each target independently becomes
\begin {equation}\label{eq3-j1}
p( {\bf x}^{i}_{k}| {\bf  {y}}_{1:k-1})= \int p( {\bf x}^i_{k}| {\bf
x}^i_{k-1})p( {\bf x}^i_{k-1}| {\bf  {y}}_{1:k-1})d{\bf x}^i_{k-1},
\end{equation}
where we assume that the prediction at time step $k$ is based on the
measurements only at $k-1$. The JPDAF defines the likelihood of
measurements with respect to target $i$ as in~\cite{single-meas1} (equation (25))
\begin{equation}\label{eq54}
 p({\bf y}_{k}|{\bf x}_k^{i}) = {\mathcal{A}}_{k}^{i,o} + \sum_{j=1}^{L}{\mathcal{A}}_k^{i,j}p({\bf  y}^j_{k}|{\bf x}_k^{i}),
\end{equation}
where ${\mathcal{A}}_k^{i,j}$ is the association probability that the
$i^{th}$ target is associated with the $j^{th}$ measurement and
${\mathcal{A}}_{k}^{i,o}$ is the association probability that the
target $i$ remains undetected; for notational convenience, in equation (\ref{eq54}) and the remainder of the paper we drop the subscript $t$ on the target likelihood.
With prediction and likelihood
functions defined by equations (\ref{eq3-j1}) and (\ref{eq54}),
respectively, the JPDAF estimates the posterior probability of the
state of target $i$
\begin{equation}\label{eq3-1j} p({\bf
x}^i_k|{\bf { y}}_{1:k}) \propto p({\bf { y}}_k|{\bf x}^i_k)p({\bf
x}^i_k|{\bf { y}}_{1:k-1}).
\end{equation}
The association probability ${\mathcal{A}}_k^{i,j}$ can be defined
as in~\cite{single-meas1} (equation (27))
\begin{equation}\label{eq36}
{\mathcal{A}}_k^{i,j} =
\sum_{\boldsymbol{\psi}_k\in\boldsymbol{\Psi}_k^{i,j}}p(\boldsymbol{\psi}_k|{\bf
{y}}_{1:k}),
\end{equation}
where $\boldsymbol{\Psi}_k^{i,j}$ represents the set of all those hypotheses which associate the $j^{th}$ measurement to the $i^{th}$ target. The probability $p(\boldsymbol{\psi}_k|{\bf {y}}_{1:k})$ can be
represented as equation (28) in \cite{single-meas1}
\begin{equation}\label{eq37}
p(\boldsymbol{\psi}_k|{\bf {y}}_{1:k}) \propto
p(\boldsymbol{\psi}_k|{\bf x}_k)\gamma^{f_k}
\prod_{j\in\mathcal{\boldsymbol{J}}} p({\bf {y}}^j_{k}|{\bf
{y}}_{1:k-1}),
\end{equation}
where $p(\boldsymbol{\psi}_k|{\bf x}_k)$ is the
probability of assignment $\boldsymbol{\psi}_k$ given the current
state of the objects. We assume that all assignments have equal
prior probability and hence $p(\boldsymbol{\psi}_k|{\bf x}_k)$ can
be approximated by a constant. If we define a normalizing constant $\pi$ ensuring that $p(\boldsymbol{\psi}_k|{\bf {y}}_{1:k})$ integrates to one
and define the predictive likelihood $p({\bf {y}}^j_{k}|{\bf {y}}_{1:k-1})$ as 
\begin{equation}\label{eq38}
\begin{split}
p({\bf {y}}^j_{k}|{\bf {y}}_{1:k-1}) = \int p({\bf {y}}_k^j|{\bf
x}_k^i) p({\bf x}_k^i|{\bf {y}}_{1:k-1})d{\bf x}_k^i,
 \end{split}
 \end{equation}
then equation (\ref{eq37}) becomes
\begin{equation}\label{eq37-b}
\hspace{-0pt} p(\boldsymbol{\psi}_k|{\bf {y}}_{1:k}) \!
= \! \pi \gamma^{f_k} \! \! \!\prod_{(j,i) \in\boldsymbol{\psi}_k}
\! \int \! \! p({\bf {y}}_k^j|{\bf x}_k^i) p({\bf x}_k^i|{\bf
{y}}_{1:k-1})d{\bf x}_k^i.
\end{equation}
\subsection{Particle Filtering Based Approach}
The standard JPDAF assumes a Gaussian marginal filtering
distribution of the individual targets
\cite{FortmanandBarshalom,single-meas1}. In this paper we adopt a
particle filtering based approach which represents the state of
every target with the help of samples. Therefore, we modify equation
(\ref{eq37-b}) as 
\begin{equation}\label{eq55-a}
 p(\boldsymbol{\psi}_k|{\bf {y}}_{1:k}) = \pi\gamma^{f_k}\prod_{(j,i)\in\boldsymbol{\psi}_k}\sum_{s=1}^{N_s} \hat{w}^{i,s}_{k}p({\bf  {y}}_k^j|{\bf x}_k^{i,s}),
 \end{equation}
where $N_s$ is the total number of particles which are required to estimate the filtering distribution over every individual target and
$\hat{w}^{i,s}_{k}$ represents the associated predictive weights as described in \cite{single-meas1}.
We can modify equation (\ref{eq55-a}) as 
\begin{equation}\label{eq55-b}
p(\boldsymbol{\psi}_k|{\bf {y}}_{1:k}) = \pi
\gamma^{f_k}\prod_{(j,i)\in\boldsymbol{\psi}_k}\frac{1}{N_s}\sum_{s=1}^{N_s}
p({\bf  {y}}_k^j|{\bf x}_k^{i,s}).
 \end{equation}
Once the probability $p(\boldsymbol{\psi}_k|{\bf {y}}_{1:k})$ is
calculated according to (\ref{eq55-b}), we can substitute it in
equation (\ref{eq36}) to calculate the association probability. We
can further use this association probability to calculate the
measurement likelihood from equation (\ref{eq54}).

After drawing particles from a suitably constructed proposal
distribution~\cite{Arulam}, the weights associated with particles
are calculated, $\hat{w}^{i,s}_{k} = p({\bf  y}_{k}|{\bf x}_k^{i})$.
The particles and their weights can then be used to approximate the
posterior distribution
\begin {equation}\label{eq52-pf}
p({\bf x}^i_{k}|{\bf  {y}}_{1:k}) \approx
\sum_{s=1}^{N_{s}}{w}^{i,s}_{k}\delta ({\bf x}^i_{k}-{\bf
x}_{k}^{i,s}),
\end{equation}
for individual targets where ${\bf x}_{k}^{i,s}$ is the $s^{th}$
particle for target $i$ and ${w}^{i,s}_{k}$ is the associated
weight. For estimation of a state we predict $N_{\varphi}$ particles with respect to every interaction mode. There are $S$ interaction modes, therefore, $N_s = S \times N_{\varphi}$ represents the total number of samples.

\subsection{Algorithm to Associate Multiple Measurements to a Target }
 \label{ztot}
In the case of tracking people in
video, every person generates multiple measurements. To avoid any
information loss we relax the assumption that every person can
generate a single measurement. Here we present a data association
technique for associating multiple measurements to every target. To
achieve this, the algorithm groups the foreground pixels into
clusters with the help of a variational Bayesian clustering
technique, and then instead of associating one measurement to one
target we associate clusters to every target. The clustering process
returns $\kappa$ clusters,  where the number of clusters is not
predefined or fixed. Clusters at discrete time $k$ are regions
represented as $  {\bf Z}^1_k, {\bf Z}^q_k,\ldots,{\bf Z}^\kappa_k
$, where ${\bf Z}^q_k$ is the $q^{th}$ cluster which contains
certain measurements, i.e. a number of vectors $ {\tilde{\bf
y}}_k^j$ of foreground pixels. The complete clustering process is
described in Section \ref{clus}. The clustering process aims to
assign measurements to the respective clusters and gives as output a
correspondence matrix ${\bf B}_k = [({\bf b}_k^1)^T,\ldots,({\bf
b}_k^j)^T,\ldots,({\bf b}_k^M)^T]^T$, where ${\bf b}_ k^j =
[b_k^{j,1}, b_k^{j,2},\ldots,b_k^{j,q},\ldots,b_k^{j,\kappa}]$
indicates, at discrete time $k$, to which cluster, the measurement
vector ${\tilde{\bf y}}_k^j$ corresponds. All but one of the
elements of ${\bf b}_ k^j$ is zero, if e.g. ${\tilde{\bf y}}_k^j$
belongs to the $q^{th}$ cluster then the correspondence vector will
be, ${\bf b}_ k^j = [0,\ldots,0,1,0,\ldots,0]$, which shows only the
$q^{th}$ element of ${\bf b}_ k^j$ is nonzero. Please note that from
this section onwards all equations are written with respect to the
measurement vector $\tilde {\bf y}_k$ that contains foreground
pixels.

We modify the measurement to target $(M \rightarrow T)$ association probability ${\mathcal{A}}_k^{i,j}$ to calculate the cluster to target $(Z\rightarrow T)$ association probability ${\mathcal{A}}_k^{i,q}$, where $q$ represents the cluster index
\begin{equation}\label{eq36-c}
{\mathcal{A}}_k^{i,q} =
\sum_{\boldsymbol{\psi}_k\in\boldsymbol{\Psi}_k^{i,q}}p(\boldsymbol{\psi}_k|
\tilde {
 \bf {y}}_{1:k}),
\end{equation}
where $\boldsymbol{\Psi}_k^{i,q}$ represents the set of all those hypotheses which associate the $q^{th}$ cluster to the $i^{th}$ target. We modify equation (\ref{eq55-b}) to define the probability
$p(\boldsymbol{\psi}_k| \tilde {\bf {y}}_{1:k})$ as
\begin{equation}\label{eq55-c}
p(\boldsymbol{\psi}_k|\tilde {\bf {y}}_{1:k}) = \pi
\gamma^{f_k}\!\!\prod_{(q,i)\in\boldsymbol{\psi}_k} \, \,\prod_{{\tilde
{\bf y}}_k^j:b_k^{j,q}=1} \!\!\frac{1}{N_s}\!\!\sum_{s=1}^{ N_s} p( \tilde
{\bf {y}}_k^j|{\bf x}_k^{i,s}),
 \end{equation}
where ${\tilde {\bf y}}_k^j:b_k^{j,q}=1$ represents only those
measurements which are associated with cluster $q$. The measurement
likelihood defined in equation (\ref{eq54}) is modified as follows
\begin{equation}\label{eq54-c}
\begin{split}
p(\tilde { \bf y}_{k}|{\bf x}_k^{i,s}) = {\mathcal{A}}_{k}^{i,o} +
\sum_{j=1}^{M}{\mathcal{A}}_k^{i,(q:b_k^{j,q}=1)}p(\tilde  {\bf y}^j_{k}|{\bf x}_k^{i,s}),
\end{split}
\end{equation}
where $q:b_k^{j,q}=1$ represents that the
$j^{th}$ measurement is associated with the $q^{th}$ cluster and
${\mathcal{A}}_{k}^{i,o}$ represents the probability that  the
cluster is not associated to the $i^{th}$ target and
${\mathcal{A}}_k^{i,(q:b_k^{j,q}=1)}$ is the
probability that the cluster is associated to the $i^{th}$ target.
To limit the number of hypotheses when the number of targets
increases, we have adopted Murty's algorithm~\cite{murty's}.
This algorithm returns the $k$-best hypotheses. The elements of the
cost matrix for Murty's algorithm are calculated with the
particles ${\bf x}_k^{i,s}$
%
\begin{equation}\label{eq-cost}
cost_q^i = \prod_{{\tilde {\bf y}}_k^j:b_k^{j,q}=1}
\frac{1}{N_s}\sum_{s=1}^{N_s} p(\tilde {\bf {y}}_k^j|{\bf
x}_k^{i,s}),
\end{equation}
where $cost_q^i$ represents the cost of assigning the $q^{th}$ cluster to the $i^{th}$ target.

The JPDAF \cite{jpdaf} data association technique proposed in
\cite{tinne} performs well when two targets do not undergo a full
occlusion. However, this data association technique sometimes fails
to associate measurements correctly, especially when two or more
targets partially occlude each other or when targets come out of
full occlusion. This is due to the fact that this data association
is performed on the basis of location information without any
information about the features of targets. This can result in missed
detections and wrong identifications. 

In order to cope with these challenges under occlusions, a data
association technique is proposed which assigns clusters to targets
by exploiting both location and features of clusters. First, we
define the term $\prod_{{\tilde {\bf y}}_k^j:b_k^{j,q}=1} \frac{1}{
N_s}\sum_{s=1}^{N_s} p(\tilde {\bf {y}}_k^j|{\bf x}_k^{i,s})$ in
equation ($\ref{eq55-c}$) as follows
\begin{equation}\label{eq61-z}
\begin{split}
\!\!\!\prod_{{\tilde {\bf y}}_k^j:b_k^{j,q}=1}\!\!\! \frac{1}{ N_s}\!\!\sum_{s=1}^{N_s}\! p(\tilde {\bf  {y}}_k^j|{\bf x}_k^{i,s}) \!=
\!\!{p({\bf Z}_k^q|{\bf x}_k^{i,s})}\!\!\!\!\!\!\!\prod_{{\tilde {\bf
y}}_k^j:b_k^{j,q}=1}\!\!\!\!\frac{1}{N_s}\sum_{s=1}^{N_s}
\!\!\mathcal{N}(\tilde {\bf {y}}_k^j|\boldsymbol{\mu}_s,{\bf \!\Sigma}_s\!),
\end{split}
\end{equation}
where $\boldsymbol{\mu}_s = {\bf x}_k^{i,s}$ is the mean vector and
${\bf \Sigma}_s$ is the fixed and known diagonal covariance matrix.
The association probability ${p({\bf Z}_k^q|{\bf x}_k^{i,s})}$ in
equation (\ref{eq61-z}) is calculated by extracting features from
the clusters at time step $k$ and comparing them with a reference feature (template) of target $i$.
The reference feature for target $i$ is extracted when it first appears in the monitored area.
To improve the computational
efficiency, the probability $p({\bf Z}_k^q|{\bf x}_k^{i,s})$ is
calculated only when targets approach each other. This association
probability is defined as
\begin{equation}\label{eq44}
 p({\bf Z}_k^q|{\bf x}_k^{i,s})\!=\!\! \left\{\begin{array}{ll}
          \!\!\!\!\exp \biggr{(}\!\!-\frac{d^i(H^i_{ref},H^q_{k})}{2{\bf \sigma}^2}\biggr{)} & \!\!\!\textrm{if} ~P_{occlusion}^i\!>\!\vartheta\\
          \!\!\!\!1 & \!\!\!\!\textrm{otherwise}
          \end{array}\right.
\end{equation}
where ${\bf \sigma}^2$ is the measurement noise variance,
$H^i_{ref}$ and $H^q_{k}$ are histograms of features of the $i^{th}$ target and
$q^{th}$ cluster, respectively and $d^i(H^i_{ref},H^q_{k})$ is the
distance between them. One possible option is to use the
Bhattacharyya distance \cite{Bhatta}. The parameter $\vartheta$ is a
predefined threshold, and the probability of occlusion is defined as
\begin{equation}\label{eq45}
P_{occlusion}^i = \exp \biggr(-\frac{d^i_{min}}{\Delta_c}\biggr),
\end{equation}
where $d^i_{min}$ is the minimum Euclidean distance among a set of distances which is calculated for target $i$ from all other targets at time $k$,  
\begin{equation}\label{eq46}
d^i_{min} = \min\limits_{i\neq n}
 {\{d_{i,n}\}},
\end{equation}
and $\Delta_c$ is the distance constant which limits the influence of target $n$ on target $i$. A higher value of minimum distance $d^i_{min}$ results in a smaller value of probability of occlusions and when this probability drops below threshold $\vartheta$ the probability $ p({\bf Z}_k^q|{\bf x}_k^{i,s})$ becomes unity.

\begin{algorithm}\label{algo-fm}
\caption{The social force model}
Input: $ N_s$ particles and the state of targets at time step $k-1$\\
Output: $ N_s$ particles for time step $k$\\
\begin{algorithmic}[1]
\STATE Convert the state at $k-1$ from pixel coordinates system to real world ground coordinates.
\STATE Use ground coordinates to calculate distances between targets.
\STATE Based on distances create links between targets.
\FOR{$i = 1,...,N$(where $N$ is the number of targets)} 
\STATE Create $S = \Gamma^{N_i}$ social modes
\STATE Calculate forces due to every social mode by using equations (\ref{fm2}), (\ref{fm4}) and (\ref{fm3}).
\FOR{$s = 1:\Gamma^{N_i}: N_s$(where $N_s$ is the number of particles)}
\STATE Convert the particle ${\bf x}_{k-1}^{i,s}$ to the ground plane.
\STATE Add random noise to get ${\bf x}_{k-1}^{i,s}+{\boldsymbol \xi}_{k-1}$ .
\FOR{$s_{\boldsymbol{\varphi}_i} = 1,...,\Gamma^{N_i}$}
\STATE Predict particle ${\bf x}_{k-1}^{i,s+s_{\boldsymbol{\varphi}_i}-1}$ by using particle ${\bf x}_{k-1}^{i,s}+{\boldsymbol \xi}_{k-1}$ w.r.t. the social mode $\boldsymbol{\varphi}_i$ in equation~(\ref{fm5}).
\STATE Convert the particle ${\bf x}_{k-1}^{i,s+s_{\boldsymbol{\varphi}_i}-1}$ to the pixel coordinates.
\ENDFOR
\ENDFOR
\ENDFOR
\end{algorithmic}
\end{algorithm}
\section{Variational Bayesian Clustering - from Measurements to Clusters}
\label{clus}
The clustering process aims to subdivide the foreground image into regions
corresponding to the different targets. A variational Bayesian
clustering technique is developed to group the $M$ foreground pixels
into $\kappa$ clusters. Each cluster at time index $k$ is represented by its center
${\bf {\boldsymbol{\mu}}}_k^q$, where $q = 1,\ldots,\kappa$. A binary
indicator variable $b_k^{j,q}\in\{0,1\}$ represents to which of the $q^{th}$ clusters data point
$ \tilde {\bf {y}}_k^j$  is assigned; for example, when the data point $\tilde  {\bf
 {y}}_k^j$ is assigned to the $q^{th}$ cluster then $b_k^{j,q} =
1$, and $b_k^{j,l} = 0$ for $l\neq q$.

We assume that the foreground pixel locations are modeled by a
mixture of Gaussian distributions. The clustering task can be viewed
as fitting mixtures of Gaussian distributions~\cite{bishop} to the
foreground measurements.
 Every cluster ${\bf Z}_k^q$ of foreground pixel locations is assumed to be
modeled by a Gaussian distribution, $\mathcal{N}(\tilde {\bf
{y}}_k^j|{\boldsymbol{\mu}}_k^q, {\bf \Sigma}_k^q)$ which is one
component of the mixture of Gaussian distributions. Each cluster has
a mean vector ${\boldsymbol{\mu}}_k^q$ and a covariance matrix
${\boldsymbol{\Sigma}}_k^q$. Hence, the probability of a data point
$\tilde {\bf {y}}^j_k$ can be represented as
\begin{equation}\label{eq6}
p(\tilde {\bf {y}}_k^j|{\bf C}_k,{\boldsymbol{\mu}}_k,{\bf \Sigma}_k) =  \sum_{q = 1}^{\kappa}
C_k^{q}\mathcal{N}(\tilde {\bf
 {y}}_k^j|{\boldsymbol{\mu}}_k^q,{\bf \Sigma}_k^q),
\end{equation}
where $\boldsymbol{\mu}_k = \{\boldsymbol{\mu}_k^q\}_{q=1}^\kappa $  and $\boldsymbol{\Sigma}_k = \{\boldsymbol{\Sigma}_k^q\}_{q=1}^\kappa $.
The mixing coefficient vector is defined as ${\bf C}_k=[C_k^{1},
\ldots,C_k^{q},\ldots,C_k^{\kappa}]^T$, where $C_k^{q}$ represents
the probability of selecting the $q^{th}$ component of the mixture
which is the probability of assigning the $j^{th}$ measurement to
the $q^{th}$ cluster.
If we assume that all the measurements are independent, then the log likelihood function becomes \cite{bishop}
 \begin{equation}\label{eq12n}
\ln p(\tilde {\bf {y}}_k|{\bf C}_k,{\boldsymbol{\mu}}_k,{\bf \Sigma}_k)
\!\!=\!\! \sum_{j=1}^{M}\!\ln\biggr{\{}\!\!\sum_{q = 1}^{\kappa}{
C}_k^{q}\mathcal{N}(\tilde {\bf {y}}_k^j|{\boldsymbol{\mu}}_k^q,{\bf
\Sigma}_k^q)\biggr{\}}.
\end{equation}
The data point $\tilde {\bf {y}}^j_k$ belongs to only one cluster at
a time and this association is characterized by a correspondence
variable vector ${\bf b}_ k^j$. Therefore, it can be represented as
$C_k^{q} = p(b_k^{j,q}=1)$
where the mixing coefficients must satisfy the following conditions:
$0<~C_k^{q}~\leq1$ and $\sum_{q = 1}^{\kappa} C_k^{q}=1$. Since
only one element of ${\bf b}_ k^j$ is equal to $1$, the probability
of the mixing coefficient can also be written as
\begin{equation}\label{eq10}
p({\bf b}_ k^j|{\bf C}_k) = \prod_{q=1}^{\kappa}{(C_k^{q})}^{b_k^{j,q}}.
\end{equation}
Similarly, we assume that $p(\tilde {\bf {y}}_k^j|{ b}_k^{j,q}=1)
= \mathcal{N}{(\tilde {\bf {y}}_k^j|{\boldsymbol{\mu}}_k^q,{\bf
\Sigma}_k^q)}$.
Since only one element of ${\bf b}_ k^j$ is equal to $1$, we can
write again
\begin{equation}\label{eq12}
p(\tilde {\bf {y}}_k^j|{\bf b}_k^j,\boldsymbol{\mu}_k,{\bf \Sigma}_k) =
\prod_{q=1}^{\kappa}\mathcal{N}{(\tilde {\bf
{y}}_k^j|{\boldsymbol{\mu}}_k^q,{\bf \Sigma}_k^q)}^{b_k^{j,q}}.
\end{equation}\\
As a result of the clustering process we will obtain estimates of
the probabilistic weight vector ${\bf C}_k$,
the mean vectors ${\boldsymbol{\mu}}_k$, the covariance matrices ${\bf \Sigma}_k$
and correspondence matrix ${\bf B}_k$.

The estimation problem can be simplified by introducing hidden
(latent) variables. We consider the correspondence variables ${\bf
B}_k$ as latent variables and suppose that they are known. Then $\{
\tilde {\bf {y}}_k,{\bf B}_k\}$ represents the complete dataset. The
log likelihood function for this complete data set becomes $\ln
p(\tilde {\bf {y}}_k,{\bf B}_k|{\bf C}_k,{\boldsymbol{\mu}}_k,{\bf
\Sigma}_k)$. Now the new set of unknown parameters contains ${\bf
B}_k, {\bf C}_k,{\boldsymbol{\mu}}_k$ and ${\bf \Sigma}_k$ which can be
estimated by the proposed adaptive variational Bayesian clustering
algorithm. If, for simplicity, these parameters are represented as a
single parameter $\boldsymbol{\Theta}_k = ({\bf B}_k, {\bf
C}_k,{\boldsymbol{\mu}}_k,{\bf \Sigma}_k)$, then the desired
distribution can be represented as $p(\boldsymbol{\Theta}_k|\tilde
{\bf {y}}_k)$. The log marginal distribution $p(\tilde {\bf {y}}_k)$
can be decomposed as described in~\cite{bishop} (Section 9.4 and
equations (10.2)-(10.4))
\begin{equation}\label{eq16}
\ln p(\tilde {\bf  {y}}_k) = L(q) + KL(q\|p).
\end{equation}
We define the distribution $\mathcal{Q}(\boldsymbol{\Theta}_k)$ as an
approximation of the desired distribution. Then the objective of the
variational Bayesian method is to optimize this distribution, by
minimizing the Kullback Leibler (KL) divergence
\begin{equation}\label{eq15}
KL(q\|p) = -\int \mathcal{Q}(\boldsymbol{\Theta}_k)\ln\biggr{\{}\frac
{p(\boldsymbol{\Theta}_k|\tilde {\bf
{y}}_k)}{\mathcal{Q}(\boldsymbol{\Theta}_k)}\biggr{\}}d\boldsymbol{\Theta}_k.
\end{equation}
The symbol $\cdot\|\cdot$ represents the KL divergence, and
\begin{equation}\label{eq17}
L(q) = \int \mathcal{Q}(\boldsymbol{\Theta}_k)\ln\biggr{\{}\frac
{p(\boldsymbol{\Theta}_k,\tilde {\bf
{y}}_k)}{\mathcal{Q}(\boldsymbol{\Theta}_k)}\biggr{\}}d\boldsymbol{\Theta}_k.
\end{equation}
Minimizing the KL divergence is equivalent to maximizing the lower
bound $L(q)$. The maximum of this lower bound can be achieved when
the approximate distribution $\mathcal{Q}(\boldsymbol{\Theta}_k)$ is
exactly equal to the desired posterior distribution
$p(\boldsymbol{\Theta}_k|\tilde {\bf {y}}_k)$.

The joint distribution $p(\boldsymbol{\Theta}_k,\tilde {\bf  {y}}_k)$
can be decomposed as \cite{bishop}
\begin{equation}\label{eq19}
\begin{split}
p(\boldsymbol{\Theta}_k,\tilde {\bf  {y}}_k) = p(\tilde {\bf  {y}}_k,{\bf B}_k, {\bf C}_k,{\boldsymbol{\mu}}_k,{\bf \Sigma}_k)~~~~~~~~~~~~~~~~~~~~~~~~~~~~~~~~~\\
\!\!\!\!\!= p(\tilde {\bf  {y}}_k|{\bf B}_k,{\boldsymbol{\mu}}_k,{\bf \Sigma}_k)p({\bf B}_k|{\bf C}_k)p({\bf
C}_k) p({\boldsymbol{\mu}}_k|{\bf \Sigma}_k)p({\bf \Sigma}_k).
\end{split}
\end{equation}
We assume that the variational distribution $\mathcal{Q}(\boldsymbol{\Theta}_k)= \mathcal{Q}({\bf B}_k, {\bf C}_k,{\boldsymbol{\mu}}_k,{\bf \Sigma}_k)$ can be factorized between latent variable ${\bf B}_k$ and parameters ${\bf C}_k$, ${\boldsymbol{\mu}}_k$ and ${\bf \Sigma}_k$.
\begin{equation}
\mathcal{Q}({\bf B}_k, {\bf C}_k,{\boldsymbol{\mu}}_k,{\bf \Sigma}_k) = \mathcal{Q}({\bf B}_k)\mathcal{Q}({\bf C}_k,{\boldsymbol{\mu}}_k,{\bf \Sigma}_k).
\end{equation}
A similar assumption is made in \cite{bishop}, please see equation (10.42). Optimization of the variational distribution can therefore be represented as
\begin{equation}
\mathcal{Q}^*({\bf B}_k, {\bf C}_k,{\boldsymbol{\mu}}_k,{\bf \Sigma}_k) = \mathcal{Q}^*({\bf B}_k)\mathcal{Q}^*({\bf C}_k,{\boldsymbol{\mu}}_k,{\bf \Sigma}_k).
\end{equation}
By considering equation (10.54) from \cite{bishop} the distribution $\mathcal{Q}^*({\bf C}_k,{\boldsymbol{\mu}}_k,{\bf \Sigma}_k)$ can further be decomposed into $\mathcal{Q}^*({\bf C}_k),\mathcal{Q}^*({\boldsymbol{\mu}}_k,{\bf \Sigma}_k)$, where $(\cdot)^*$ represents the optimum distribution. Therefore, the optimum distribution can be written as
\begin{equation}
\mathcal{Q}^*({\bf B}_k, {\bf C}_k,{\boldsymbol{\mu}}_k,{\bf \Sigma}_k) = \mathcal{Q}^*({\bf B}_k)\mathcal{Q}^*({\bf C}_k)
\mathcal{Q}^*({\boldsymbol{\mu}}_k,{\bf \Sigma}_k).
\end{equation}
This shows that the optimum over the joint distribution is equivalent to obtaining $\mathcal{Q}^*({\bf B}_k)$, $\mathcal{Q}^*({\bf C}_k)$
and $\mathcal{Q}^*({\boldsymbol{\mu}}_k,{\bf \Sigma}_k)$.
Therefore, the optimum distributions over $\boldsymbol{\Theta}_k$ can be evaluated by optimizing $L(q)$ with respect to all parameters one-by-one. A general form of optimization can be written as in equation (10.9) of \cite{bishop}
\begin{equation}\label{eq18}
\ln \mathcal{Q}^*(\boldsymbol{\Theta}^c_k) = E_{d\neq c}[\ln
p(\boldsymbol{\Theta}_k,\tilde {\bf  {y}}_k)]+constant,
\end{equation}
where $E_{d \neq c}[.]$  represents the mathematical expectation
with respect to the distributions over all the elements in
$\boldsymbol{\Theta}_k$ for $d \neq c$, where $d$ is the $d^{th}$
component of $\boldsymbol{\Theta}_k$.
$\mathcal{Q}^*(\boldsymbol{\Theta}^c_k)$ represents the optimum
approximate distribution over the $c^{th}$ component of
$\boldsymbol{\Theta}_k$. We next evaluate the optimum distributions
$\mathcal{Q}^*({\bf B}_k|{\bf C}_k)$, $\mathcal{Q}^*({\bf C}_k)$ and
$\mathcal{Q}^*({\boldsymbol{\mu}}_k^q, {\bf \Sigma}_k^q)$ by using
equation (\ref{eq18}).
\subsubsection{Optimum distribution over ${\bf B}_k$}
According to equation (\ref{eq18}), the optimum distribution over ${\bf B}_k$ can be written as
\begin{equation}\label{eq18b}
\ln \mathcal{Q}^*({\bf B}_k) = E_{{\bf C}, \boldsymbol{\mu}_k,
\boldsymbol{\Sigma}_k}[\ln p({\bf B}_k,  \tilde {\bf  {y}}_k, {\bf
C}_k,\boldsymbol{\mu}_k, \boldsymbol{\Sigma}_k)]+constant.
\end{equation}\\
Probabilities $p({\bf B}_k|{\bf C}_k)$ and $p(\tilde {\bf  {y}}_k|{\bf
B}_k,{\boldsymbol{\mu}}_k,{\bf \Sigma}_k)$ can be defined by using
equations (\ref{eq10}) and (\ref{eq12}) respectively
\begin{equation}\label{eq22}
p({\bf B}_k|{\bf C}_k) = \prod_{j=1}^M\prod_{q=1}^{\kappa}{({\bf C}_k^{q})}^{b_k^{j,q}},
\end{equation}
and
\begin{equation}\label{eq23}
p(\tilde {\bf  {y}}_k|{\bf B}_k,{\boldsymbol{\mu}}_k,{\bf \Sigma}_k) =
\prod_{j=1}^M\prod_{q=1}^{\kappa}\mathcal{N}{(\tilde {\bf  {y}}_k^j|{\boldsymbol{\mu}}_k^q,{\bf \Sigma}_k^q)}^{b_k^{j,q}}.
\end{equation}

By using equations (\ref{eq18b}), (\ref{eq22}) and (\ref{eq23}), the optimum distribution over the correspondence variable ${\bf B}_k$ becomes
\begin{equation}\label{eq24}
\mathcal{Q}^*({\bf B}_k) =  \prod_{j=1}^M\prod_{q=1}^{\kappa}{(r_k^{j,q})}^{b_k^{j,q}},
\end{equation}
where $r_k^{j,q}$ is the responsibility that component $q$ takes to
explain the measurement $\tilde {\bf  {y}}_k^j$.
The derivation of equation (\ref{eq24}) and
$r_k^{j,q}$ can be found in Appendix-A.

\subsubsection{Optimum distribution over ${\bf C}_k$}
Before evaluating the optimum distributions $\mathcal{Q}^*({\bf
C}_k)$ and $\mathcal{Q}^*({\boldsymbol{\mu}}_k, {\bf \Sigma}_k)$ we need
to first define their priors. The Dirichlet distribution is chosen
as a prior over the mixing coefficient ${\bf C}_k$
\begin{equation}\label{eq20}
p({\bf C}_k) = Dir({\bf C}_k|\alpha_\circ),
\end{equation}
where $Dir(\cdot)$ denotes the Dirichlet distribution, and $\alpha_\circ$ is an effective prior number of observations associated with each component of the mixture.
Using  equation (\ref{eq18}) we can write
\begin{equation}\label{eq18c}
\ln \mathcal{Q}^*({\bf C}_k, {\boldsymbol{\mu}}_k,
\boldsymbol{\Sigma}_k) = E_{{\bf B}} [\ln p(\tilde {\bf {y}}_k, {\bf
B_k}, {\bf C}_k,\boldsymbol{\mu}_k, \boldsymbol{\Sigma}_k)]+const,
\end{equation}
where const. represents a constant. The optimum distributions $\mathcal{Q}^*({\bf C}_k)$ over ${\bf C}_k$ can be calculated by using equations (\ref{eq23}), (\ref{eq20}) and (\ref{eq18c}), which becomes
\begin{equation}\label{eq25}
\mathcal{Q}^*({\bf C}_k) = Dir({\bf C}_k|{\boldsymbol \alpha_k}),
\end{equation}
where ${\boldsymbol \alpha}_k = [\alpha_k^1,\ldots,\alpha_k^\kappa]
$ and one of its components $\alpha_k^q$ can be defined as
\begin{equation}\label{eq26}
\alpha_k^q = \alpha_\circ + N_k^q,
\end{equation}
\begin{equation}\label{eq27}
N_k^q = \sum_{j=1}^{M}r_k^{j,q},
\end{equation}
and $r_k^{j,q}$ is the responsibility that component $q$ takes to
explain the measurement $\tilde {\bf  {y}}_k^j$. The derivation
of~(\ref{eq25}) can be found in Appendix-B.
\subsubsection{Optimum distribution over ${\boldsymbol{\mu}}_k$ and ${\bf \Sigma}_k$ }
The prior over the mean ${\boldsymbol{\mu}}_k$ and the covariance matrix ${\bf \Sigma}_k$ is defined by the independent Gaussian-Wishart distribution
\begin{equation}\label{eq21}
p({\boldsymbol{\mu}}_k|{\bf \Sigma}_k)p({\bf \Sigma}_k) = \prod_{q=1}^{\kappa}\mathcal{N}({\boldsymbol{\mu}}_k^q|{\bf m}_\circ,\beta_\circ^{-1}{\bf \Sigma}_k^q)\mathcal{W}({\bf \Sigma}_k^q|{\bf W}_\circ,\upsilon_\circ),
\end{equation}
where ${\bf m}_\circ$, $\beta_\circ$, ${\bf W}_\circ$ and $\upsilon_\circ$ are the prior parameters.
By using equations (\ref{eq23}), (\ref{eq20}) and (\ref{eq18c}), decomposition of equation (\ref{eq19}) and following the steps explained in Appendix-C, the distribution becomes 
\begin{equation}\label{eq28}
\mathcal{Q}^*({\boldsymbol{\mu}}_k|{\bf \Sigma}_k)\mathcal{Q}^*({\bf \Sigma}_k) = \prod_{q=1}^{\kappa}\mathcal{N}({\boldsymbol{\mu}}_k^q|{\bf m}_q,\beta_q^{-1}{\bf \Sigma}_k^q)\mathcal{W}({\bf \Sigma}_k^q|{\bf W}_q,\upsilon_q),
\end{equation}
where ${\bf m}_q, \beta_q, {\bf W}_q$ and $\upsilon_q$ are defined from
\begin{equation}\label{eq29}
\beta_q = \beta_\circ + N_k^q,
\end{equation}
\begin{equation}\label{eq30}
{\bf m}_q = \frac{1}{\beta_q}(\beta_\circ + N_k^q {\bf \bar{y}}_k^q),
\end{equation}
where, 
\begin{equation}\label{eq33}
{\bf \bar{y}}_k^q = \frac{1}{N_k^q}\sum_{j=1}^{M}r_k^{j,q}\tilde {\bf  {y}}_k^j,
\end{equation}
\begin{equation}\label{eq31}
{\bf W}_q^{-1} = {\bf W}_\circ^{-1} + N_k^q{\bf S}_k^q + \frac{\beta_\circ N_k^q}{\beta_\circ + N_k^q}({\bf \bar{y}}_k^q - {\bf m}_\circ){({\bf \bar{y}}_k^q - {\bf m}_\circ)}^T,
\end{equation}
\begin{equation}\label{eq32}
\upsilon_q = \upsilon_\circ + N_k^q,
\end{equation}
and 
\begin{equation}\label{eq34}
{\bf S}_k^q = \frac{1}{N_k^q}\sum_{j=1}^{M}r_k^{j,q}(\tilde {\bf  {y}}_k^j
-{\bf \bar{y}}_k^q) {(\tilde {\bf  {y}}_k^j -{\bf \bar{y}}_k^q)}^T.
\end{equation}

The variational Bayesian technique operates in two steps to optimize the posterior distributions of unknown variables and parameters. In the first step it calculates the responsibilities $r_k^{j,q}$ using equation (\ref{aa15})
and in the second step it uses these responsibilities to optimize the distributions by using equations (\ref{eq24}), (\ref{eq25}) and (\ref{eq28}). These steps are repeated until some convergence criterion is met. In our work we monitor the lower bound $L(q)$ after every iteration to test the convergence. When the algorithm converges, the value of the lower bound does not change more than a small amount. The clustering algorithm is further summarized in {\bf Algorithm~2} given in Section~\ref{vnt}.

One of the important advantages of variational Bayesian clustering
is that it can automatically determine the number of clusters by
using the measurements. This can be achieved if we set the parameter
$\alpha_\circ$ less then $1$. This helps to obtain a solution which
contains a minimum number of clusters to represent the
data~\cite{tinne}.

The position and shape of clusters are defined using the parameters
${\bf m}_\circ$ and ${\bf W}_\circ$, respectively. A possible choice
for selecting these priors is given in Section~\ref{sec3}. This
stage returns the minimum possible number of clusters and their
associated measurements which are defined by the correspondence
matrix ${\bf B}_k$. In the next stage of the algorithm these
clusters are assigned to the targets by using a data association
technique as described after Section \ref{ztot}.

\section{Variable Number of Targets}
\label{vnt}
Many multi-target tracking algorithms assume that the number of
targets is fixed and
known~\cite{Mohsin-j,color,Godsill,Godsill-Group1}. In \cite{sjpdaf}
the JPDAF is extended to a variable number of targets. In other
works, including~\cite{tinne} the posterior probability of the
number of targets $N$ is estimated given the number of clusters
$\kappa$ at each discrete time step $k$
\begin{equation}\label{eq62}
\begin{split}
p(N_k|\kappa_{1:k})\!\! \propto\!\! p(\kappa_k|N_k)\!\!\sum_n\{p(N_k|N_{k-1}\!=\!n)
 p(N_{k-1}\!=\!n|\kappa_{1:k-1})\},
\end{split}
\end{equation}
where $p(\kappa_k|N_k)$ is the probability of the $\kappa_k$
clusters given $N_k$ targets. 
Here we deal with changeable shapes and calculate the number of
targets in a different way. The number of targets is estimated based
on the: 1) location ${\boldsymbol{\mu}}_k^q$ of clusters at time step
$k$, 2) size of clusters at time step $k$, and 3) position vector
${\bf x}_{k-1}$ of the targets at the previous time step $k-1$.

The number 
of targets in the first video frame is determined based on the
variational Bayesian clustering. At every following video frame the
number $N_k$ of targets is evaluated, by the following stochastic
relationship, similarly to~\cite{Ng:2007:MIT:1349545.1349557}
\begin{equation}\label{eq-vt-a}
N_k = N_{k-1} + \mathcal{\varrho}_k.
\end{equation}
%
The variable $\mathcal{\varrho}_k \in \{-1,0,1\}$ is defined as
\begin{equation}\label{eq-vt}
\mathcal{\varrho}_k\!\! =\!\! \left\{\begin{array}{ll}
         \!\!-1 & \!\!\!\textrm{if}~p(death)\!>\!p(birth)~\&~p(death)\! >\! Thr\\
         \!\!  1 & \!\!\!\textrm{if}~p(birth)\!>\!p(death)~\&~p(birth)\! >\! Thr\\
         \!\! 0 &\!\!\! \textrm{otherwise}
          \end{array}\right.
\end{equation}
where $p(death)$ corresponds to the probability of decrementing the number of targets, similarly, $p(birth)$ represents the probability of incrementing number of targets and $Thr$ is an appropriately chosen threshold.  

We assume that at a given time only one target can enter or leave.
In the monitored area people can only enter or leave through a
specifically known region in a video frame. This known region is
called the \emph{red region}. This red region is modeled by a
Gaussian distribution with its center $\boldsymbol{\mu}^r$ and
covariance matrix~$\boldsymbol{\Sigma}^r$. Multiple Gaussian distributions can be
used if the red region was disjoint.
\subsection{Probability of death}
The probability of death of a target depends on two things: 1) the
probability $p({\bf Z}_k^r)$ of finding a cluster ${\bf Z}_k^r$ in
the red region and 2) the probability $p(Existing~Target|{\bf
Z}_k^r,{\bf x}_{k-1})$ that the cluster found in the red region is
due to an existing target (denoted by the variable
$Existing~Target$). This means that when a target is leaving the
monitored area, a cluster is found in the red region, i.e. at the
entrance or exit point, and this cluster will be due to an existing
target. Therefore, probability of death is modeled as
\begin{equation}\label{eq63}
p(death) = p({\bf Z}_k^r)p(Existing~Target|{\bf Z}_k^r,{\bf
x}_{k-1}).
\end{equation}
In order to estimate $p({\bf Z}_k^r)$, we calculate the probability
of the measurements (foreground pixels) from
\begin{equation}\label{eq63-vt}
p({\bf {\tilde y}}_k^j|{\boldsymbol{\mu}}^r,{\bf \Sigma}^r) =
\mathcal{N}({\bf {\tilde y}}_k^j|{\boldsymbol{\mu}}^r,{\bf \Sigma}^r).
\end{equation}
Note that we are not considering measurements of one cluster only,
because a target in the red region may generate more than one
cluster. Therefore, measurements which have probability $p(\tilde
{\bf  {y}}_k^j| {\boldsymbol{\mu}}^r,{\bf \Sigma}^r)$ greater than a
set threshold value are considered as measurements found in the red
region and all these measurements are therefore grouped to form
cluster ${\bf Z}_k^r$. Since we assume that maximum one target can
be in the red region, then, all the measurements of ${\bf Z}_k^r$
are considered to be generated by one target. For an $N_p$ number of
measurements in cluster ${\bf Z}_k^r$, the probability of finding a
cluster in the red region depends on the number of pixels in ${\bf
Z}_k^r$ and is calculated as
\begin{equation}
p({\bf Z}_k^r) = 1- \exp(-\frac {N_p}{\Delta_p}),
\end{equation}
where ${\Delta_p}$ is a pixel constant which limits the range of pixels to be considered
for calculating $p({\bf Z}_k^r)$.
The second probability in equation (\ref{eq63}) is the probability
that cluster ${\bf Z}_k^r$ is generated by an existing target. This
probability depends on the location of existing targets and the
location of cluster ${\bf Z}_k^r$. The probability
$p(Existing~Target|{\bf Z}_k^r,{\bf x}_{k-1})$ is calculated as
\begin{equation}
p(Existing~Target|{\bf Z}_k^r,{\bf x}_{k-1}) = \exp(-\frac
{\tilde{d}_{min}}{\Delta_d}),
\end{equation}
where ${\Delta_d}$ is a constant which can be chosen experimentally.
To calculate the distance $\tilde{d}_{min}$, we chose the
minimum distance among the distances which are found from the
centroid of ${\bf Z}_k^r$, $centroid({\bf Z}_k^r)$ to each of the existing targets at time
step $k-1$. Note that this $\tilde{d}_{min}$ is different from the
$d_{min}$ calculated in equation (\ref{eq45}).
\begin{equation}\label{eq:Num_obj}
\tilde{d}_{min} = \min\limits_{i= 1 \dots N_{k-1}}
\parallel centroid({\bf Z}_k^r) - {\bf
x}_{k-1}^{i}\parallel,
\end{equation}
where $\parallel\cdot\parallel$ denotes the Euclidean distance
operator. Note that the probability ${\bf Z}_k^r$ depends on the
distance between the centroid of ${\bf Z}_k^r$ and the closest
existing target. The probability that cluster ${\bf Z}_k^r$ is
generated by an existing target will be high if $\tilde{d}_{min}$ is
small and vice versa.
\subsection{Probability of Birth}
The probability of birth of a target depends on two different
probabilities: 1) the probability of finding a cluster ${\bf Z}_k^r$
in the red region $p({\bf Z}_k^r)$ and 2) the probability that the
cluster found in the red region is not due to an existing target.
According to our assumption that at a given time step the cluster in
the red region can only be either from an existing target or a new
target, the probability that the cluster in the red region is due to
a new target is $ 1 - p(Existing~Target|{\bf Z}_k^r,{\bf x}_k)$. The
probability of birth can be calculated as
\begin{equation}\label{eq64}
p(birth) = p({\bf Z}_k^r) (1 - p(Existing~Target|{\bf Z}_k^r,{\bf x}_k)).
\end{equation}
Finally, (\ref{eq63}) and (\ref{eq64}) are applied in (\ref{eq-vt-a})
and the number of targets is updated by using (\ref{eq:Num_obj}).
This completes the full description of our algorithm. A summary of
the overall proposed algorithm is described in {\bf Algorithm~2}.

\begin{algorithm}\label{algo}
\caption{Summary of Proposed Algorithm}
Input: Video frame at time step $k$\\
Output: 2-D position of all the targets in each frame $k$\\
\begin{algorithmic}[1]
\STATE Perform the background subtraction to extract $M$ foreground pixels.
\STATE Save the coordinates of foreground pixels which represent data points $\tilde {\bf  {y}}_k$.\\
\vspace{0.2cm}
\underline{\bf Clustering}\\
\vspace{0.2cm}
\STATE Initialize the parameters $\alpha_\circ$, $\beta_\circ$, ${\bf m}_\circ$ ${\bf W}_\circ$ and $\upsilon_\circ$.
\FOR{$j = 1,...,M$} \label{step1}
\FOR{$q = 1,...,\kappa$}
\STATE Evaluate initial ${\boldsymbol{\rho}}^{j,q}_{\circ}$ by using initialized parameters $\alpha_\circ$,
$\beta_\circ$, ${\bf m}_\circ$ ${\bf W}_\circ$ and $\upsilon_\circ$ in
equations (\ref{aa11}) and (\ref{aa11-b}).
\ENDFOR
\STATE Calculate the normalization factor $\tilde{\eta} = {\sum_{q=1}^{\kappa}{\boldsymbol{\rho}}^{j,q}_{\circ}}$
\FOR{$q = 1,...,\kappa$}
\STATE Evaluate normalized responsibilities: ${r}_\circ^{j,q} = \frac{{r}_\circ^{j,q}}{\tilde{\eta}}$.
\ENDFOR
\ENDFOR \label{step2}
\WHILE{Convergence criterion is not satisfied}
\FOR{$q = 1,...,\kappa$}
\STATE Evaluate new $N_k^q$, ${\bf \bar y}_{k}^{q}$, $\alpha_k^{q}$, $\beta_{q}$, ${\bf m}_{q}$, ${\bf W}_{q}^{-1}$, ${\bf \upsilon}_{q}$ and ${\bf S}_{k}^{q}$ with equations (\ref{eq27}), (\ref{eq33}), (\ref{eq26}), (\ref{eq29}), (\ref{eq30}), (\ref{eq31}), (\ref{eq32}) and (\ref{eq34}) respectively.
For first iteration use initial responsibilities ${r}_\circ^{j,q}$.

\ENDFOR
\STATE Evaluate new responsibilities $r_k^{j,q}$ for all $j$ and
$q$ by using new $N_k^q$, ${\bf \bar y}_{k}^{q}$, $\alpha_k^{q}$, $\beta_{q}$, ${\bf m}_{q}$, ${\bf W}_{q}^{-1}$, ${\bf \upsilon}_{q}$ and ${\bf S}_{k}^{q}$
and repeating steps (\ref{step1}) through (\ref{step2}).
\ENDWHILE
\STATE Assign the $l^{th}$ cluster to
measurement $\tilde {\bf  {y}}_k^j$, when $r_k^{j,l} = \max _{q =
1:\kappa}r_k^{j,q}$ and repeat it for all the measurements. \STATE
Delete the small clusters. \\
\vspace{0.2cm}
\underline{\bf Identify Number of Targets}\\
\vspace{0.2cm}
\STATE Evaluate probability of death and
birth by using equations (\ref{eq63}) and (\ref{eq64}) respectively.
\STATE Identify number of targets by using equation (\ref{eq-vt-a}).\\
\vspace{0.2cm}
\underline{\bf Data Association and Tracking}\\
\vspace{0.2cm}
\STATE Evaluate the cost matrix
by using equation (\ref{eq-cost}). \STATE Evaluate k-best hypotheses
by using Murty's algorithm \cite{murty's}.
\FOR {$i = 1,....,N$(where $N$ is the total number of targets)}
\STATE Draw $N_s$ samples by using the state transition equation
(\ref{fm5}) as explained in {\bf Algorithm~1}.
\FOR{$q = 1,...,\kappa$}
\STATE Evaluate the set of hypotheses $\boldsymbol{\Psi}_k^{i,q}$
which assign the $q^{th}$ cluster to the $i^{th}$ target. \STATE For
every hypothesis $\boldsymbol{\psi}_k \in \boldsymbol{\Psi}_k^{i,q}$
evaluate $\ln p(\boldsymbol{\psi}_k|\tilde {\bf  {y}}_{1:k})$ by using
equations (\ref{eq55-c}) and (\ref{eq61-z}).
\STATE Evaluate ${\mathcal{A}}_k^{i,q}$ with equation (\ref{eq36-c}).
\ENDFOR
\STATE Weight all the particles using equation (\ref{eq54-c}).
\STATE Update states of all the targets according to equation (\ref{eq52-pf}).
\ENDFOR
\end{algorithmic}
\end{algorithm}
\section{Experimental Results}
\label{sec3}
To examine the robustness of the algorithm to close interactions,
occlusions and varying number of targets, the algorithm is evaluated
by tracking a variable number of people in three different publicly
available video datasets: CAVIAR, PETS2006 and AV16.3. The test
sequences are recorded at a resolution of $288 \times 360$ pixels at
$25$ frames/sec and in total there are $45$ sequences. The
performance of the proposed algorithm is compared with recently
proposed techniques \cite{tinne}, \cite{zia} and \cite{Czyz}. All
the parameters are discussed in the following subsections. The
algorithm automatically detects and initializes the targets when
they enter the monitored area. For evaluation the tracking results
we convert the final tracked locations of targets from pixel coordinates
to the ground coordinates by using four point homography.
\subsection{Background Subtraction Results}
\begin{figure}[h]
\centering
$\begin{array}{c@{\hspace{0.15in}}c@{\hspace{0.15in}}c@{\hspace{0.15in}}c}
\includegraphics[width=0.70in]{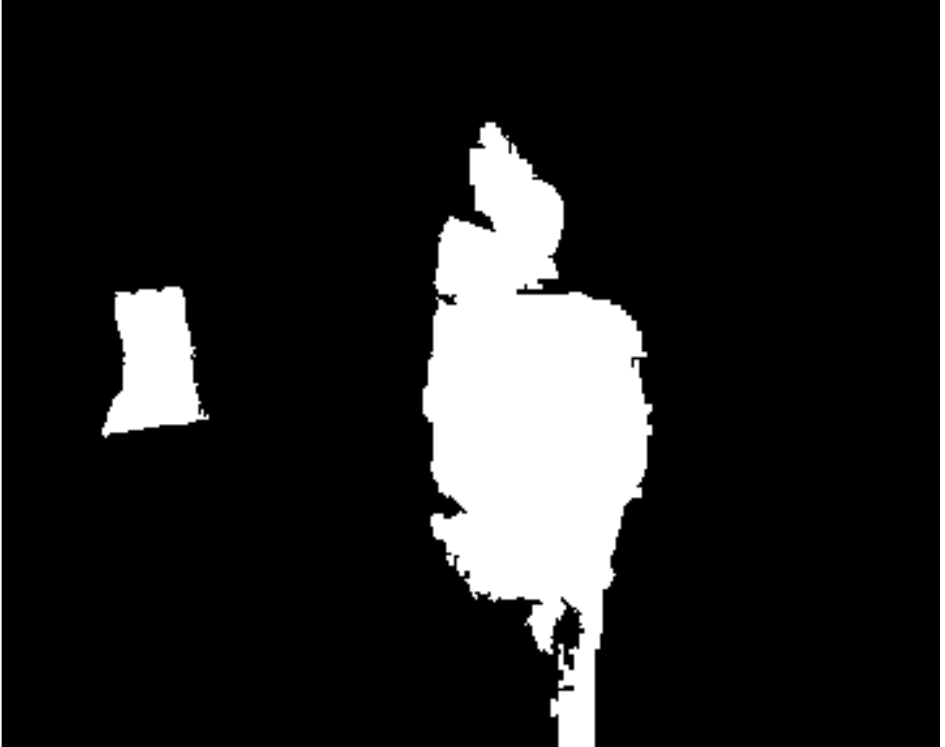} &
\includegraphics[width=0.70in]{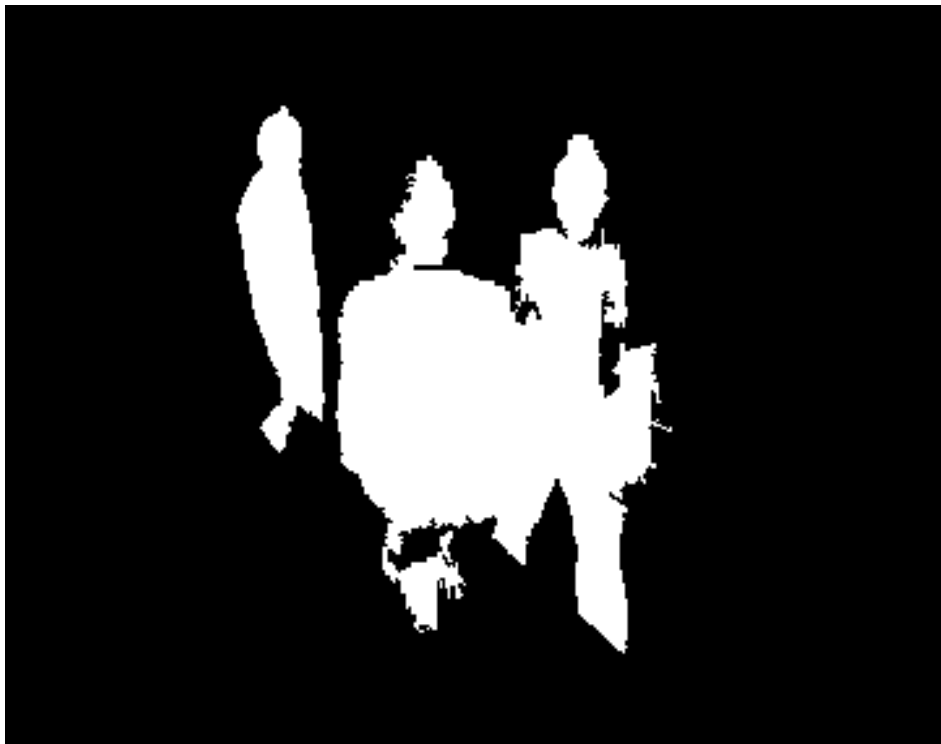} &
\includegraphics[width=0.70in]{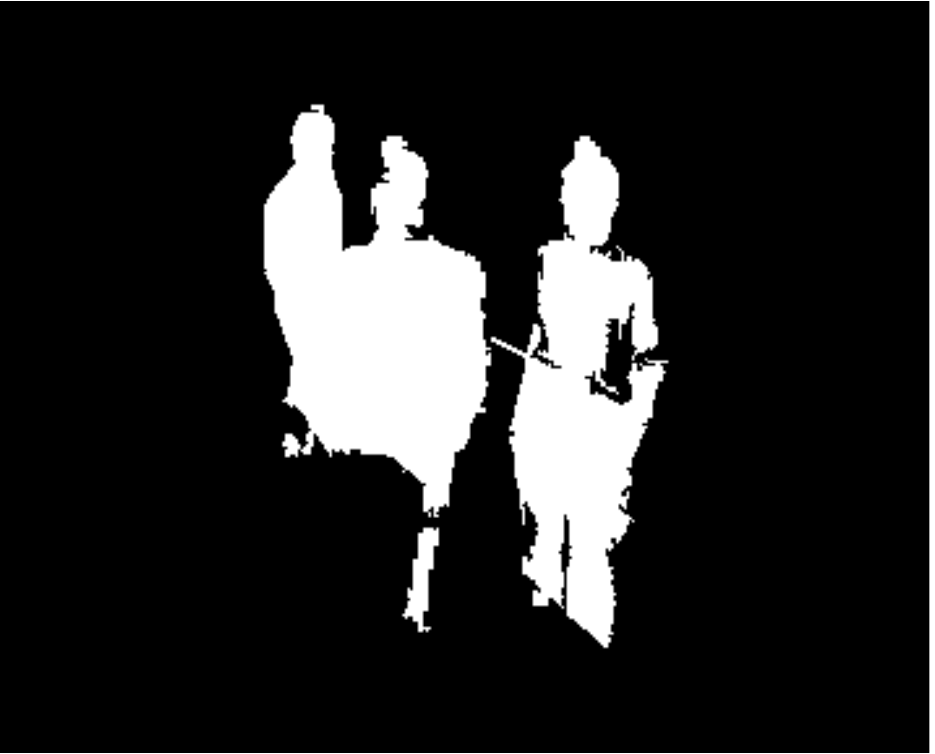}&
\includegraphics[width=0.70in]{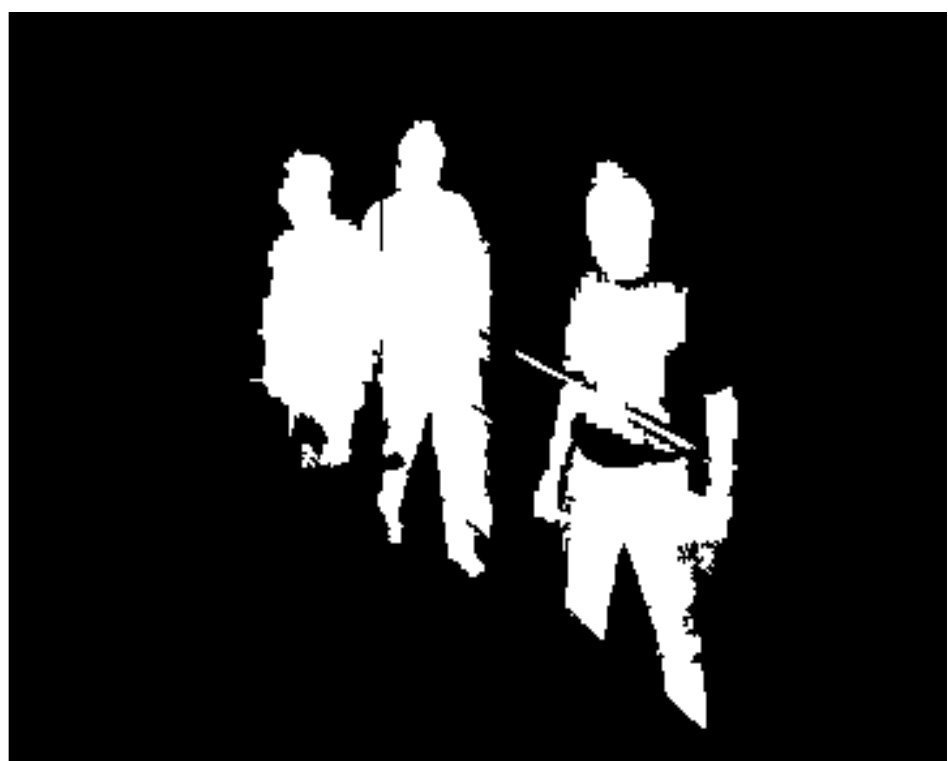}\\
\small{\mbox{(a) Frame 255}} & \small{\mbox{(b) Frame 278}}& \small{\mbox{(c) Frame 288}} & \small{\mbox{(d) Frame 313}}
\end{array}$
\caption{Background subtraction results for certain frames of sequence ``seq45-3p-1111\_cam3\_divx\_audio'' of the AV16.3 dataset: codebook background subtraction is implemented to separate the foreground pixels from the background.}
\label{fig1}
\end{figure}

\begin{figure}[h]
\centering
$\begin{array}{c@{\hspace{0.15in}}c@{\hspace{0.15in}}c@{\hspace{0.15in}}c}
\includegraphics[width=0.70in]{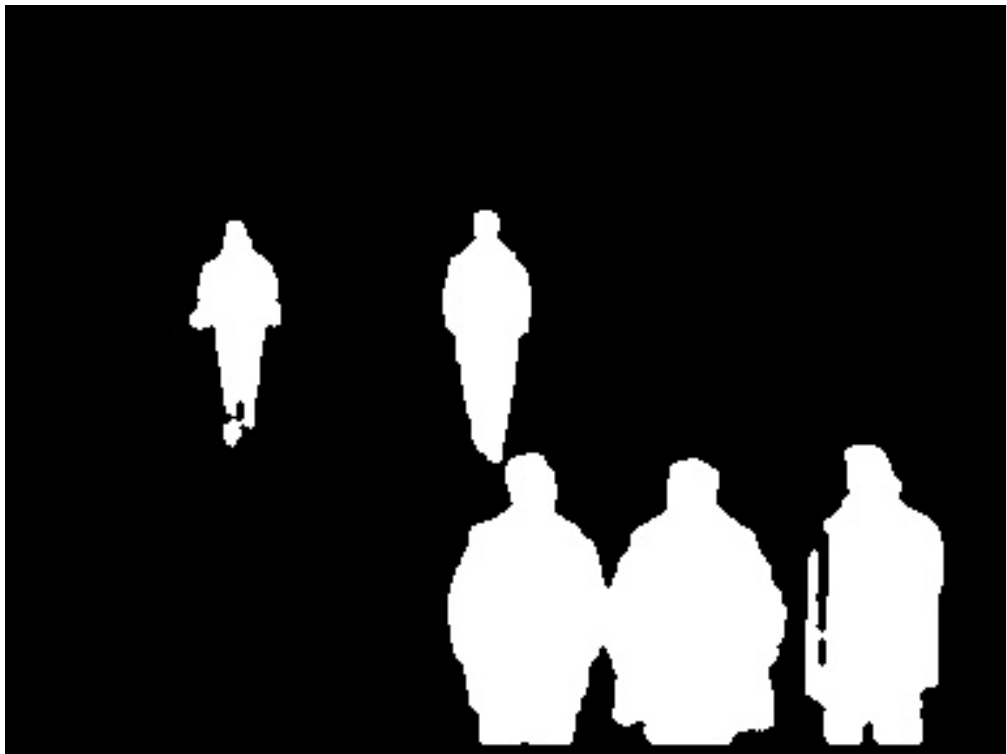} &
\includegraphics[width=0.70in]{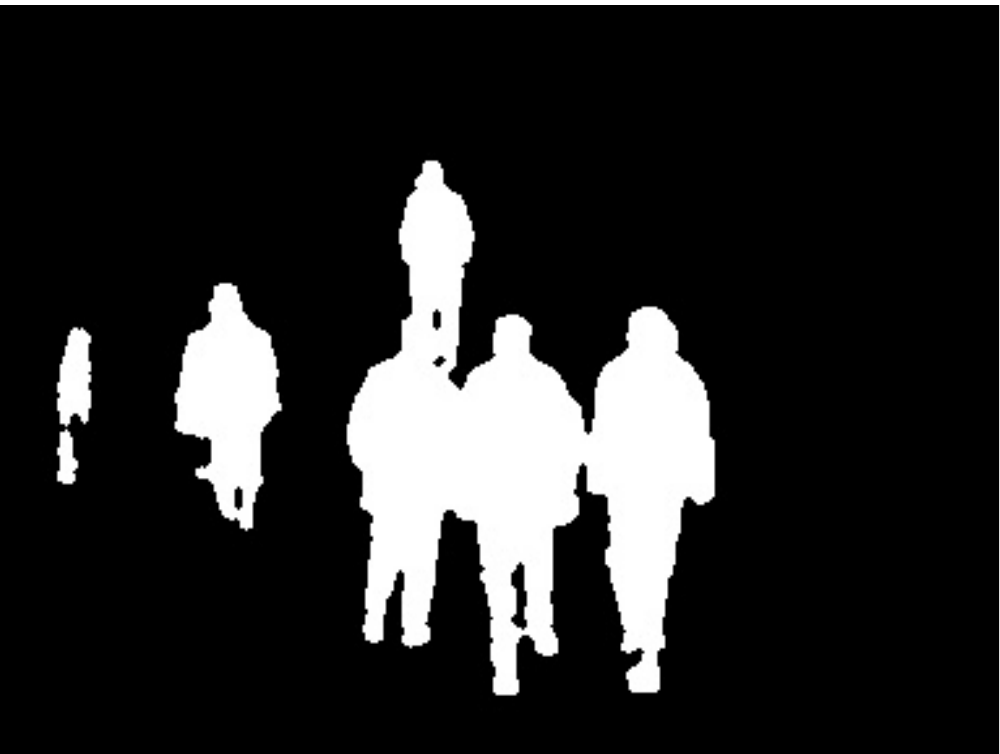}&
\includegraphics[width=0.70in]{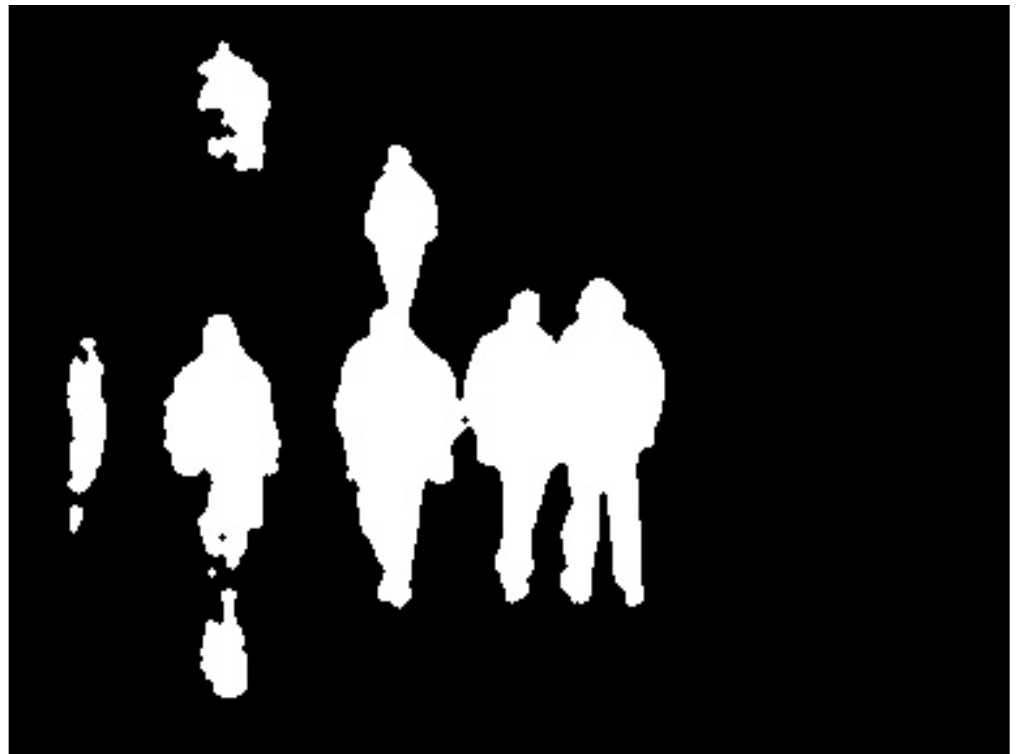}&
\includegraphics[width=0.70in]{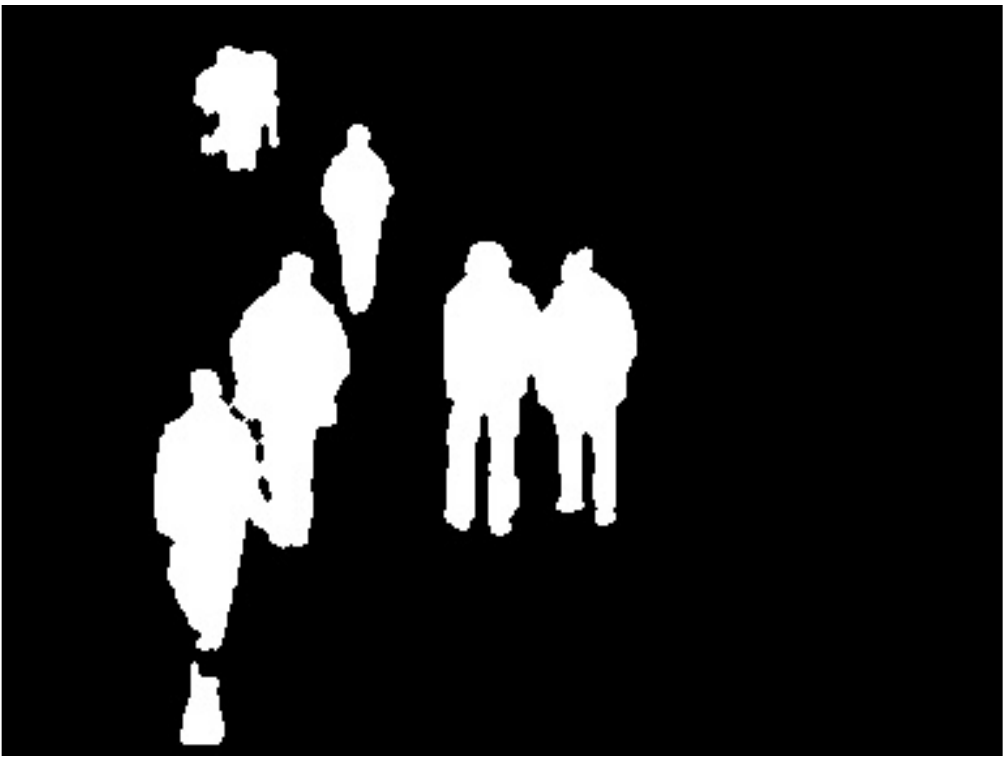}\\
 \small{\mbox{(a) Frame 334}}& \small{\mbox{(b) Frame 440}} & \small{\mbox{(c) Frame 476}} & \small{\mbox{(d) Frame 524}}\\[0.19cm]
\end{array}$
\caption{Background subtraction results for certain frames of sequence ``ThreePastShop2cor'' of the CAVIAR dataset: codebook background subtraction is implemented to separate the foreground pixels from the background.}
\label{figbs2}
\end{figure}

\begin{figure}[h]
\centering
$\begin{array}{c@{\hspace{0.15in}}c@{\hspace{0.15in}}c@{\hspace{0.15in}}c}
\includegraphics[width=0.70in]{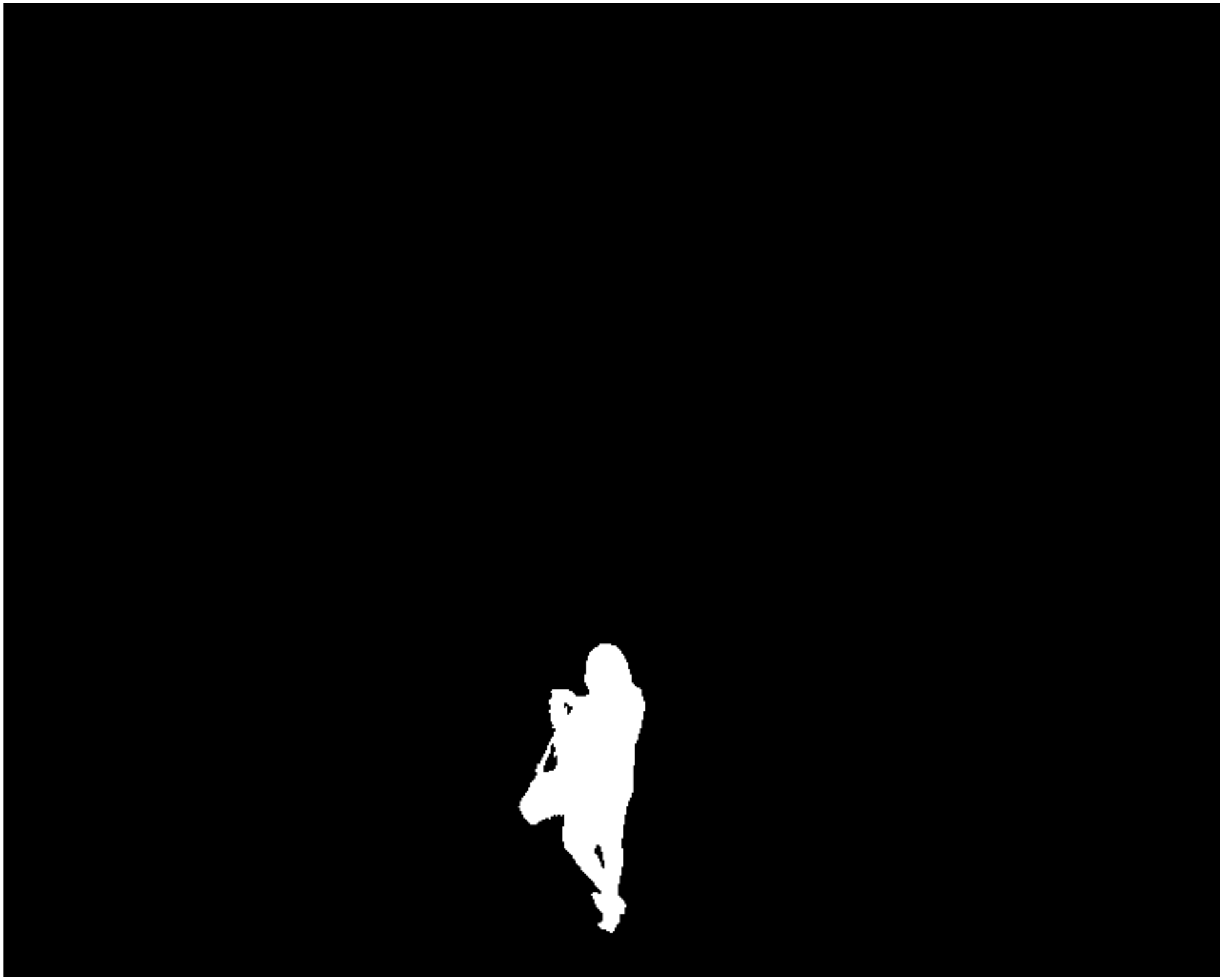} &
\includegraphics[width=0.70in]{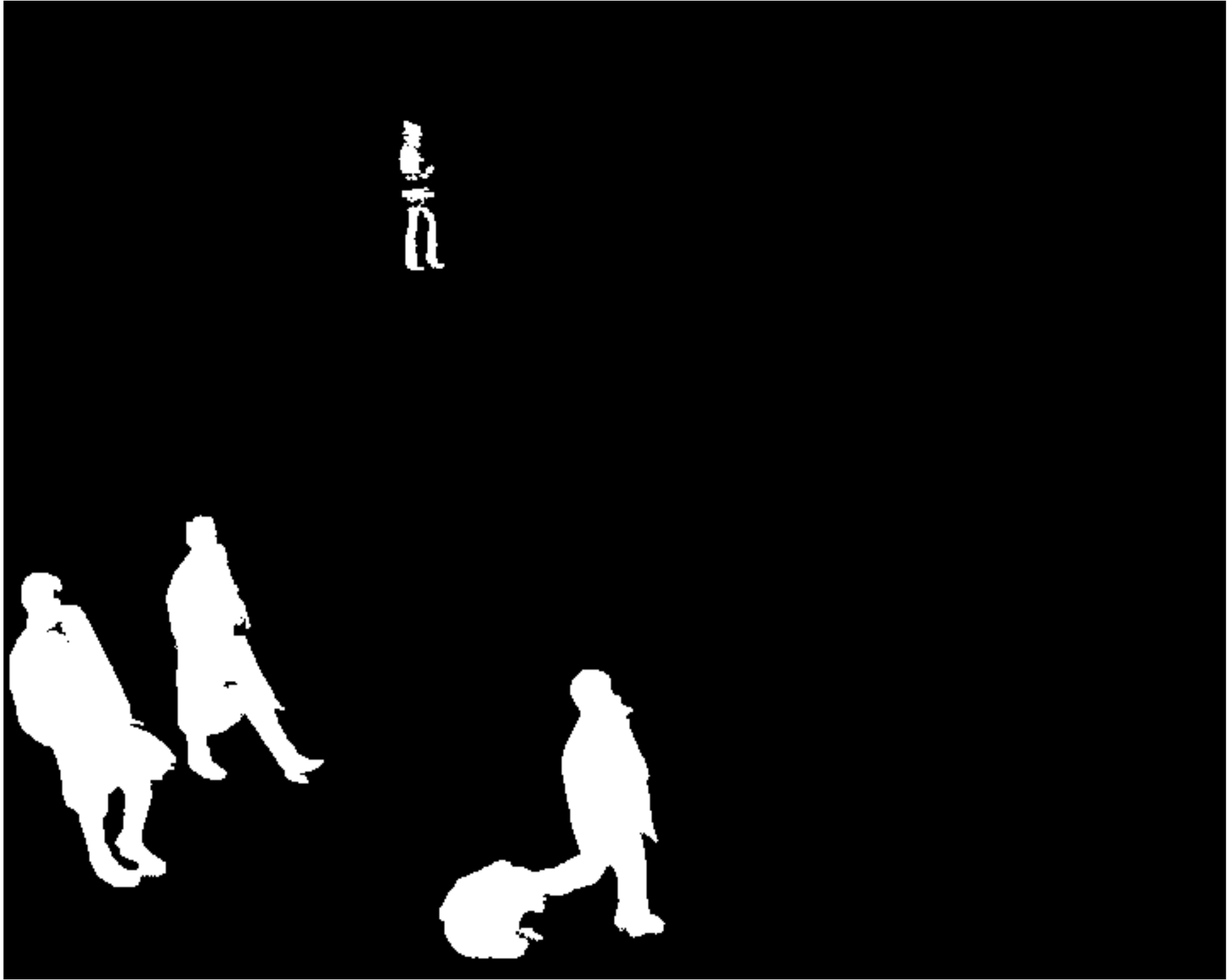} &
\includegraphics[width=0.70in]{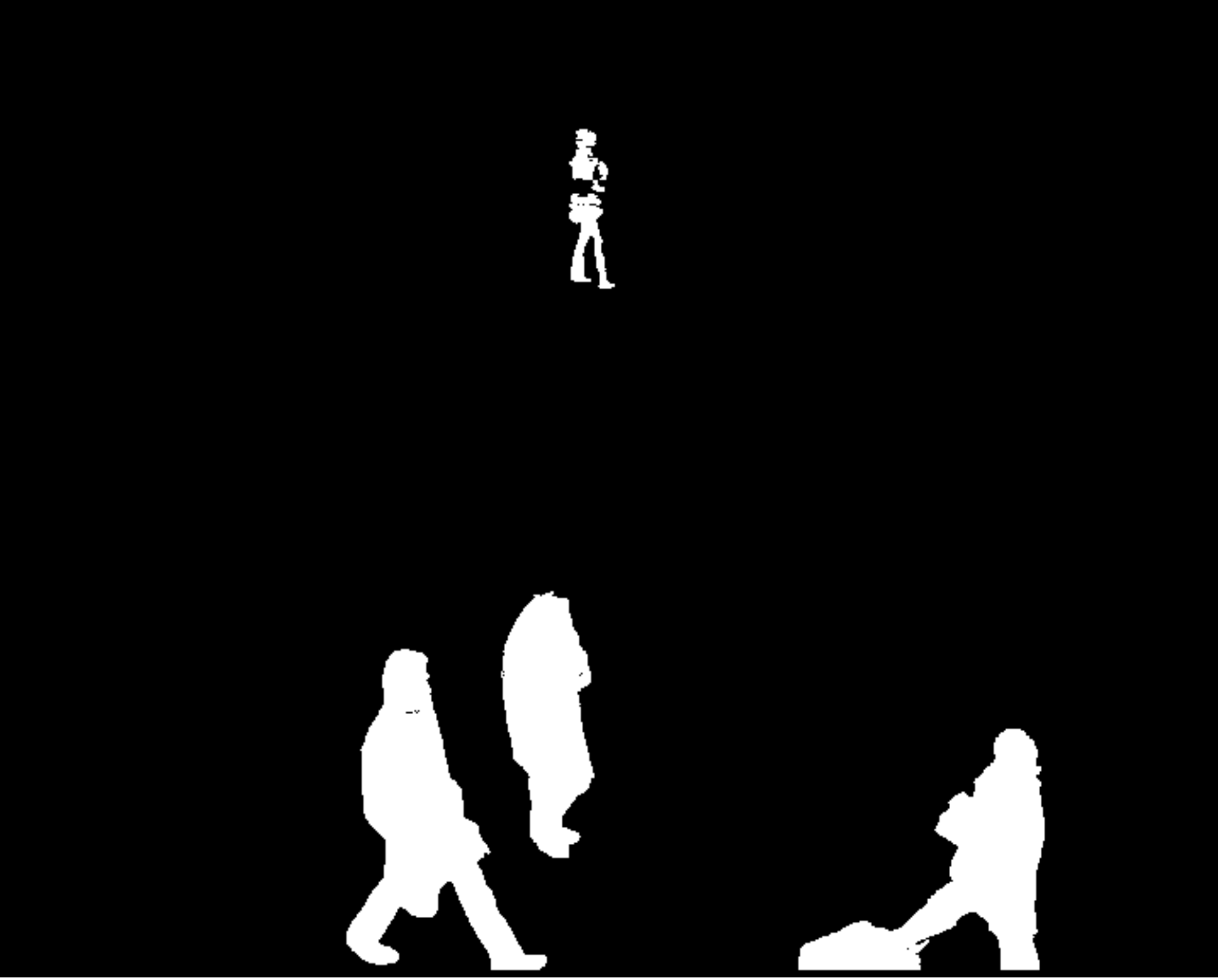}&
\includegraphics[width=0.70in]{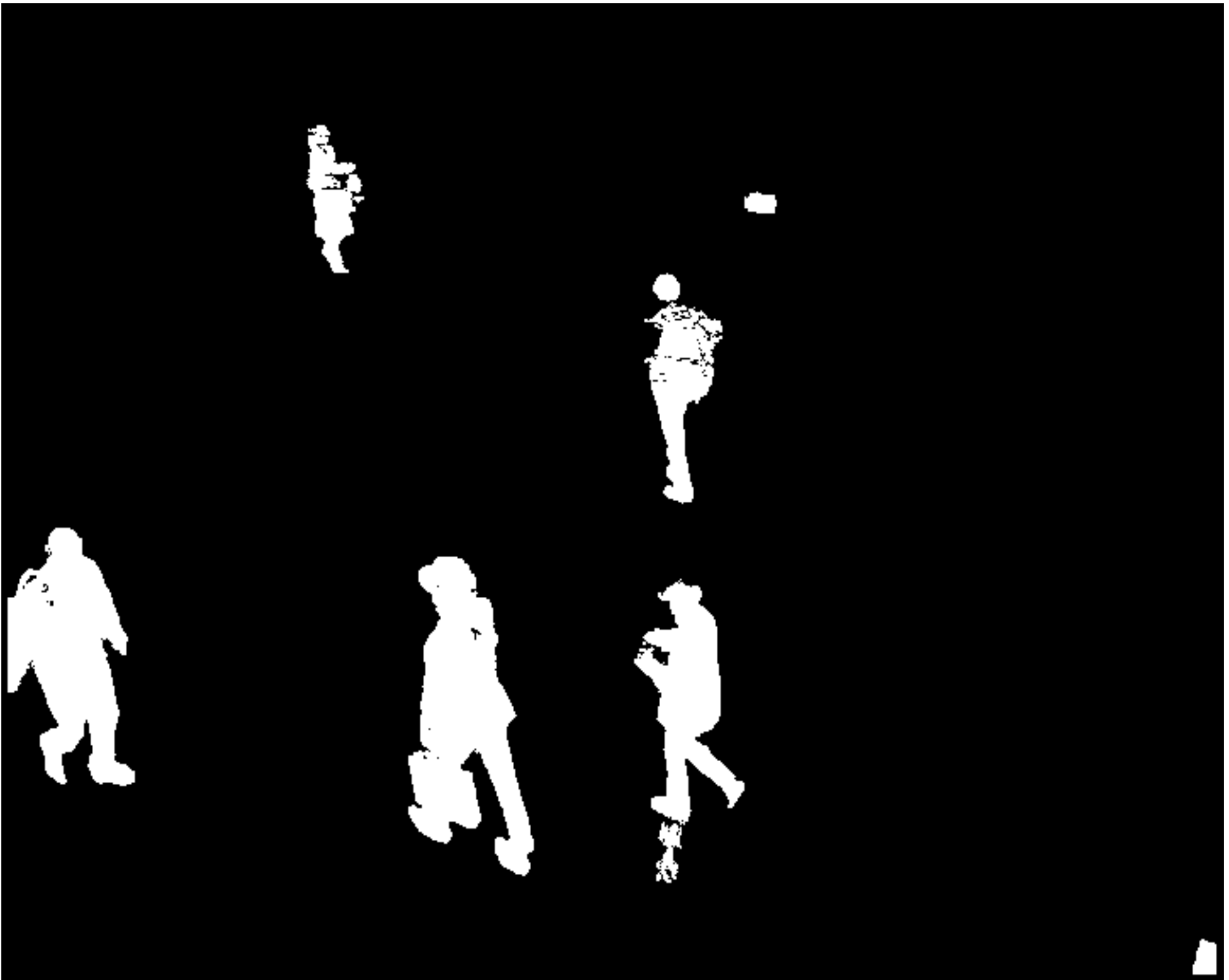}\\
\small{\mbox{(a) Frame 85}} & \small{\mbox{(b) Frame 292}}& \small{\mbox {(c) Frame 327}}& \small{\mbox{(d) Frame 1038}} \\[0.19cm]
\end{array}$
\caption{Background subtraction results for certain frames of sequence ``S1-T1-C'' of the PETS2006 dataset: codebook background subtraction is implemented to separate the foreground pixels from the background.}
\label{figbs3}
\end{figure}

The codebook background subtraction method~\cite{K.Kim} is
implemented which is one of the best background subtraction methods
since it is resistant to illumination changes and can capture the
structure of background motion. Adaptation to the background changes
and noise reduction is additionally achieved with the blob technique
\cite{M.Y}. Figs. \ref{fig1}, \ref{figbs2} and \ref{figbs3} show the
results obtained from the codebook background subtraction method
\cite{K.Kim} for a few selected video frames from the AV16.3, CAVIAR
and PETS2006 datasets respectively. In the experiment, we set the
shadow bound $\alpha=0.5$, highlight bound $\beta=2$ and the color
detection threshold $\varepsilon =20$ (see \cite{K.Kim} for further
details about these parameters). These parameters are the same for
all the sequences of all three datasets. The background subtraction
results provide the coordinates of the foreground pixels (the moving
objects) which represent data points ${\tilde{\bf y}}_k$. Frames
$440$ and $476$ in Fig. \ref{figbs2} show that we get a few extra
foreground pixels due to reflections on the floor and in the glass
which can be eliminated at the clustering stage.

\subsection{Clustering results}
For the clustering process at each time step we assume that there
are three types of clusters: 1) clusters for the existing targets,
2) clusters in the neighbors of the existing targets and 3) clusters
near boundaries of the field of view. Due to the Dirichlet prior
over the mixing coefficient and by setting $\alpha_\circ = 0.6$ the
clustering algorithm converges automatically to the minimum possible
number of clusters. The means ${\bf m}_\circ$ of clusters are
initialized with the tracked location of existing targets and
hypothesized location of new targets. Other prior parameters are
defined as: $\upsilon_\circ = 3$ and $\beta_\circ = 1$. However, a
full study about sensitivity of the system to the choice of
different parameters is beyond the scope of the paper.

\begin{figure}[h]
\centering
$\begin{array}{c@{\hspace{0.1in}}c@{\hspace{0.1in}}c@{\hspace{0.1in}}c}
\includegraphics[width=0.75in]{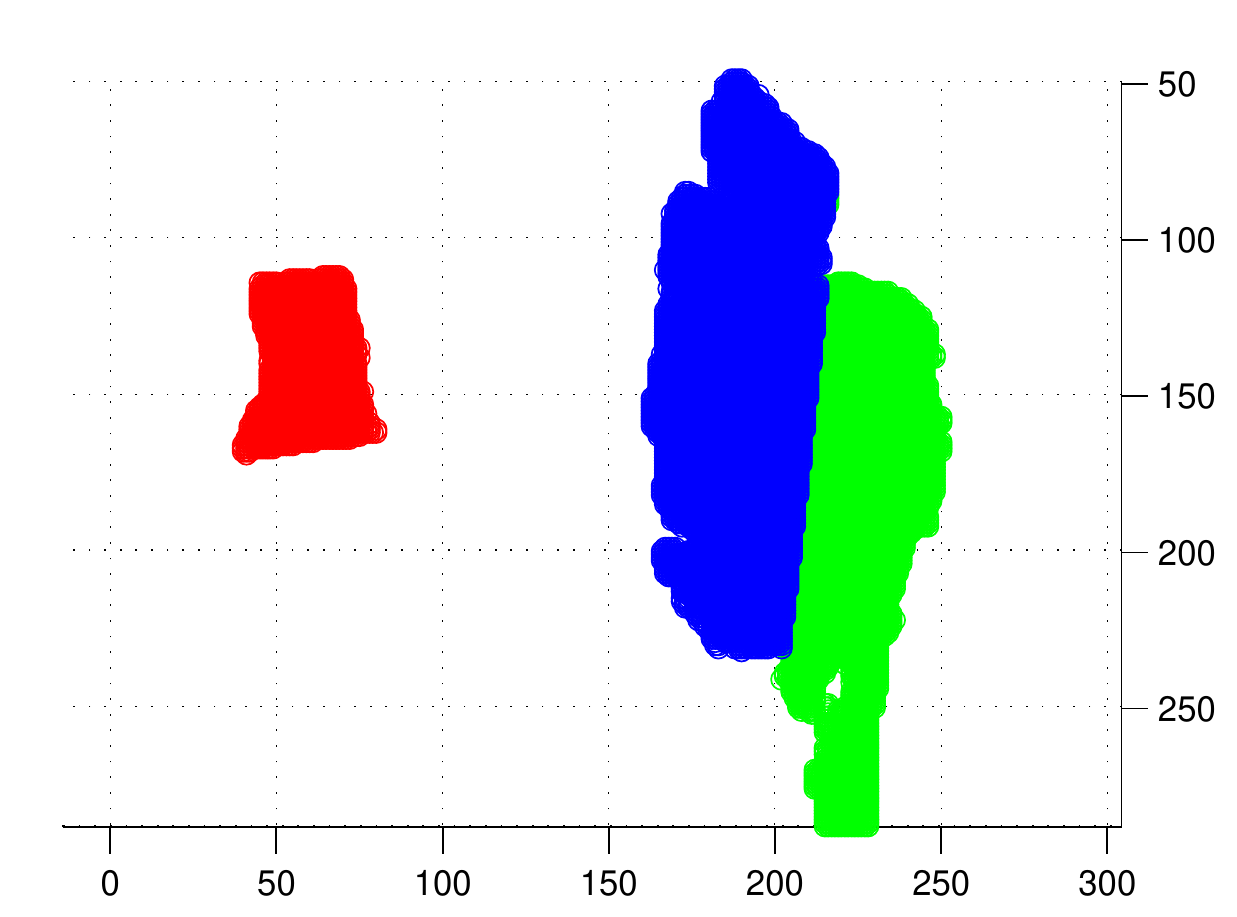} &
\includegraphics[width=0.75in]{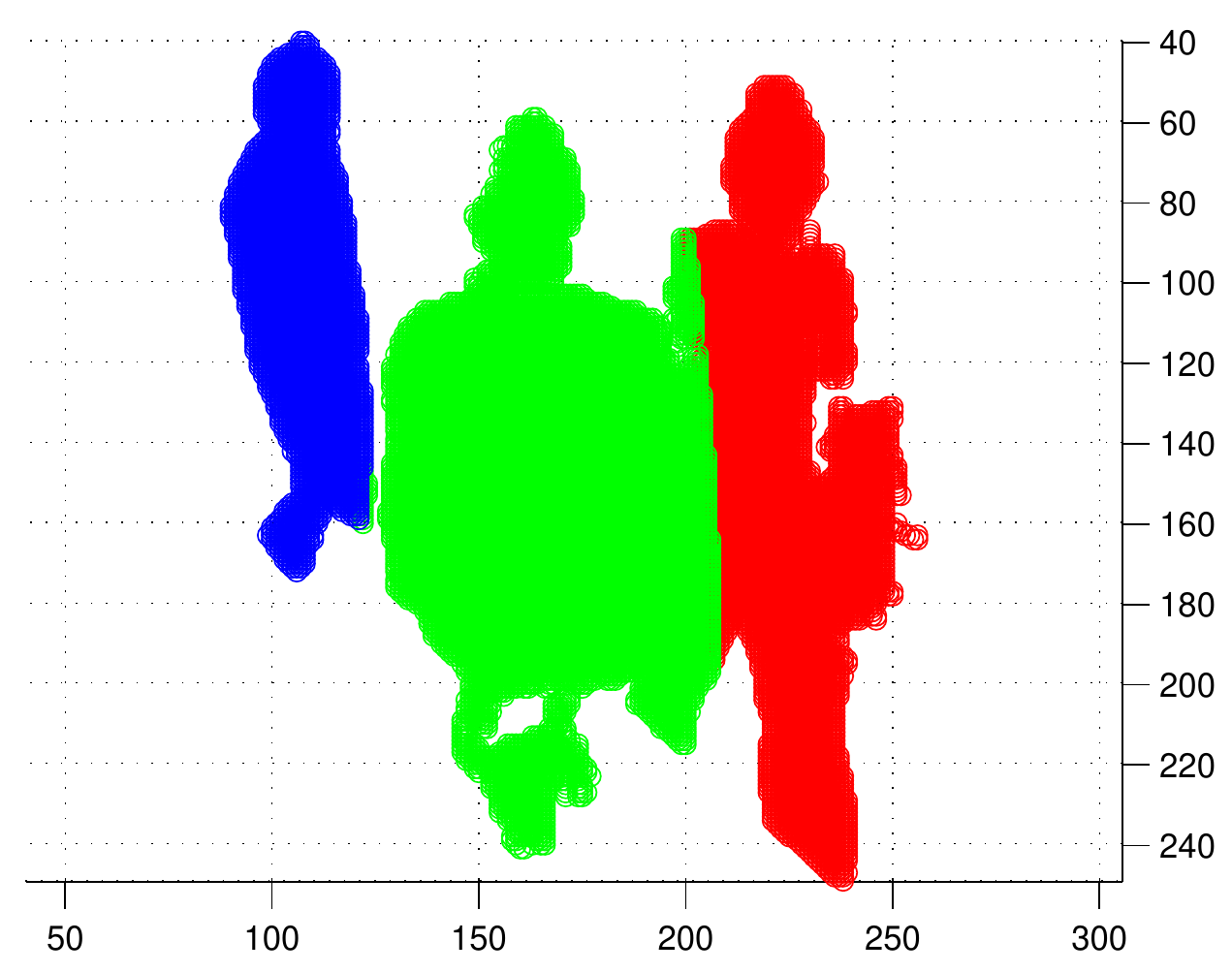} &
\includegraphics[width=0.75in]{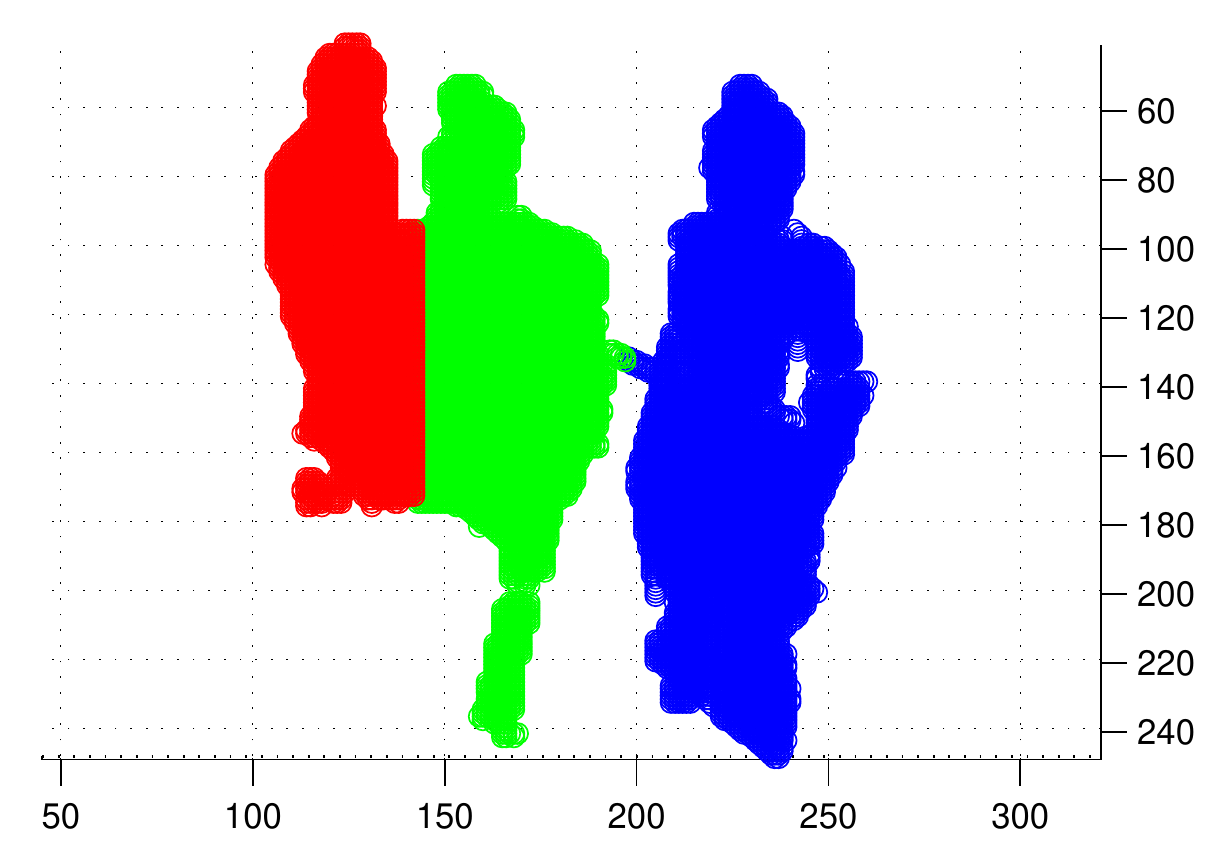}&
\includegraphics[width=0.75in]{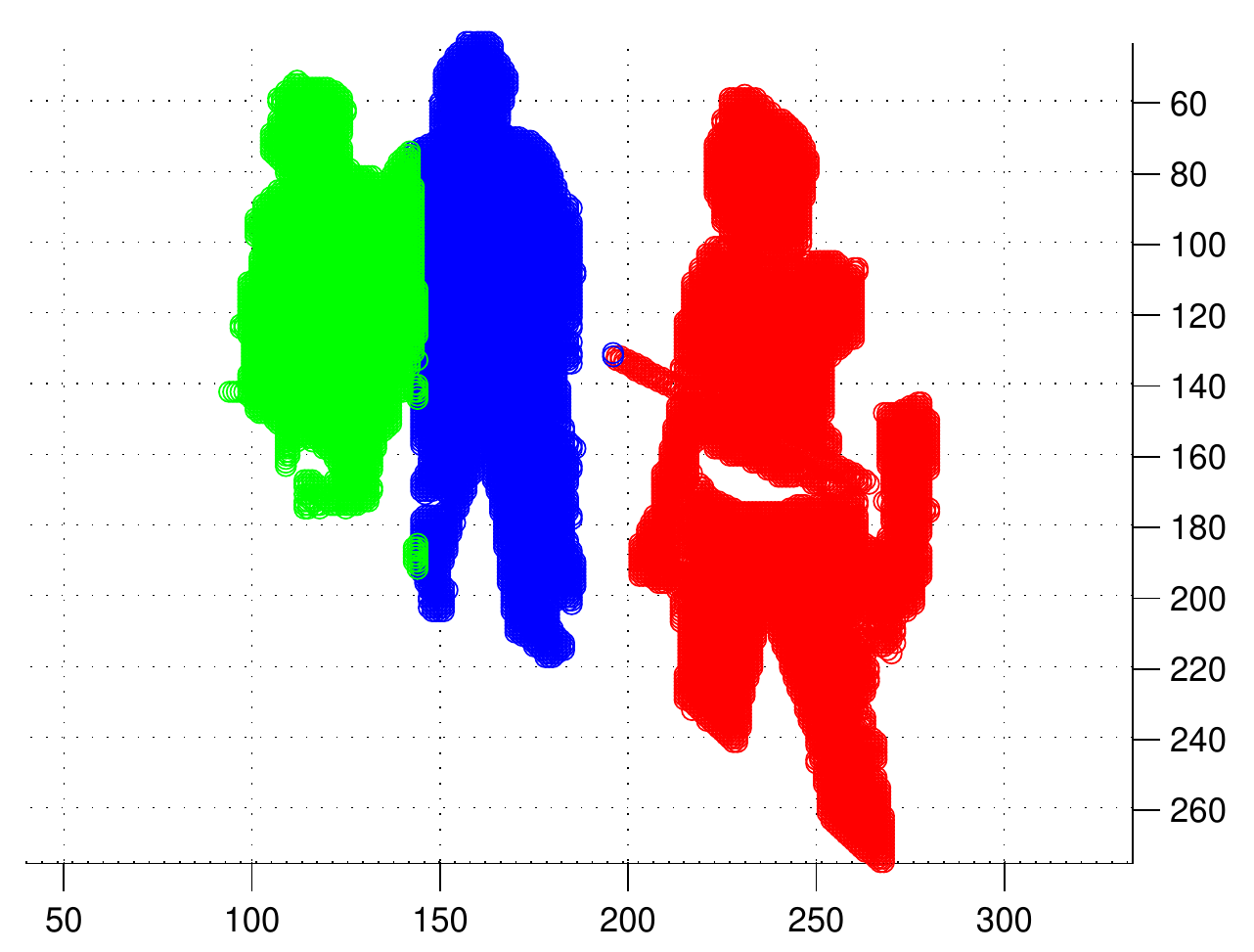}\\
\small{\mbox{(a) Frame 255}} & \small{\mbox{(b) Frame 278}} & \small{\mbox{(c) Frame 288}} & \small{\mbox{(d) Frame 313}}
\end{array}$
\caption{Clustering results for certain frames of sequence ``seq45-3p-1111\_cam3\_divx\_audio'' of the AV16.3 dataset:  First, second and third clusters are represented by blue, red and green colors respectively (a) target $2$ starts occluding target $1$ (b) target $2$ appearing again after occlusion (c) target $2$ is approaching target $3$ (d) target $3$ comes out of occlusion.}
\label{fig3}
\end{figure}

\begin{figure}[h]
\centering
$\begin{array}{c@{\hspace{0.1in}}c@{\hspace{0.1in}}c@{\hspace{0.1in}}c}
\includegraphics[width=0.80in]{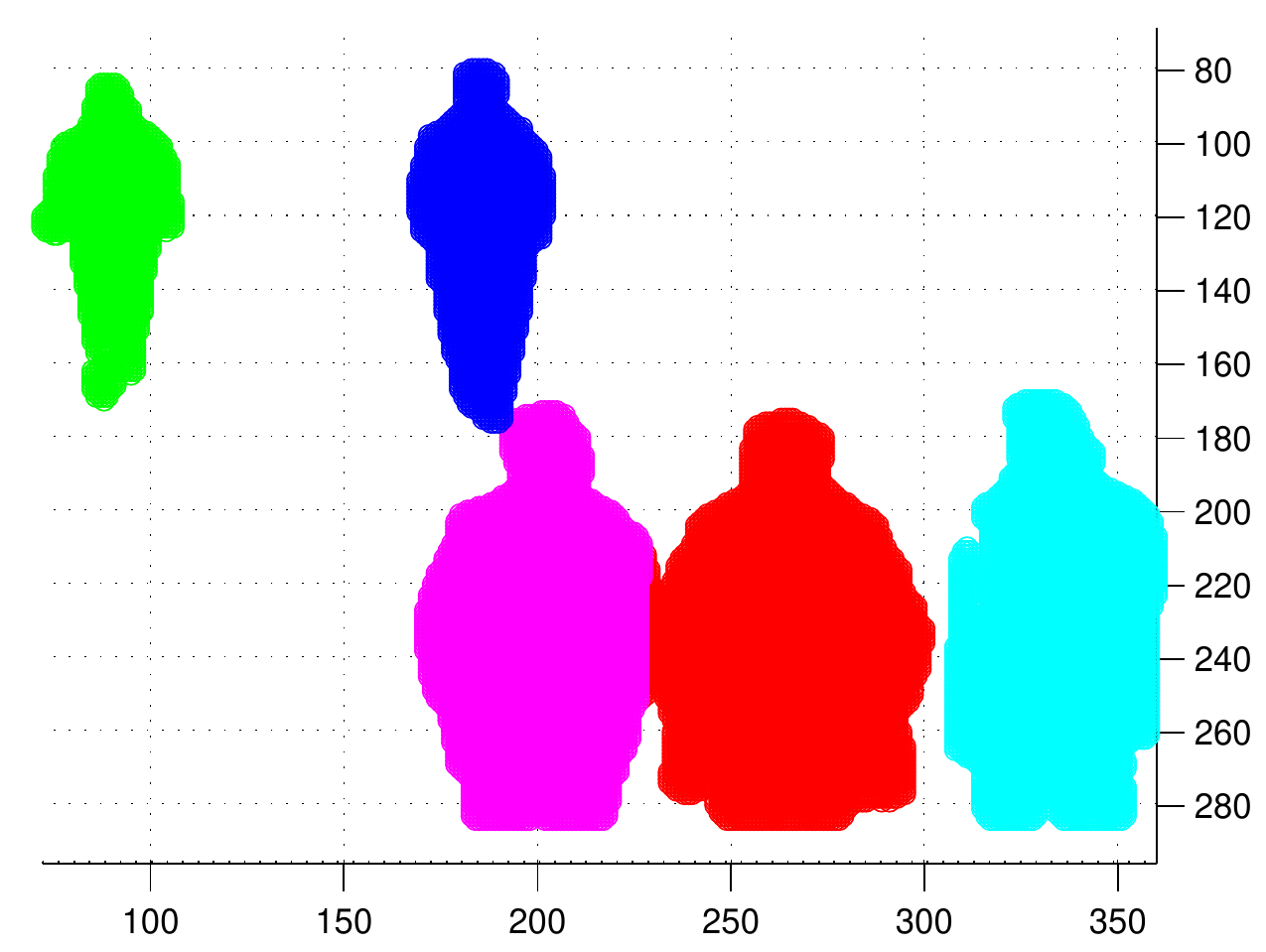} &
\includegraphics[width=0.80in]{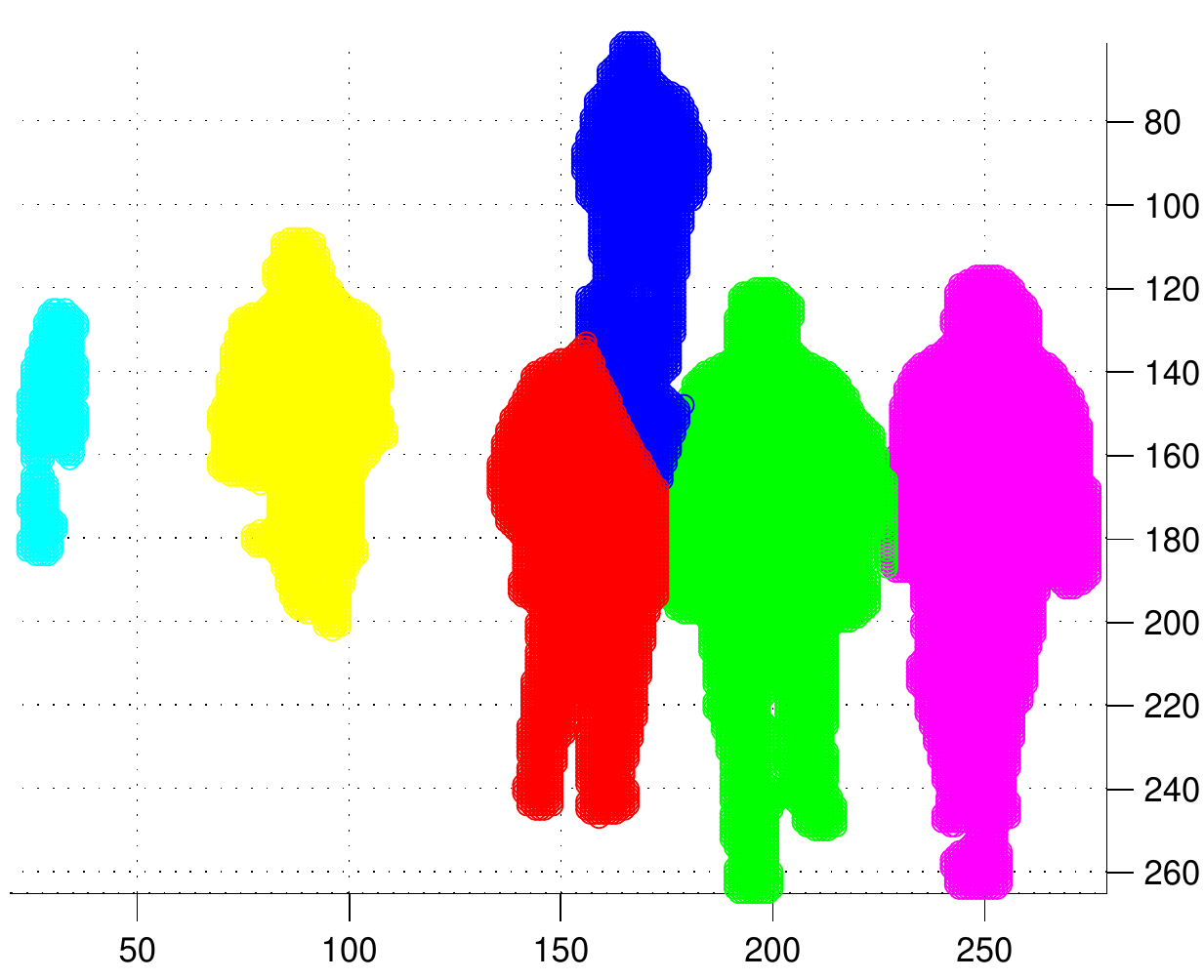}&
\includegraphics[width=0.80in]{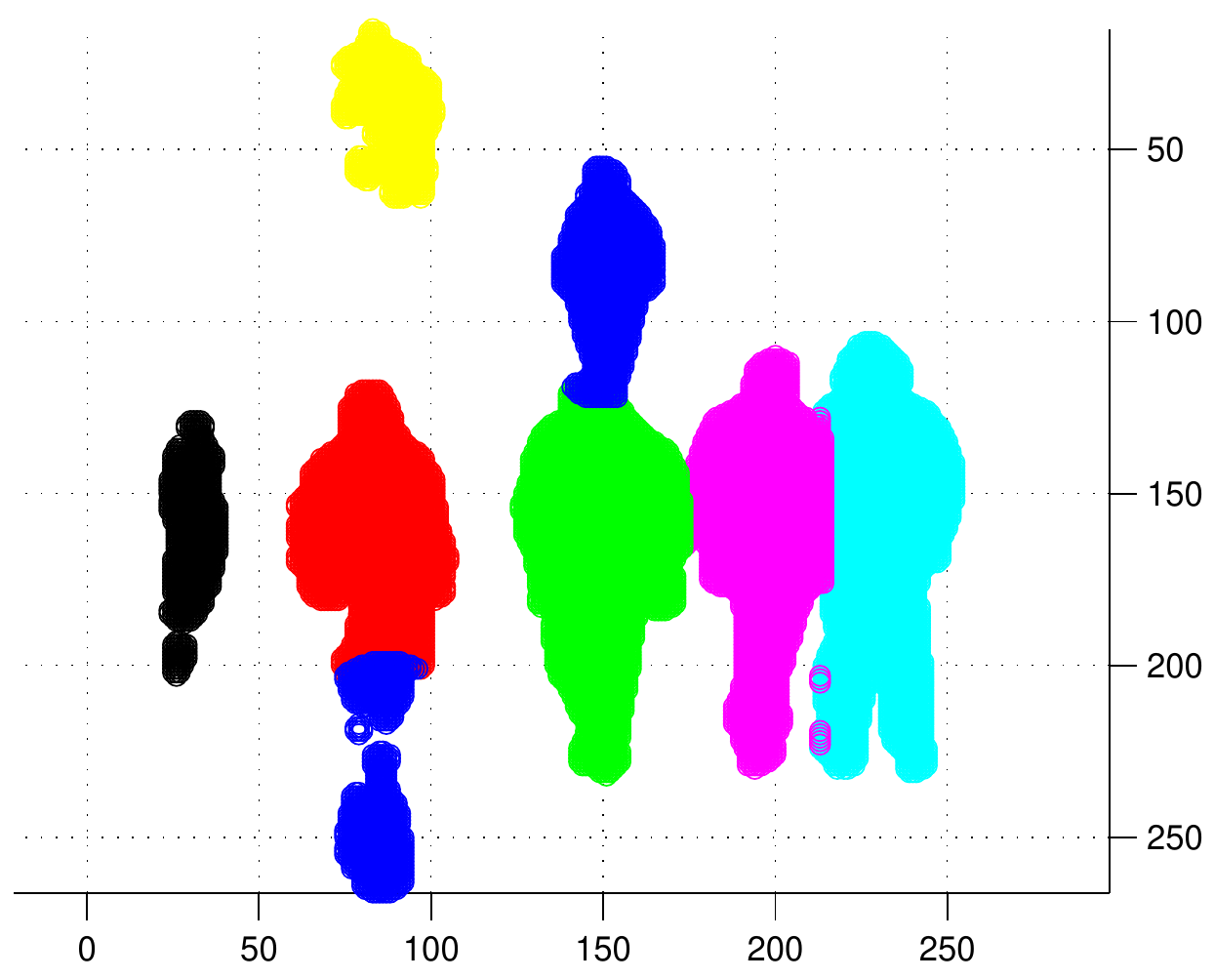}&
\includegraphics[width=0.80in]{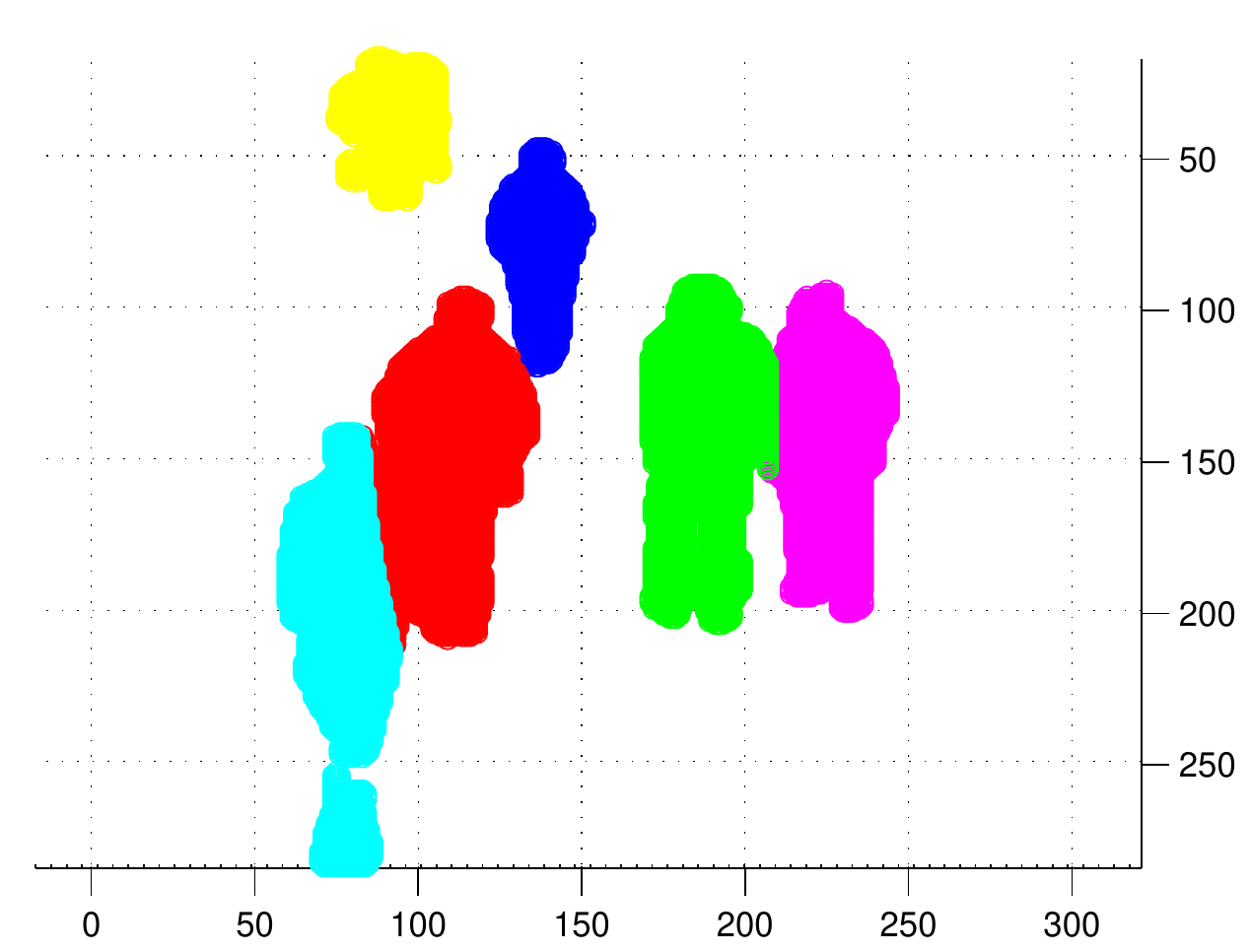}\\
\small{ \mbox{(a) Frame 334}} & \small{\mbox{(b) Frame 440}} & \small{\mbox{(c) Frame 476}} & \small{\mbox{(d) Frame 524}}
\end{array}$
\caption{Clustering results for certain frames of sequence
``ThreePastShop2cor'' of the CAVIAR dataset: clusters 1, 2, 3, 4, 5,
6 and 7 are represented by blue, red, green, magenta, cyan, yellow
and black colors respectively. In frame $334$, we have $8$ clusters
and hence the $8^{th}$ cluster is again represented by blue.}
\label{figcls2}
\end{figure}

\begin{figure}[h]
\centering
$\begin{array}{c@{\hspace{0.1in}}c@{\hspace{0.1in}}c@{\hspace{0.1in}}c}
\includegraphics[width=0.75in]{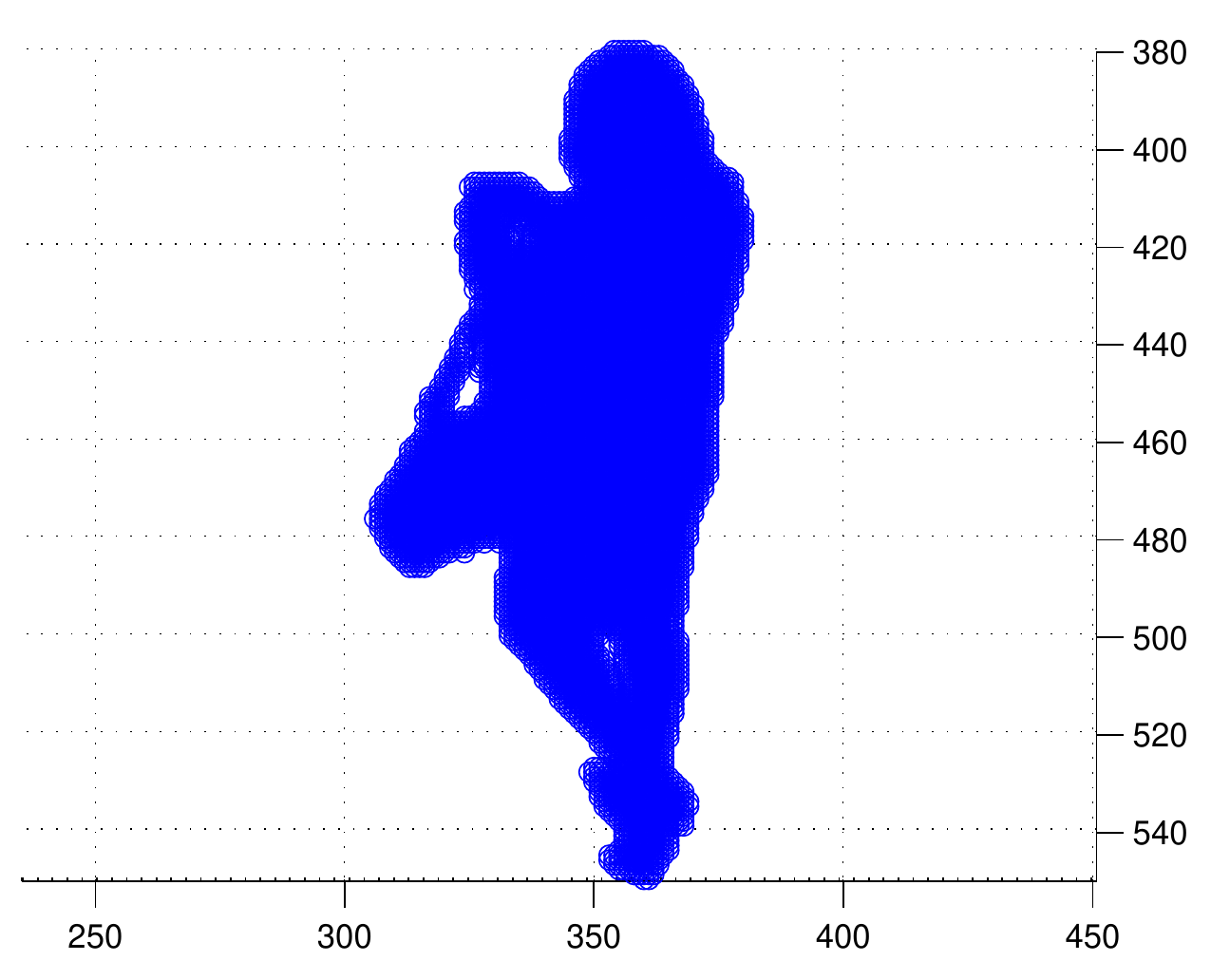} &
\includegraphics[width=0.75in]{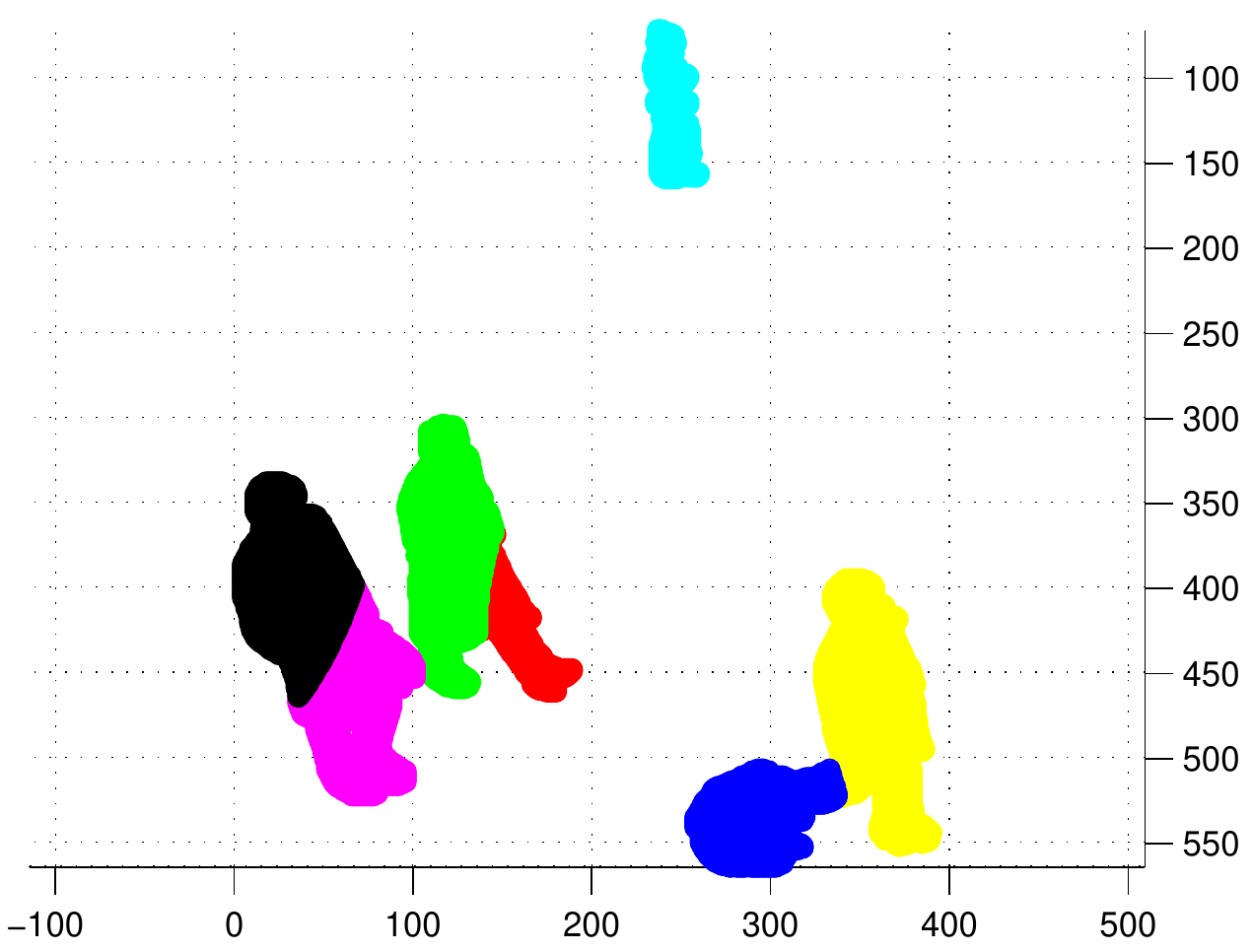} &
\includegraphics[width=0.75in]{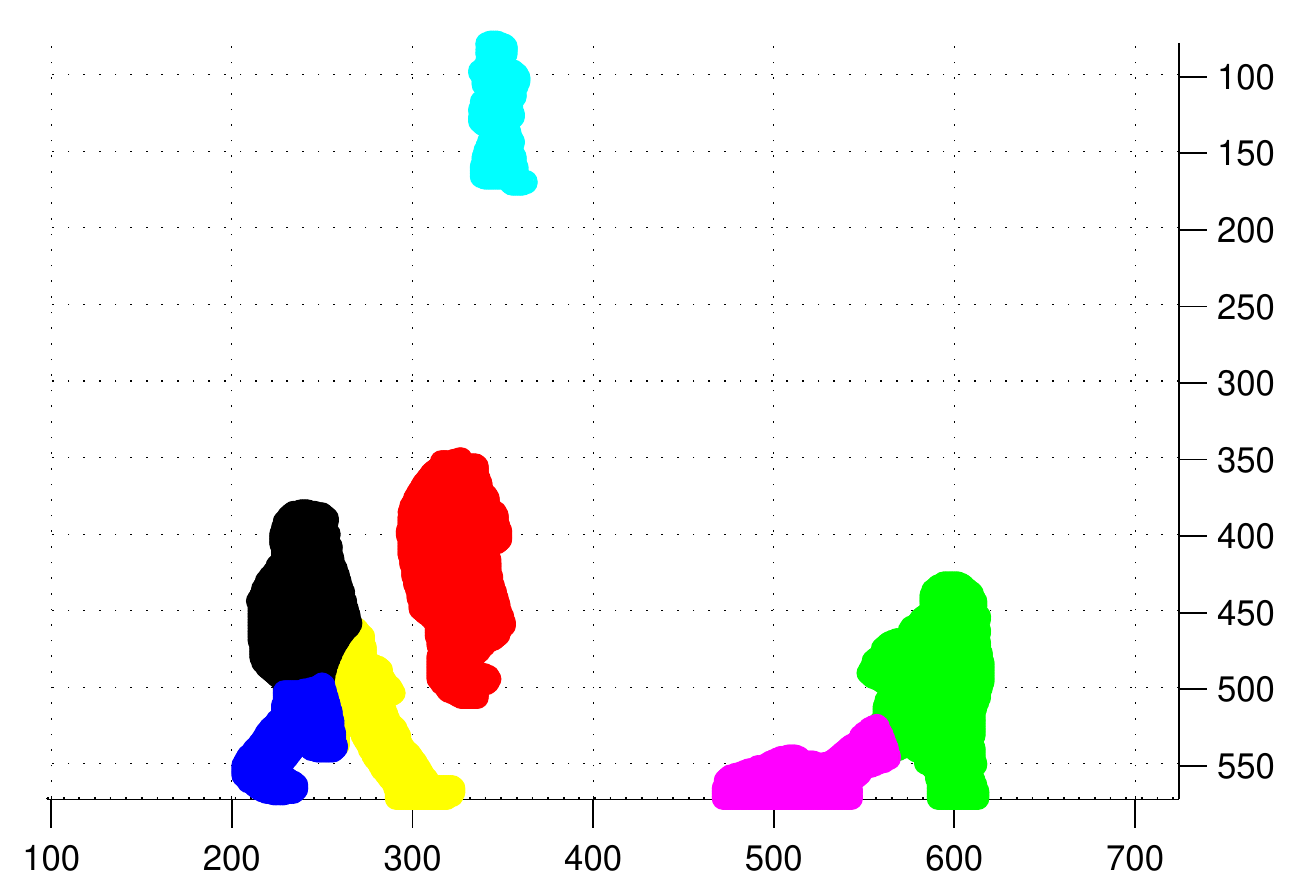}&
\includegraphics[width=0.75in]{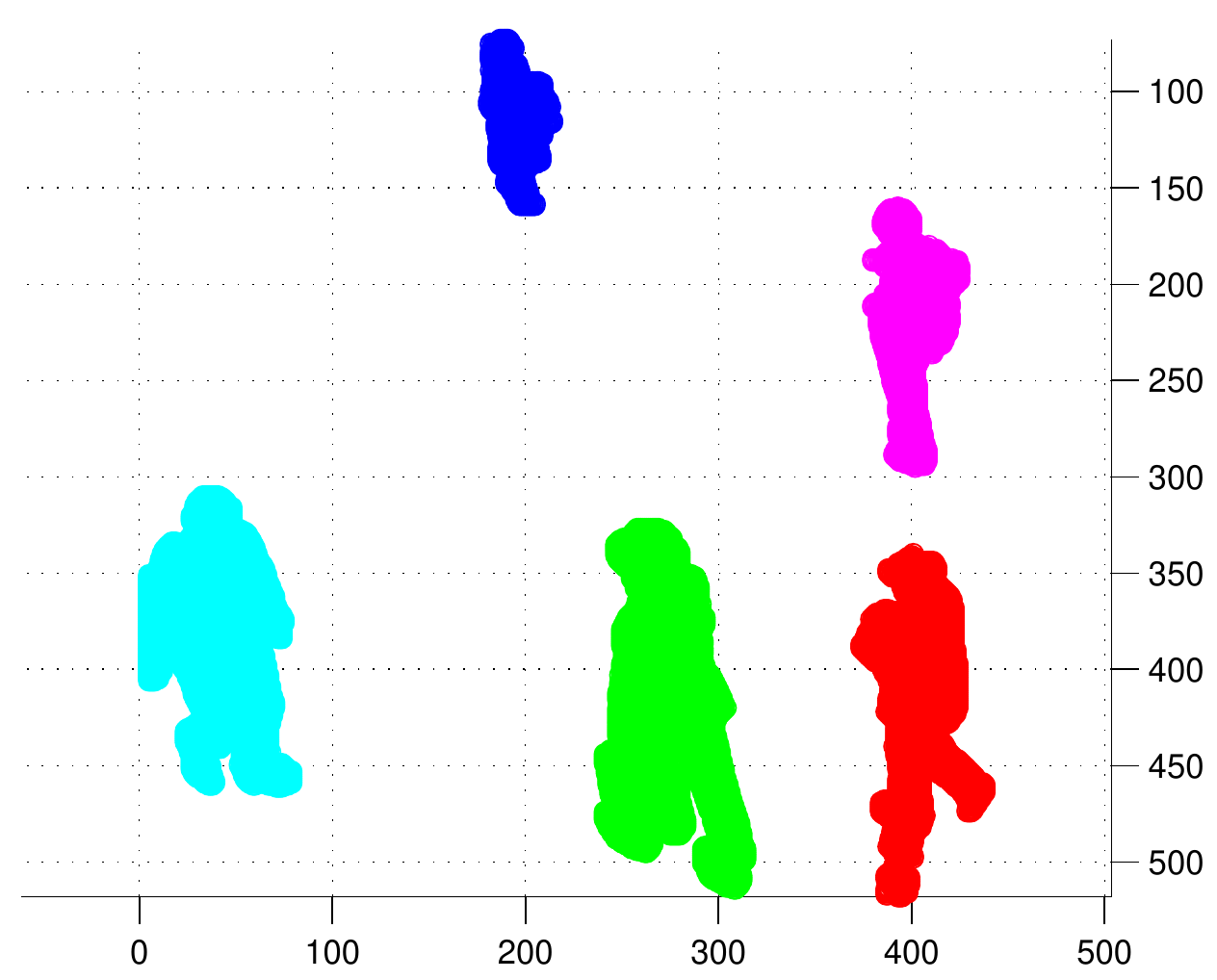}\\
\small{\mbox{(a) Frame 85}} & \small{\mbox{(b) Frame 292}} & \small{\mbox{(c) Frame 327}} & \small{\mbox{(d) Frame 1038}}
\end{array}$
\caption{Clustering results for certain frames of sequence
``S1-T1-C'' of the PETS2006 dataset:  clusters 1, 2, 3, 4, 5, 6 and
7 are represented by blue, red, green, magenta, cyan, yellow and
black colors respectively.} \label{figcls3}
\end{figure}

The prior parameter ${\bf W}_\circ$ used in equation (\ref{eq31}) determines the human shape which is modeled as an ellipse. It is defined with the help of the following equation
\begin{equation}
{\bf W}_\circ = ({\bf U})^T*{[}\begin{smallmatrix} l1&0\\ 0&\l2
\end{smallmatrix}{]}*{\bf U},
\end{equation}
where $l1$ and $l2$ are equatorial radii of the ellipse which models the human shape. The equatorial radii $l1$ and $l2$ are set to $500$ and $300$ respectively while $U$ is defined as
\begin{equation}
{\bf U} = \biggr{[}\begin{smallmatrix} \cos(\pi/2)&-\sin(\pi/2)\\ -\sin(\pi/2)&\cos(\pi/2) \end{smallmatrix}\biggr{]}.
\end{equation}
The clustering results for a few of the video frames are shown in Figs. \ref{fig3}, \ref{figcls2} and \ref{figcls3}. Blue, red, green, magenta, cyan, yellow and black represent first, second, third, fourth, fifth, sixth and seventh clusters respectively. If there are more than $7$ clusters we repeat the color scheme.

These figures show that the clustering performs well and clusters do not contain regions of multiple targets even when targets are in close interaction or partially occlude each other.
To eliminate the extra foreground pixels due to the reflections, the
small clusters consisting of less than $100$ pixels are deleted.
\subsection{Data Association and Occlusion Handling}
For data association the ten best hypotheses of joint association
event $\boldsymbol{\psi}_k$ are considered which are obtained with
the help of Murty's algorithm \cite{murty's}. The Bhattacharyya
distance
between the color histograms of cluster
$q$ and target $i$ is used to calculate the distance $d^i(H_{ref}^i,H^q_{k})$
\begin{equation}\label{eq42}
d^i(H_{ref}^i,H^q_{k}) = \sqrt{1 - \rho(H^i_{ref},H^q_{k})},
\end{equation}
where $H^i_{ref}$ is the reference histogram which is created by
using the cluster associated with target $i$ at the time step when
the target $i$ first appears in the video. $H^q_{k}$ is the
histogram created for cluster $q$ at the current time step $k$ and
$\rho(H^i_{ref},H^q_{k})$ is the Bhattacharyya coefficient
\begin{equation}\label{eq43}
\rho(H^i_{ref},H^q_{k}) = \sum_{g=1}^{G}\sqrt{H^{i,g}_{ref},H^{q,g}_{k}},
\end{equation}
where $G$ represents the number of histogram bins and we have used  $16 \times 16 \times 16$ color histograms bins.

A tracking algorithm with only a JPDAF, without the variational
Bayesian clustering and without a social force model fails to
identify an object during close interactions. In the proposed
algorithm these tracking failures are overcome with the help of the
proposed data association technique. Sometimes, due to the varying
shape of the targets, the clustering stage may produce more than one
cluster per target. Therefore, the proposed data association
technique assigns multiple clusters to every target with some
association probability.
\subsection{Variable Number of Targets Results}
\begin{figure}[!htb]
\centering
\includegraphics[width=3.2in]{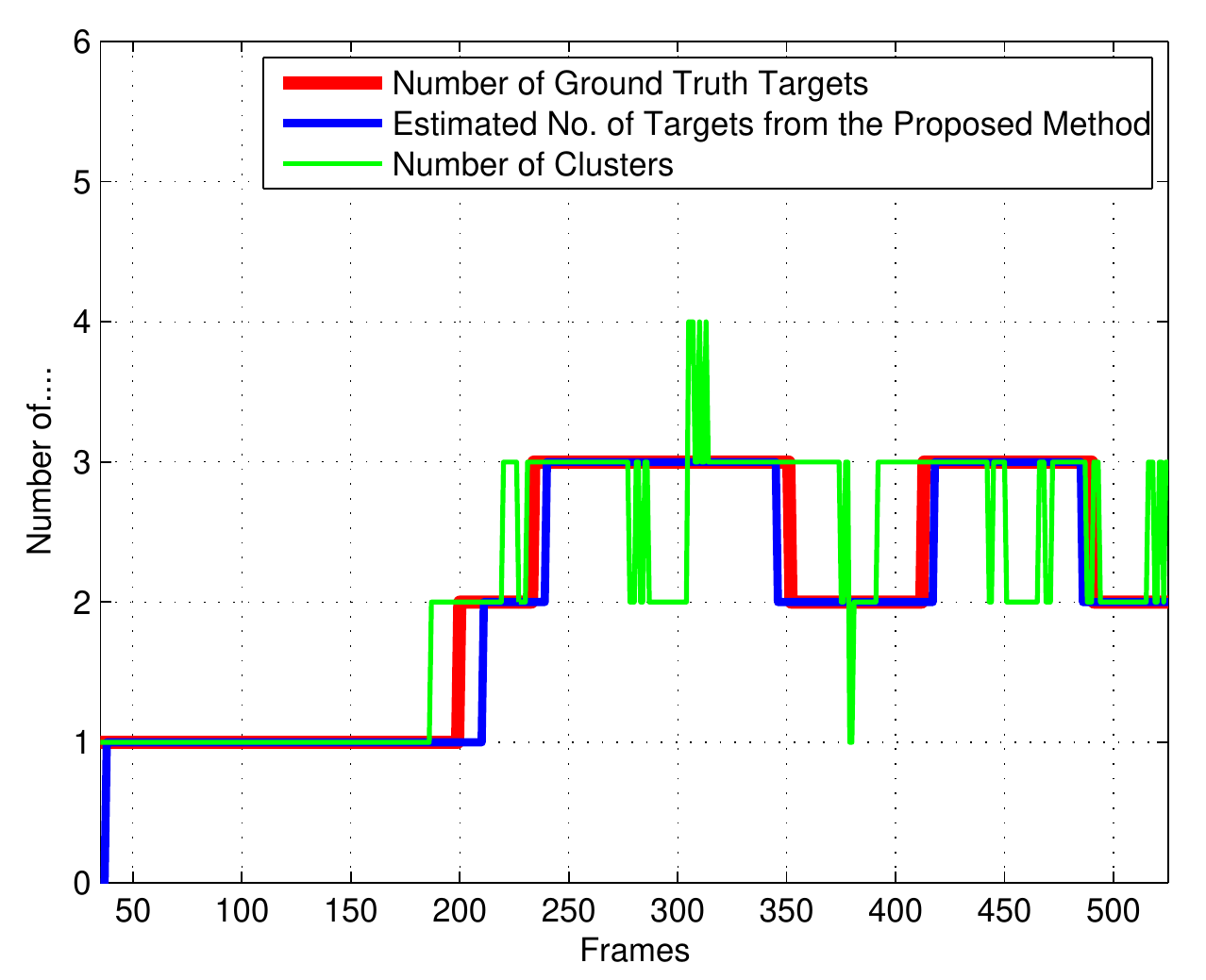}
\caption{The graph shows the actual number of targets, estimated number of targets by using the proposed method and the number of clusters as a function of video frames for sequence ``seq45-3p-1111\_cam3\_divx\_audio'' of the AV16.3 dataset.}\label{var-tar-av}
\end{figure}

\begin{figure}[!htb]
\centering
\includegraphics[width=3.2in]{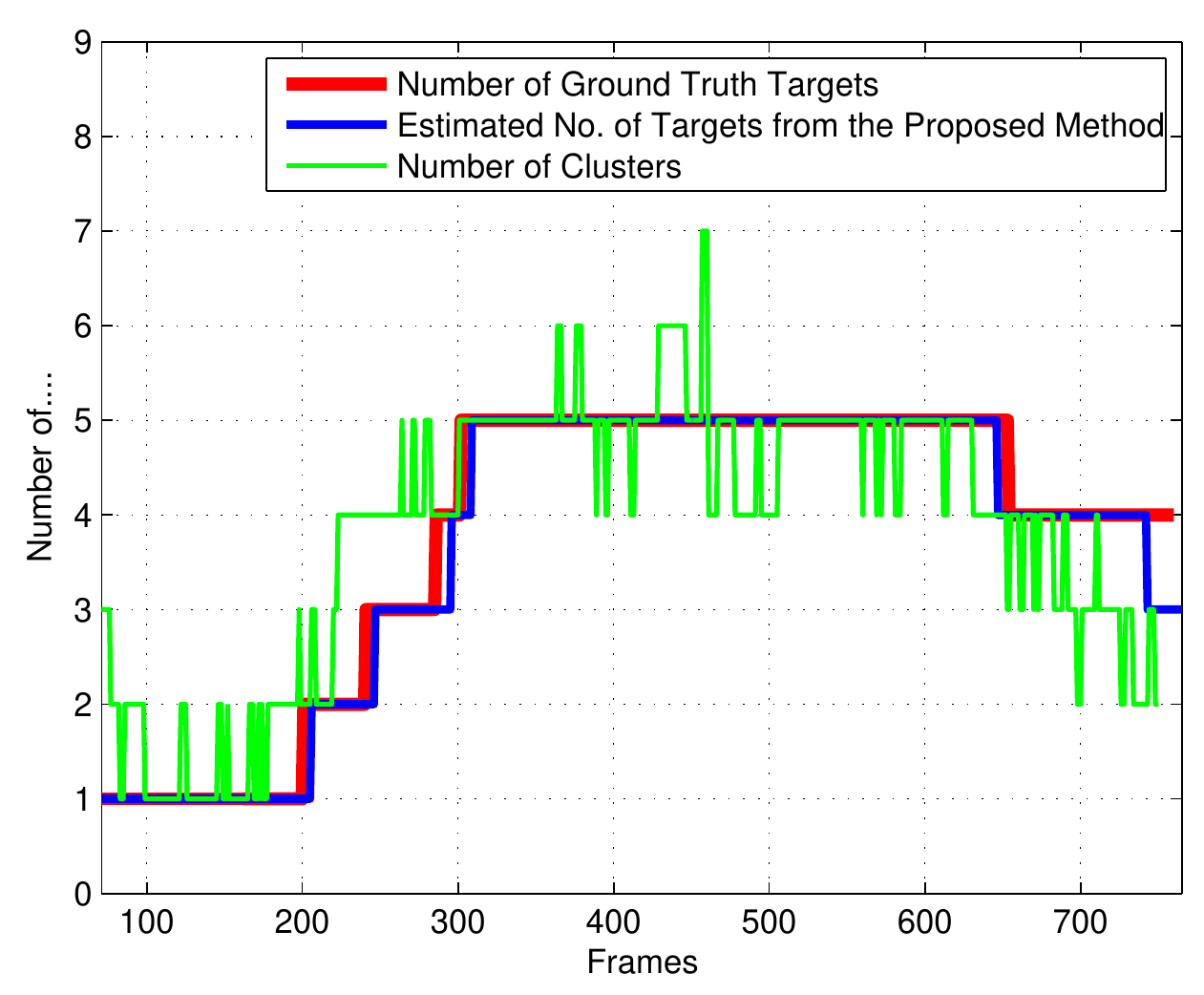}
\caption{The graph shows the actual number of targets, estimated number of targets by using the proposed method and the number of clusters as a function of video frames for sequence ``ThreePastShop2cor'' of the CAVIAR dataset.}\label{var-tar-cav}
\end{figure}

The robustness of the algorithm for estimating the correct number of targets in a video sequence is compared with the framework developed in \cite{tinne}. In \cite{tinne} the number of targets is calculated on the basis of only the number of clusters. Figs.~ \ref{var-tar-av} and \ref{var-tar-cav} present a comparison of number of targets and number of clusters in video frames.

%

%


%


It is apparent from Figs. \ref{var-tar-av} and \ref{var-tar-cav} that the number of targets does
not always match the number of clusters and hence it is
difficult to train the system to estimate the number of targets on
the basis of the number of clusters. In the proposed algorithm,
instead of using the number of clusters we have exploited the size
and location of clusters to estimate the number of targets on all
the sequences. This leads to accurate estimation of the number of
targets and is demonstrated on Figs.~\ref{var-tar-av} and
\ref{var-tar-cav} for two different data sets and in Figs.
\ref{fig6}, \ref{figorg2} and \ref{figorg3}.
\subsection{Tracking Results}
\label{trackn-res}
A minimum mean square error (MMSE) particle filter is used to
estimate the states of the multiple targets. When a new target
enters the room, the algorithm automatically initializes it with the
help of the data association results by using the mean value of the
cluster assigned to that target. Similarly, the algorithm removes
the target which leaves the room. The particle size $N_s$ is chosen
to be equal to $60$ and the number $S$ of social links are updated
at every time step. Table \ref{par} shows the values of the other
parameters used in the social force model. The prediction step is
performed by the social force model, described in Section~\ref{dmm}.
\begin{table}
\centering
\caption{Parameters of social force model}\label{par}
 {
\begin{tabular}{|c|c|c|}
\hline

Symbol & Definition & Value \\
\hline
  $b$  &   Boundary of social force in metres  &  3m     \\
\cline{1-3}
$r_i$ &   Radius of target's influence in metres   &  0.2m  \\
\cline{1-3}
$m$ &   Mass of target  &  80kg  \\
\cline{1-3}
$f_a$ &   Magnitude of attractive force  &  500N  \\
\cline{1-3}
$f_r$ &   Magnitude of repulsive force  &  500N  \\
\cline{1-3}
$\Delta t$ &   Time interval  &  ${1}/{25}$  \\
\hline
\end{tabular}
}
\end{table}

\begin{figure*}
\centering
$\begin{array}{c@{\hspace{0.35in}}c@{\hspace{0.35in}}c@{\hspace{0.35in}}c@{\hspace{0.35in}}c}
\includegraphics[width=1.0in]{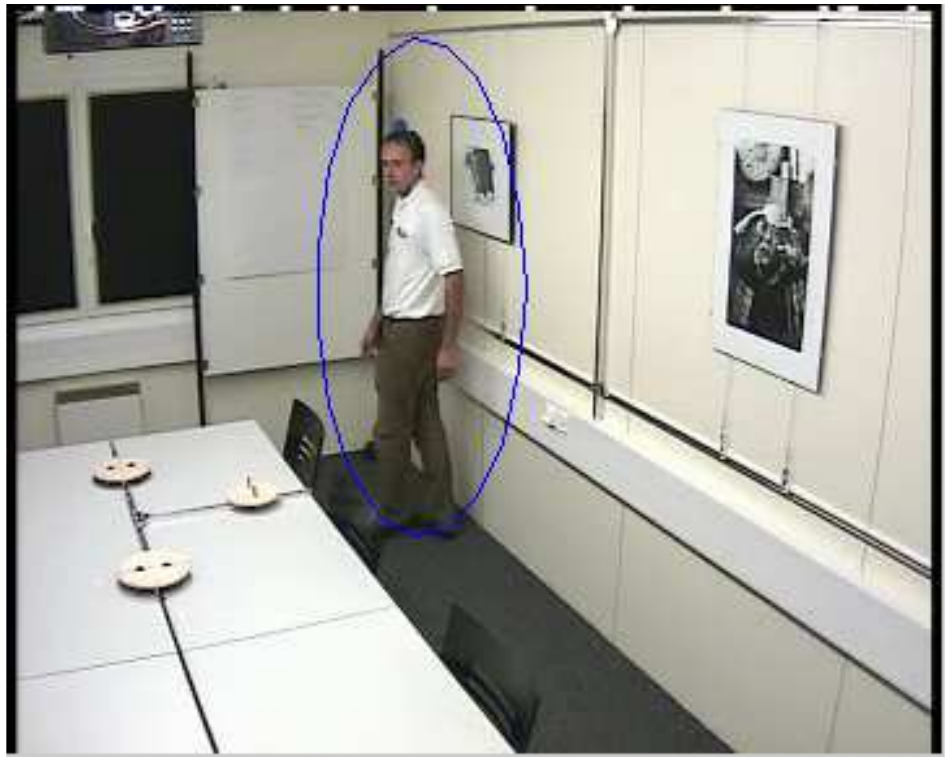} &
\includegraphics[width=1.0in]{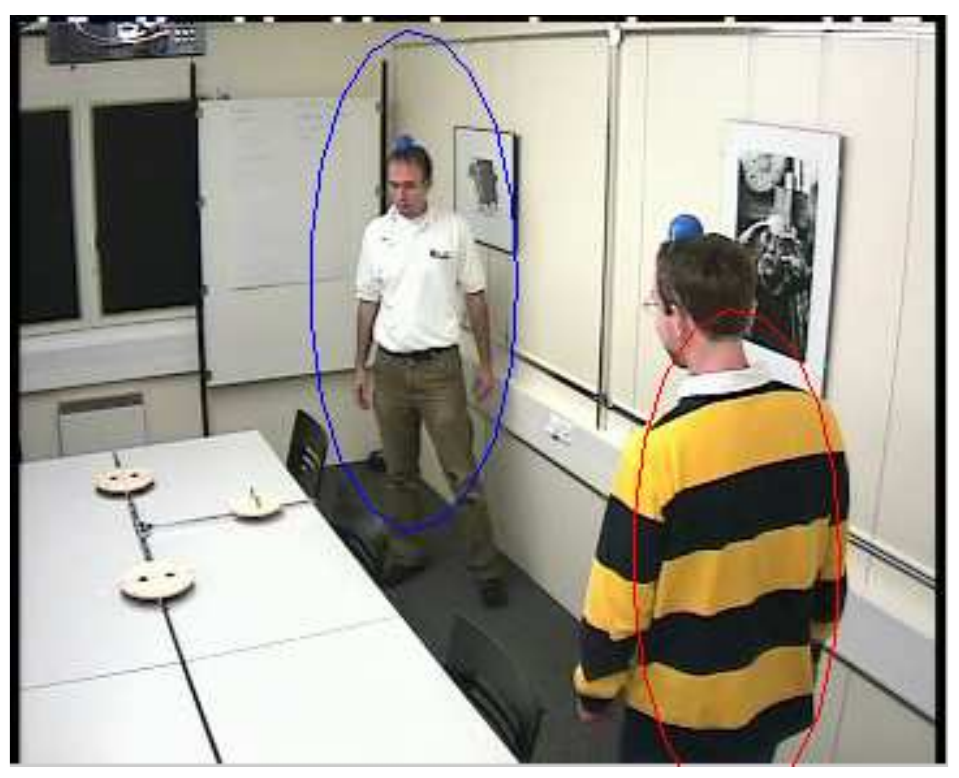} &
\includegraphics[width=1.0in]{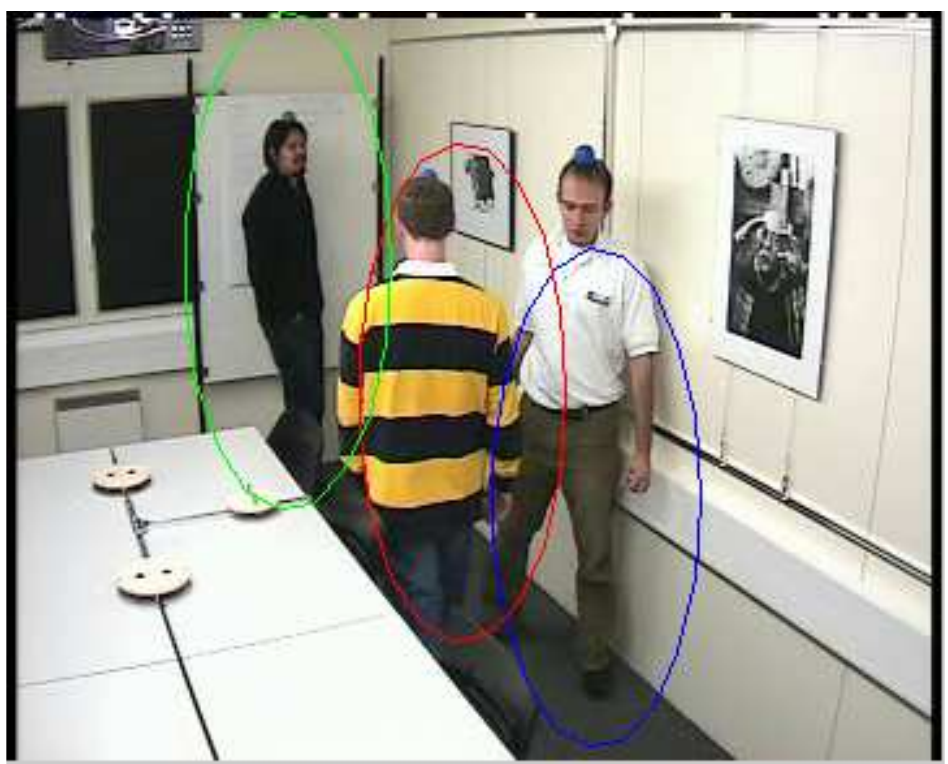}&
\includegraphics[width=1.0in]{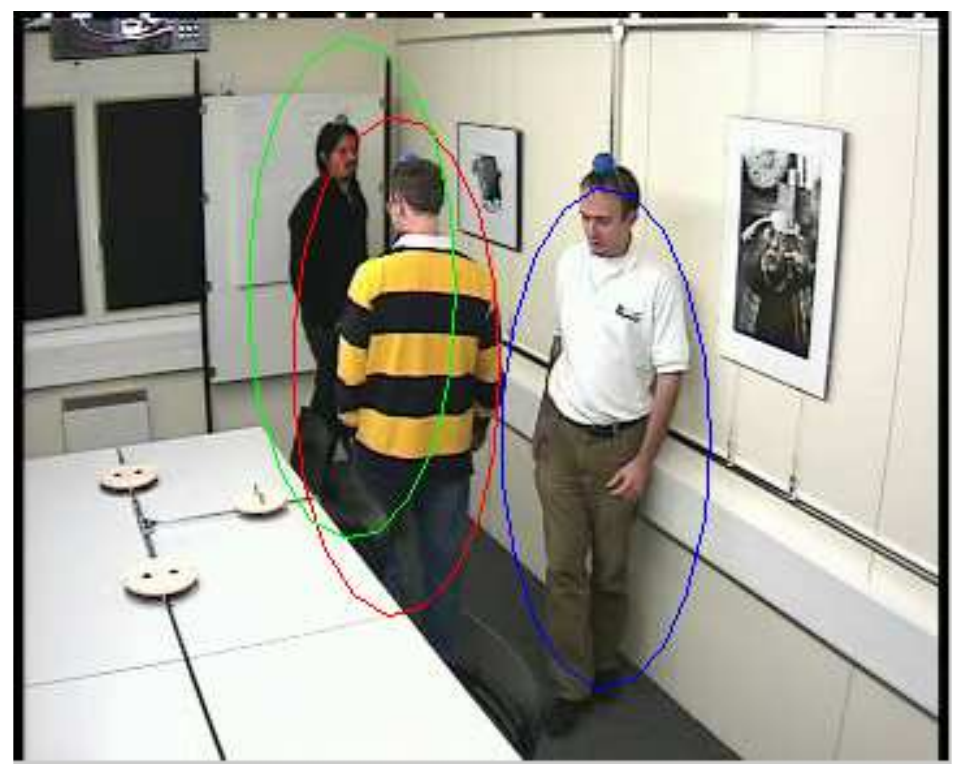}&
\includegraphics[width=1.0in]{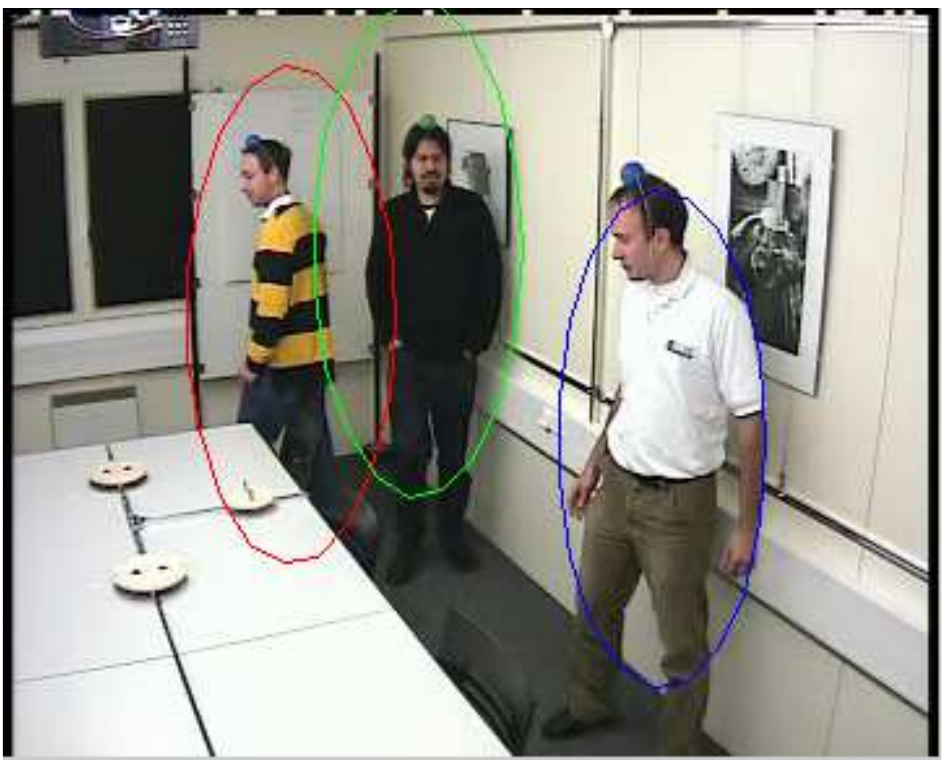}\\
\mbox{(a) Frame 124} & \mbox{(b) Frame 225} & \mbox{(c) Frame 278} & \mbox{(d) Frame 288} & \mbox{(e) Frame 320}\\[0.3cm]
\end{array}$
$\begin{array}{c@{\hspace{0.35in}}c@{\hspace{0.35in}}c@{\hspace{0.35in}}c@{\hspace{0.35in}}c}
\includegraphics[width=1.0in]{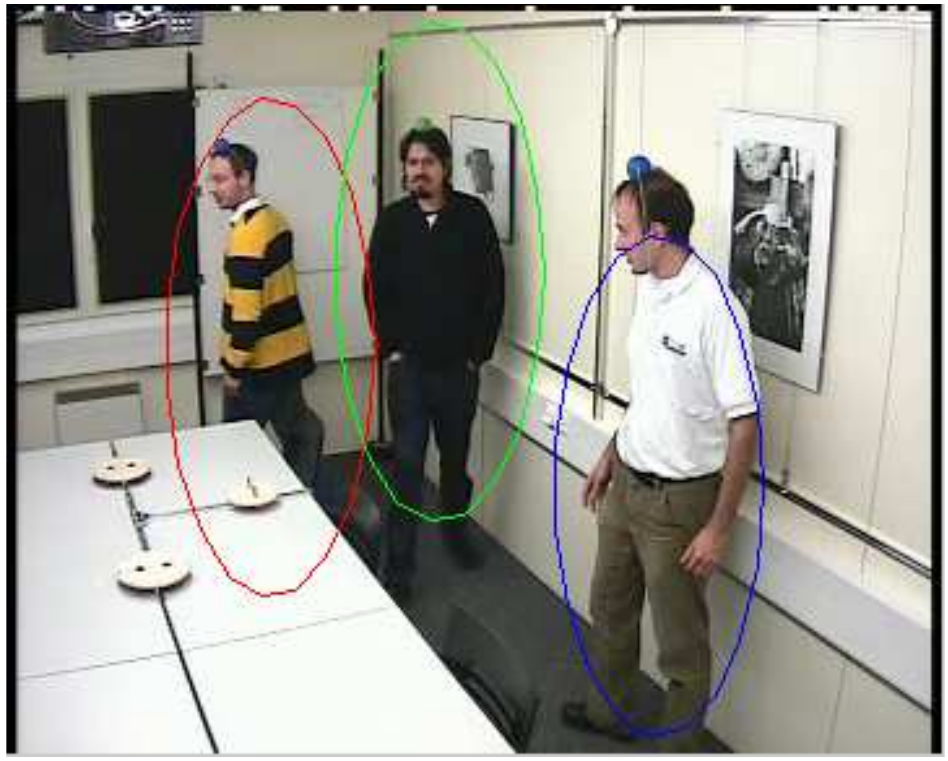}&
\includegraphics[width=1.0in]{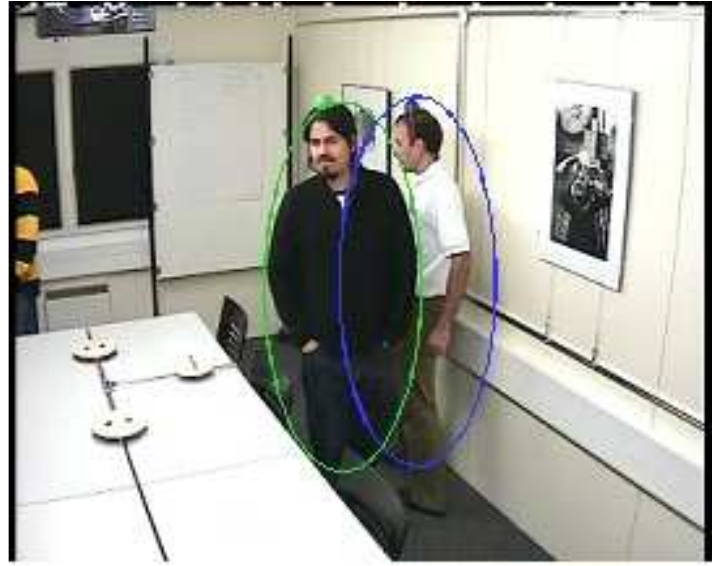}&
\includegraphics[width=1.0in]{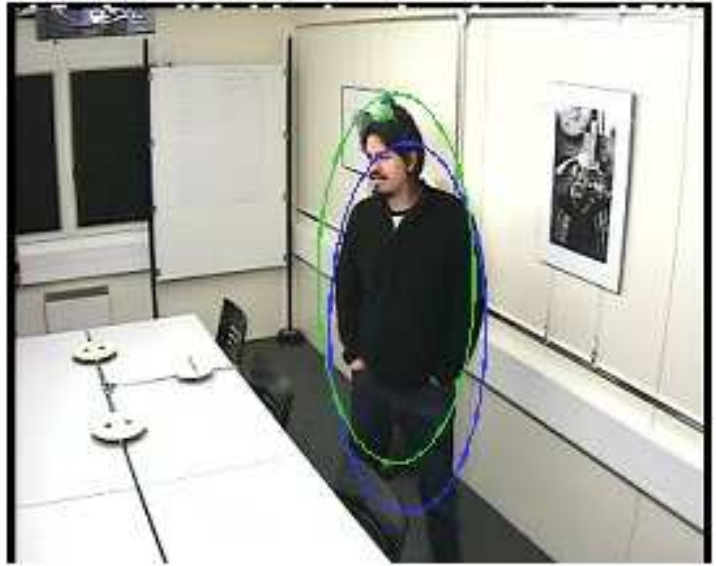}&
\includegraphics[width=1.0in]{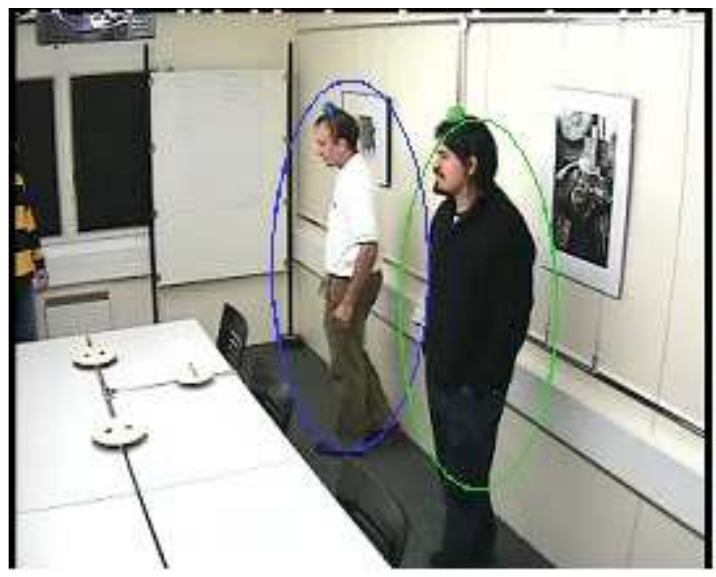}&
\includegraphics[width=1.0in]{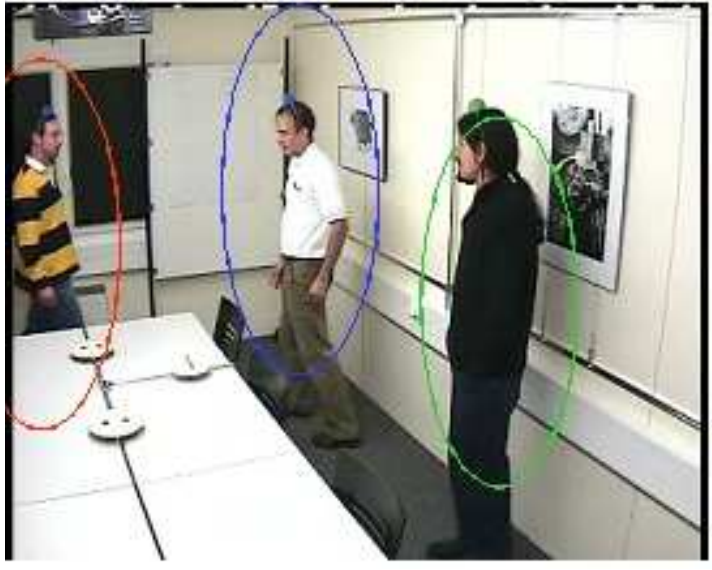}\\
\mbox{(f) Frame 326} & \mbox{(g) Frame 375} & \mbox{(h) Frame 389} & \mbox{(i) Frame 405} & \mbox{(j) Frame 420}
\end{array}$

\caption{Tracking results for certain frames of sequence
``seq45-3p-1111\_cam3\_divx\_audio'' of the AV16.3 dataset with 
a variable number of targets while handling complex occlusions. (a)
tracking of one target, (b), (c) and (j) show that the algorithm
successfully initializes the new tracker, (g) shows that the tracker
is deleted when the target leaves the room, (c)-(i) show successful
occlusion handling.} \label{fig6}
\end{figure*}

\begin{figure*}
\centering
$\begin{array}{c@{\hspace{0.35in}}c@{\hspace{0.35in}}c@{\hspace{0.35in}}c@{\hspace{0.35in}}c}
\includegraphics[width=1.0in]{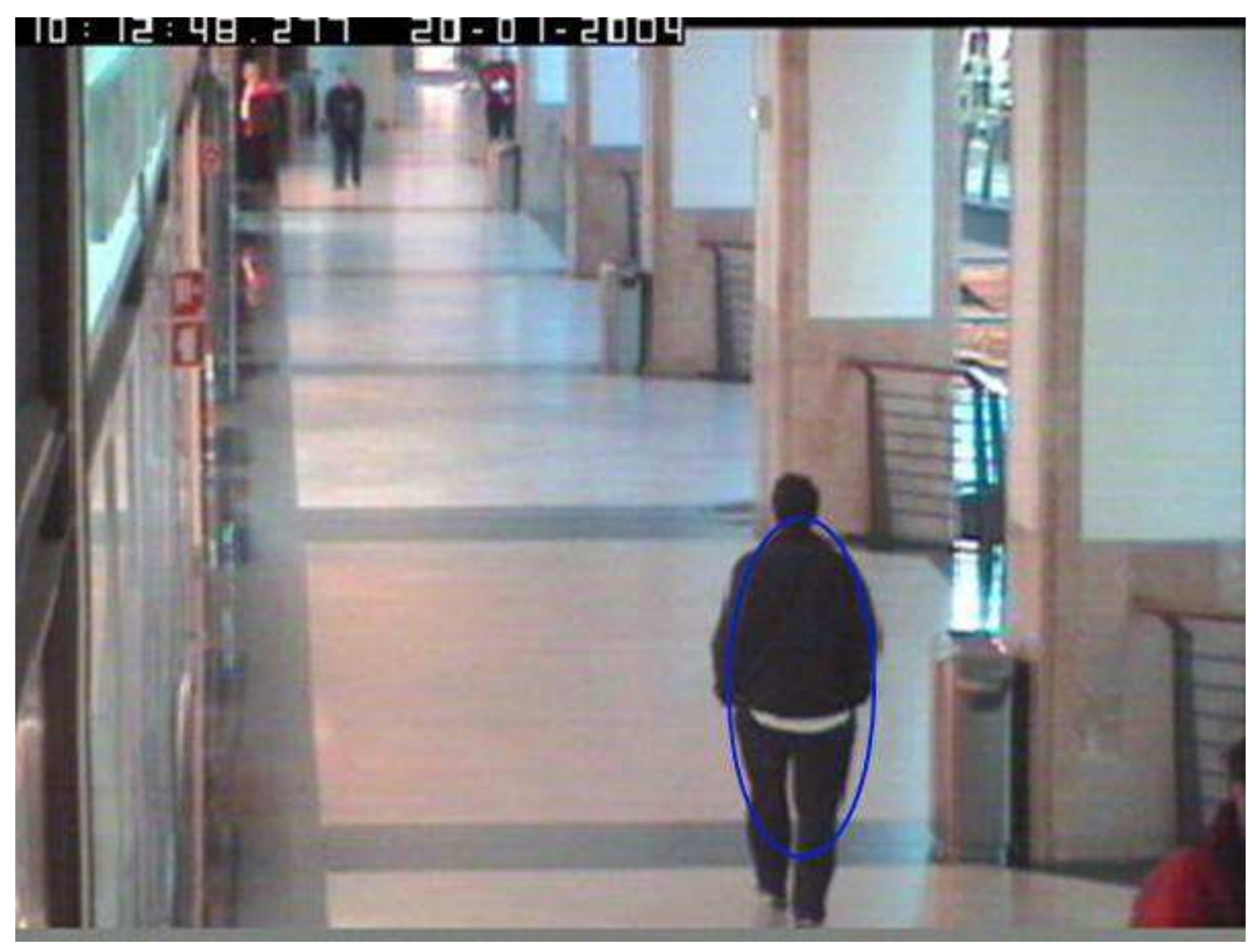} &
\includegraphics[width=1.0in]{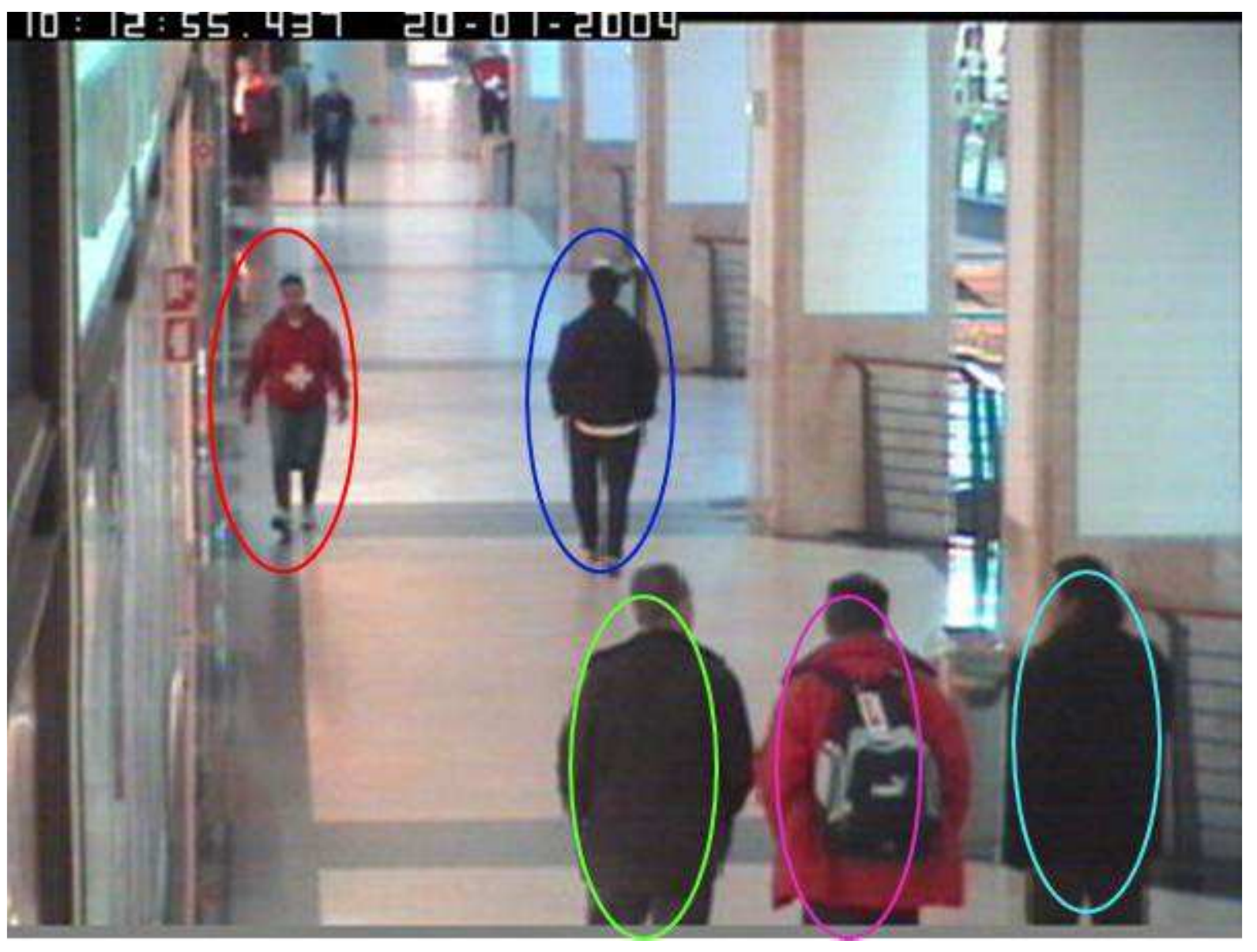} &
\includegraphics[width=1.0in]{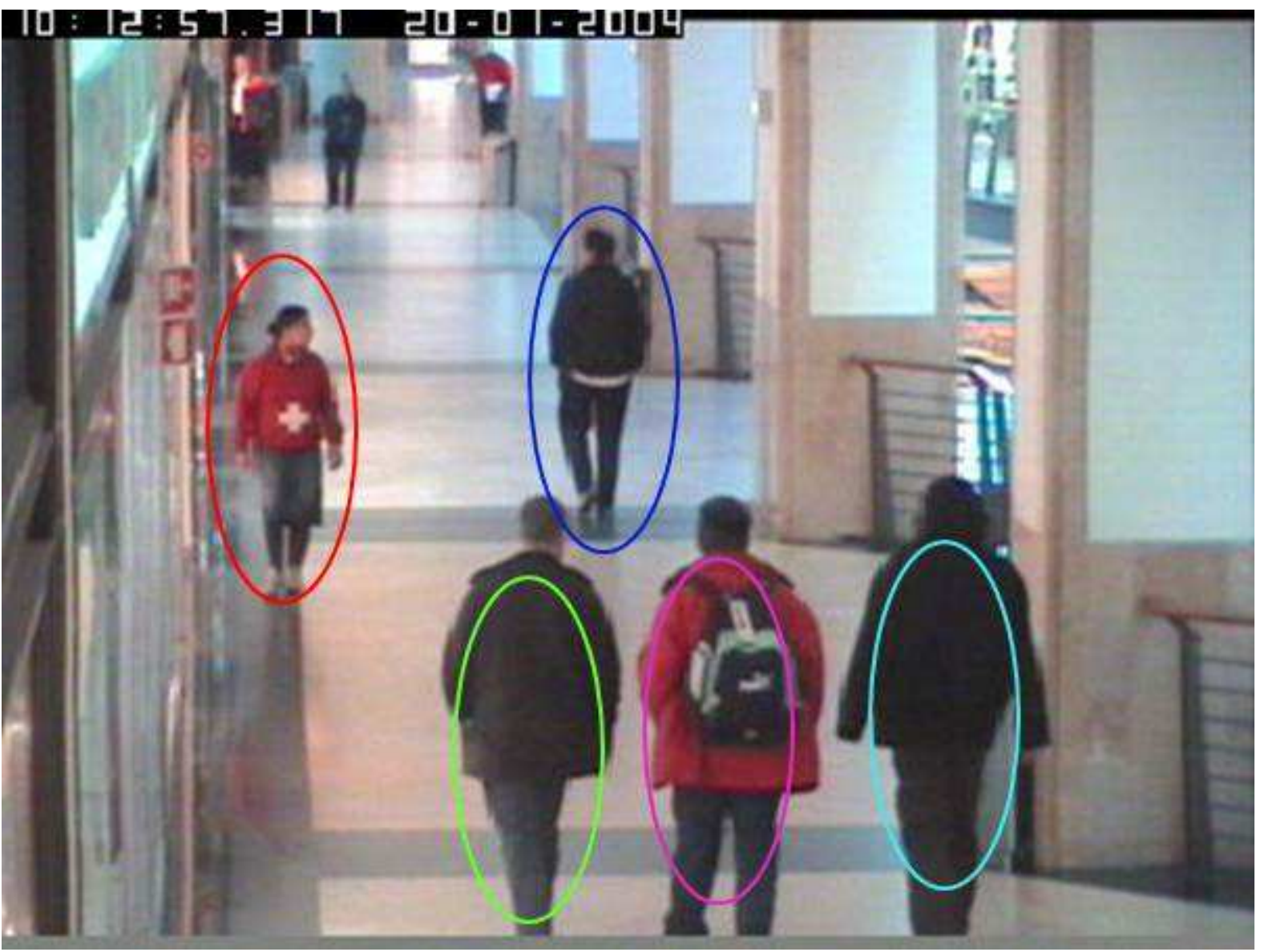}&
\includegraphics[width=1.0in]{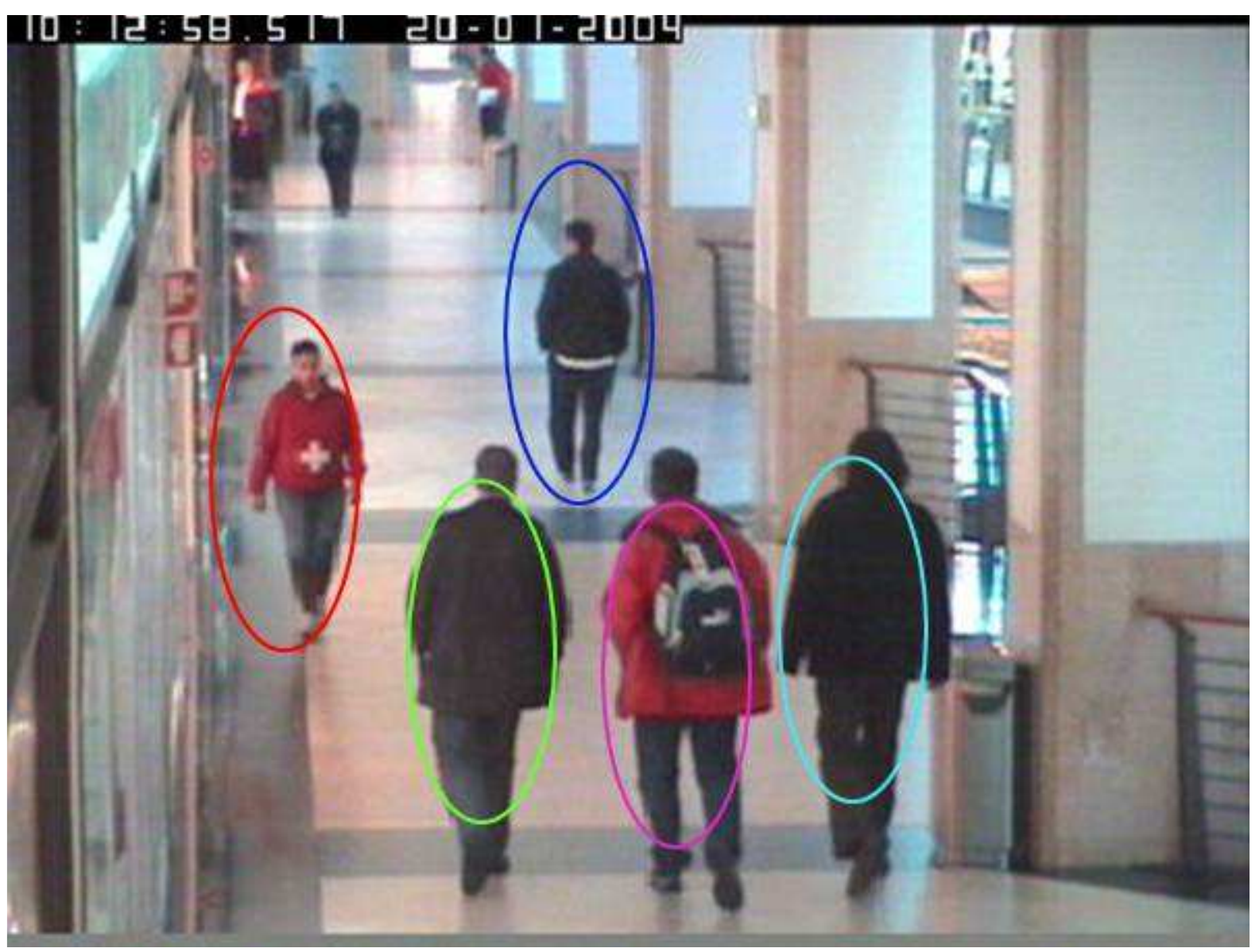}&
\includegraphics[width=1.0in]{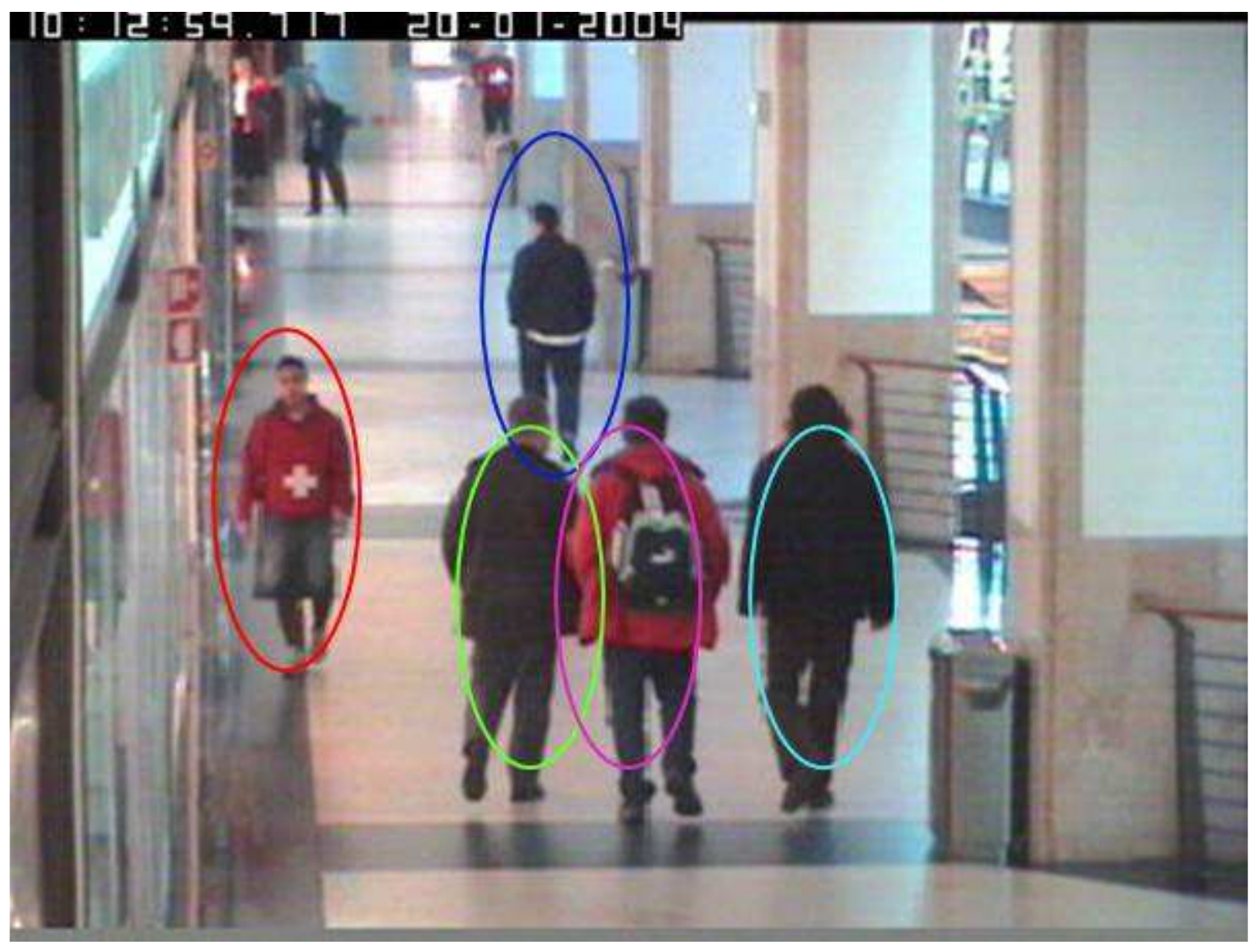}\\

\mbox{(a) Frame 152} & \mbox{(b) Frame 334}&\mbox{(c) Frame 379} & \mbox{(d) Frame 409}&\mbox{(e) Frame 440}
\end{array}$

$\begin{array}{c@{\hspace{0.35in}}c@{\hspace{0.35in}}c@{\hspace{0.35in}}c@{\hspace{0.35in}}c}
\includegraphics[width=1.0in]{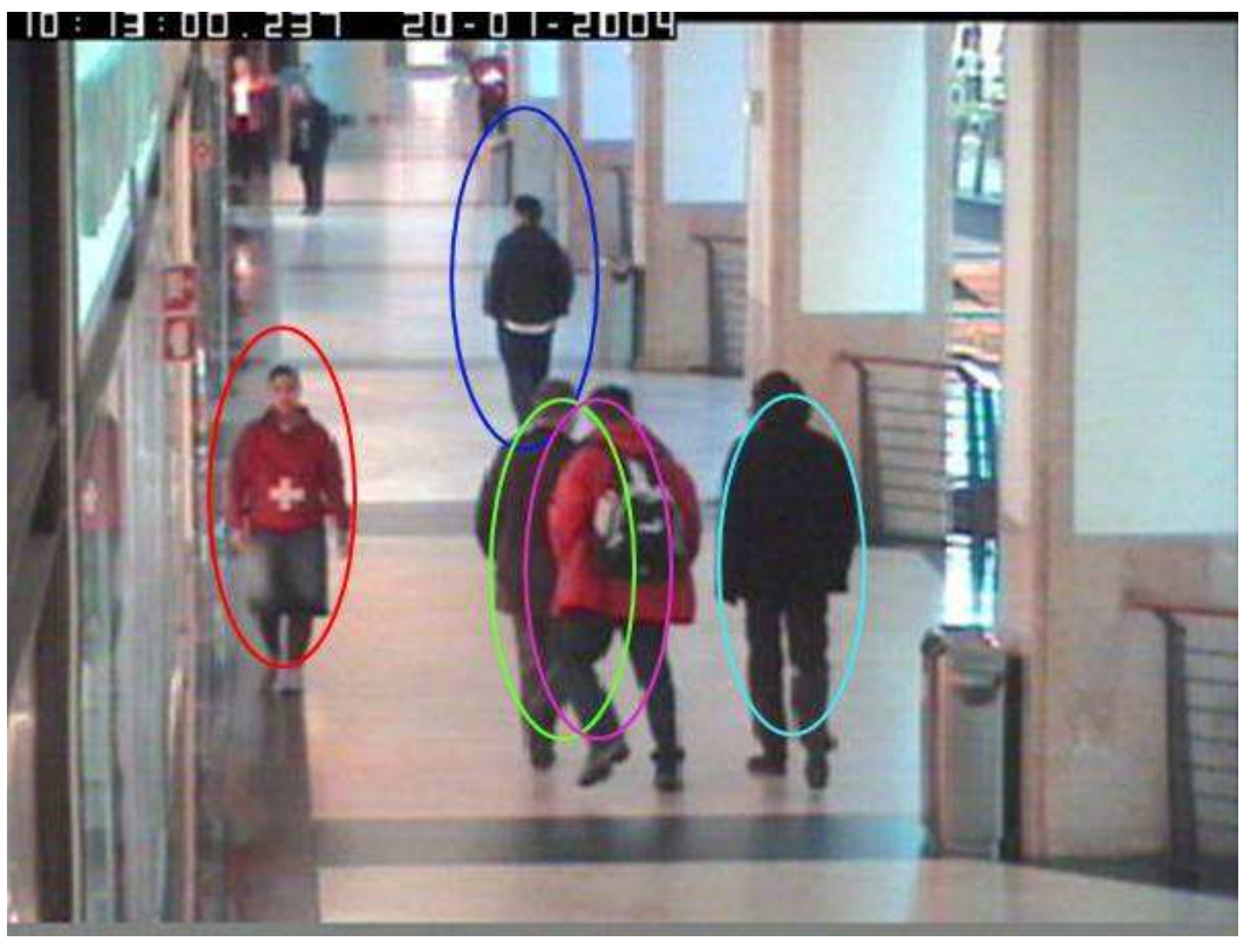}&
\includegraphics[width=1.0in]{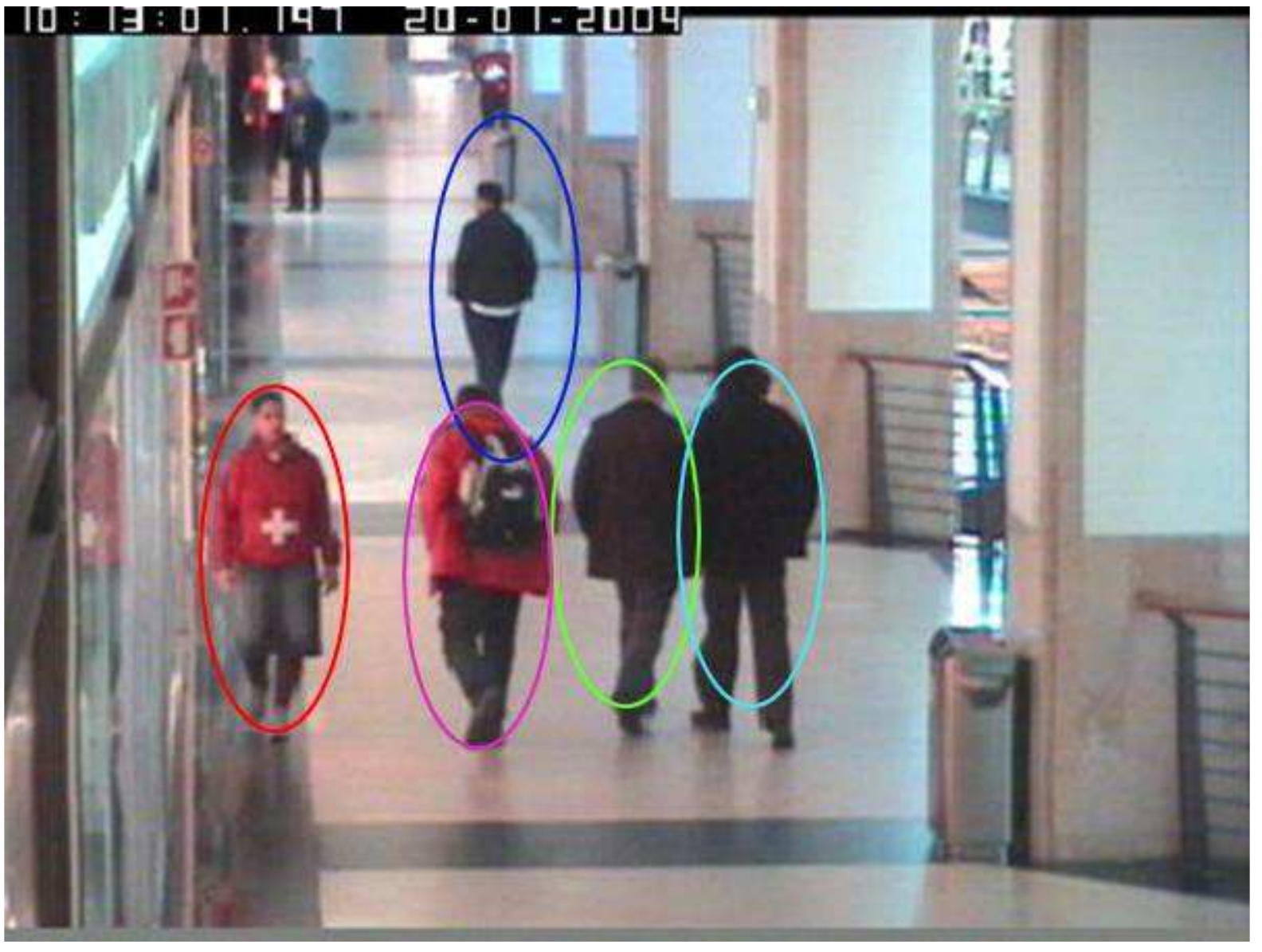}&
\includegraphics[width=1.0in]{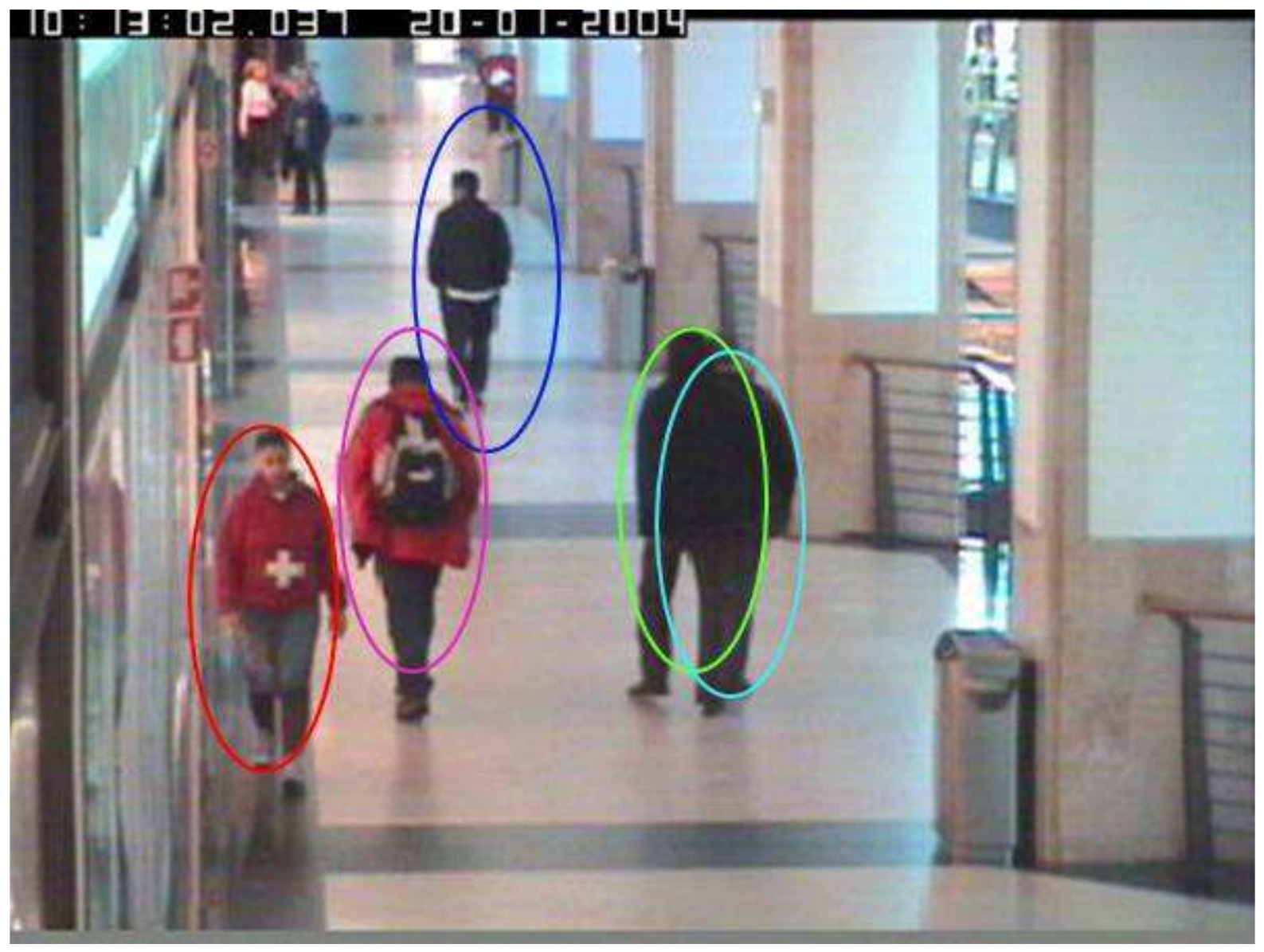}&
\includegraphics[width=1.0in]{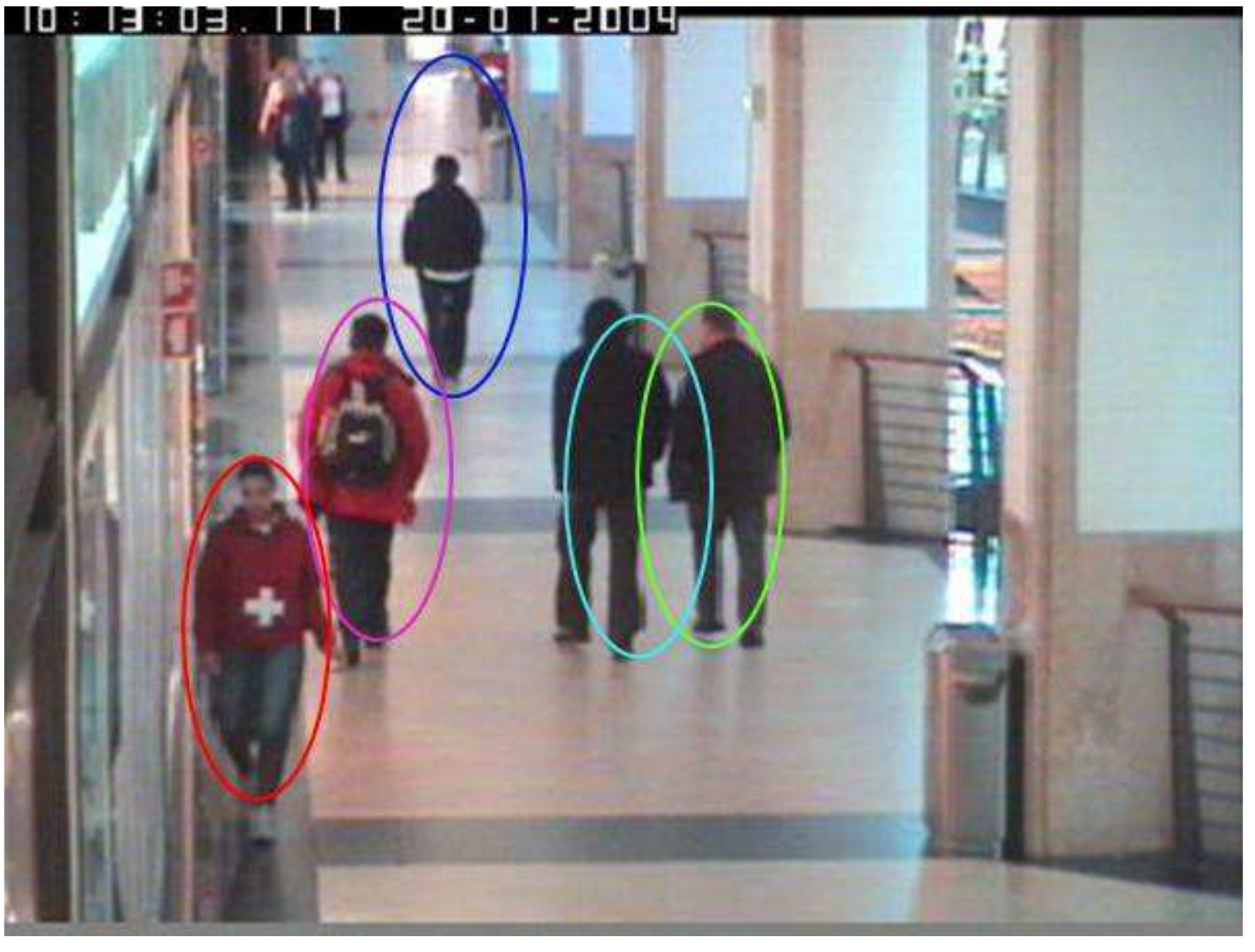}&
\includegraphics[width=1.0in]{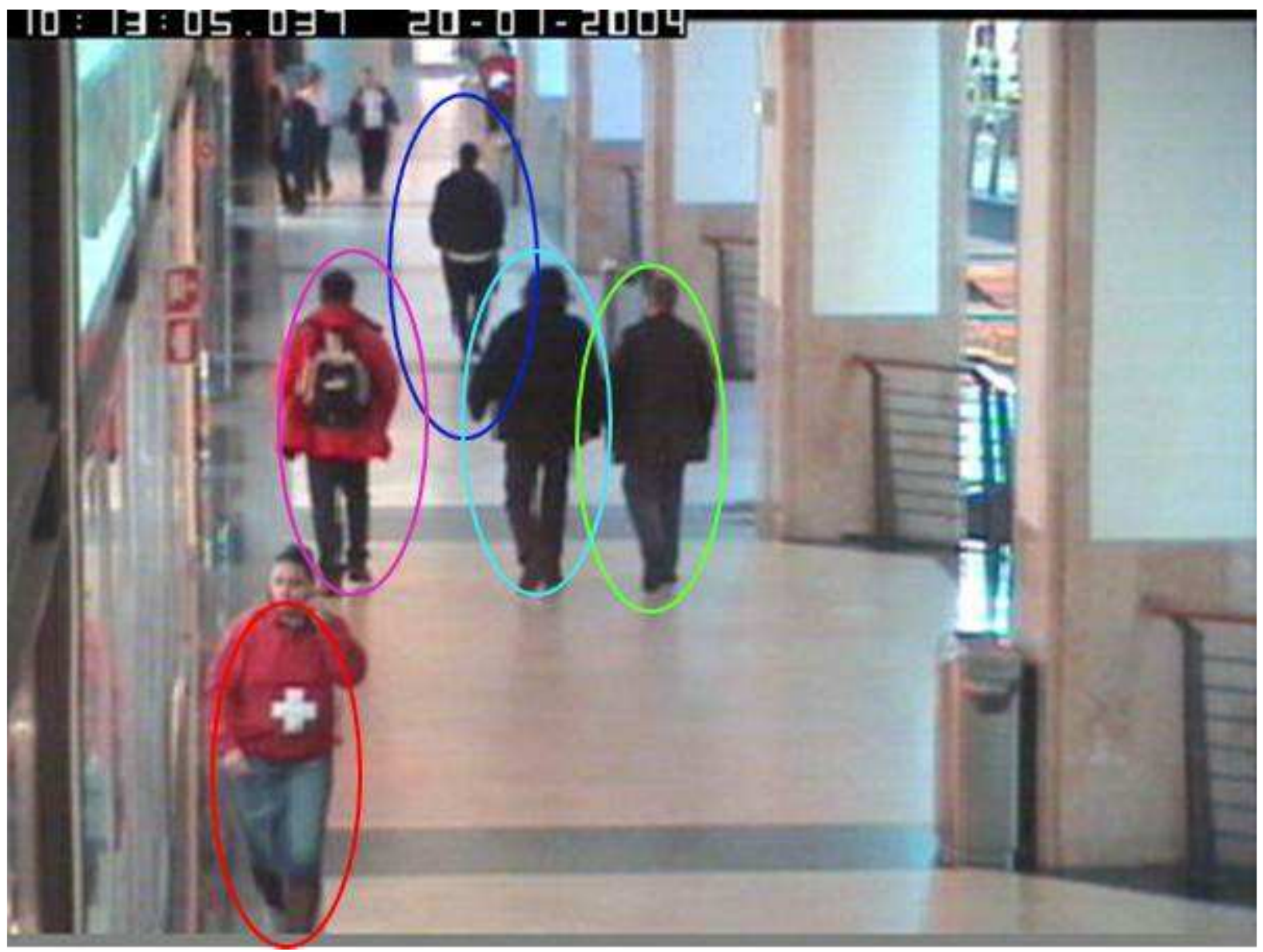}\\
\mbox{(f) Frame 452}&\mbox{(g) Frame 476} & \mbox{(h) Frame 497}&\mbox{(i) Frame 524} & \mbox{(j) Frame 572}
\end{array}$

\caption{Tracking results for certain frames of sequence ``ThreePastShop2cor'' of the CAVIAR dataset : the proposed tracking algorithm can successfully track a variable number of targets while handling complex occlusions.}
\label{figorg2}
\end{figure*}

\begin{figure*}
\centering
$\begin{array}{c@{\hspace{0.35in}}c@{\hspace{0.35in}}c@{\hspace{0.35in}}c@{\hspace{0.35in}}c}
\includegraphics[width=1.0in]{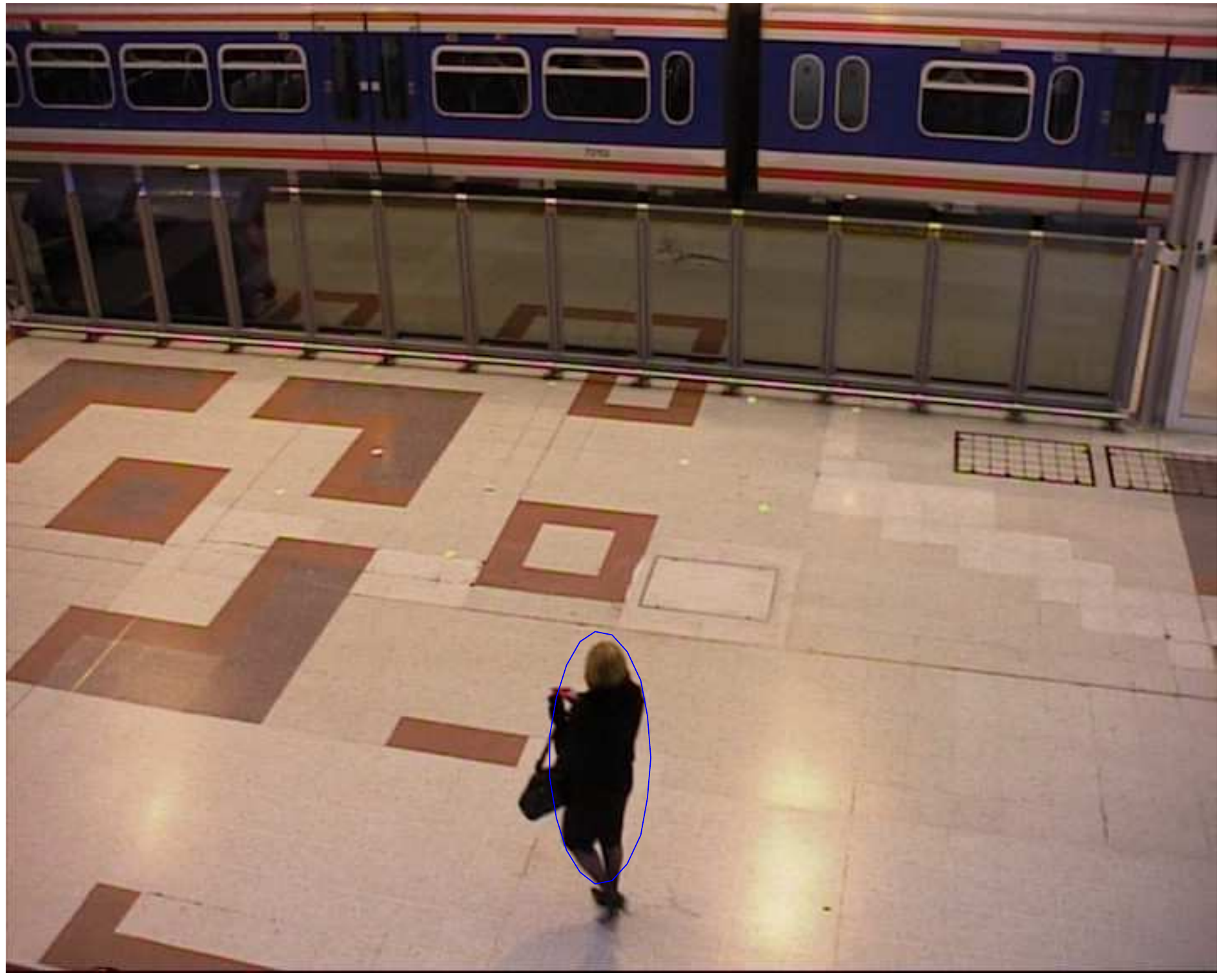} &
\includegraphics[width=1.0in]{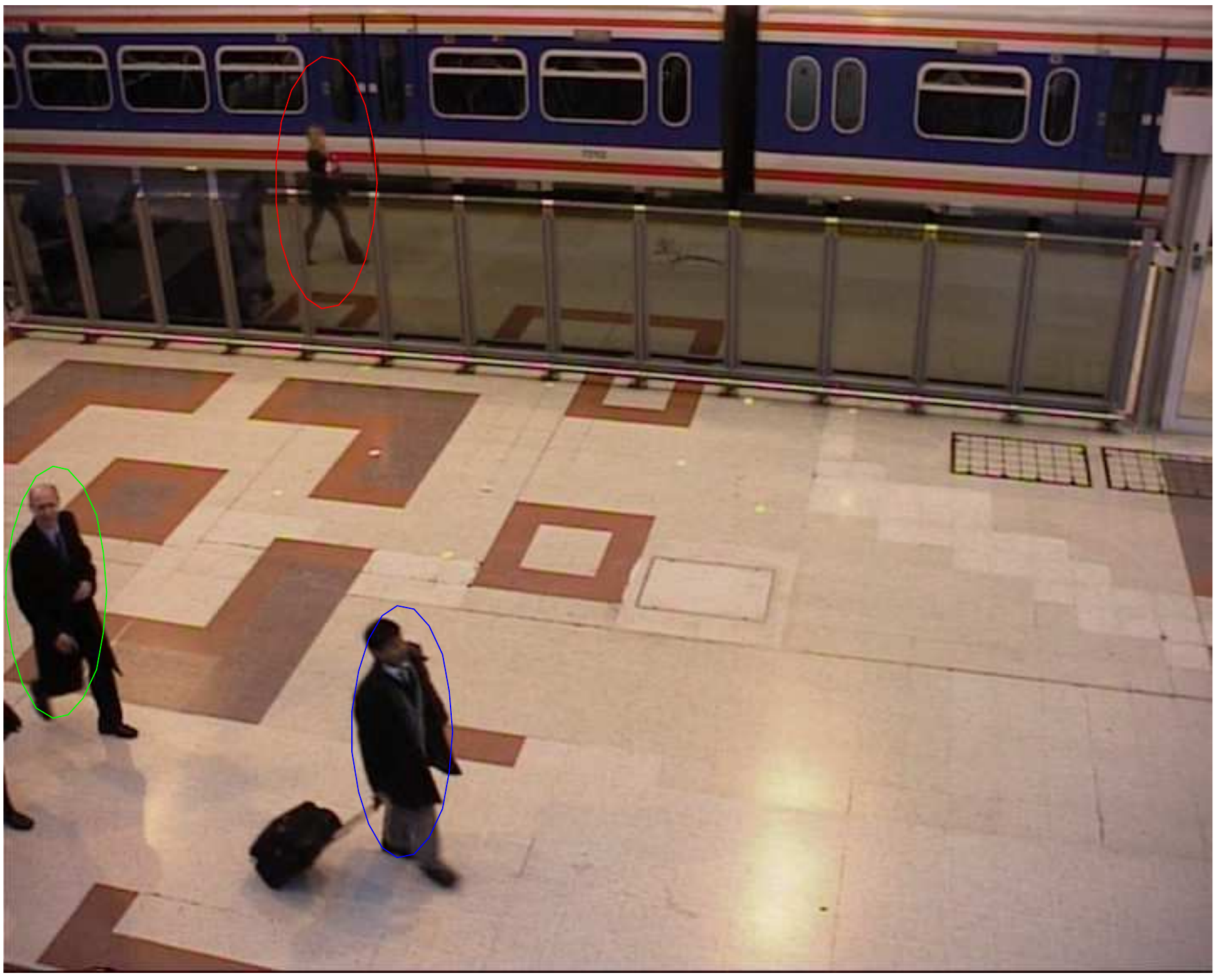} &
\includegraphics[width=1.0in]{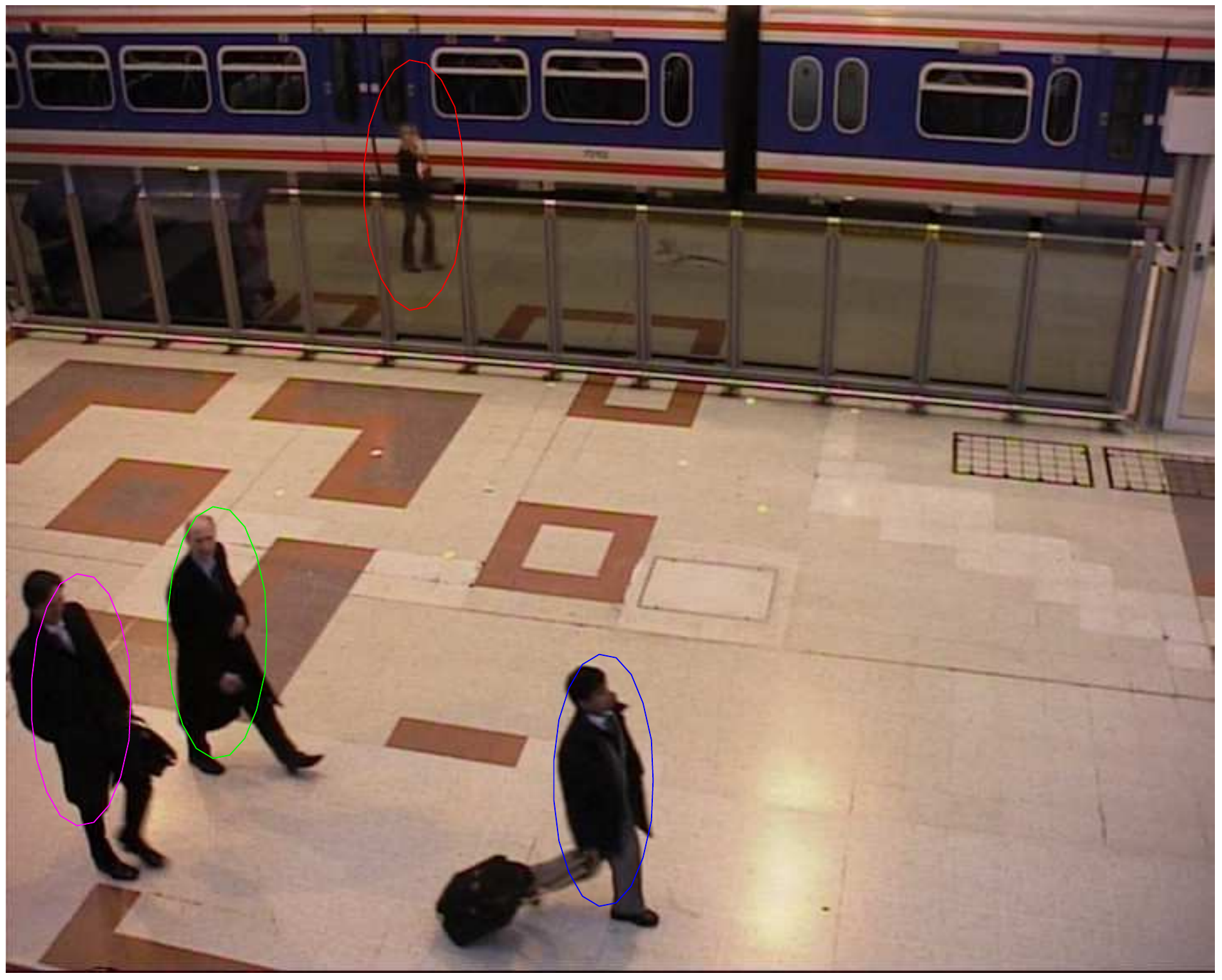}&
\includegraphics[width=1.0in]{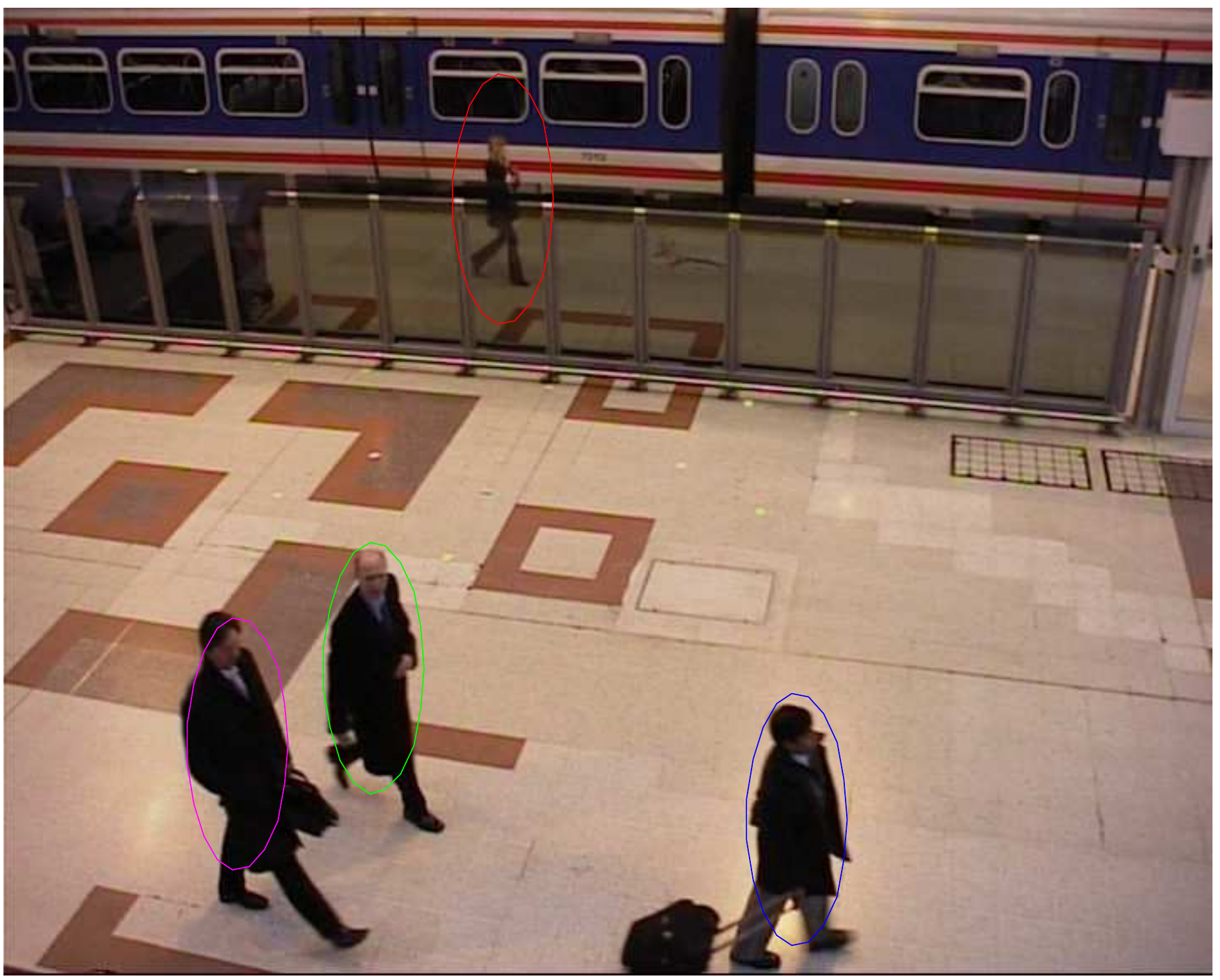}&
\includegraphics[width=1.0in]{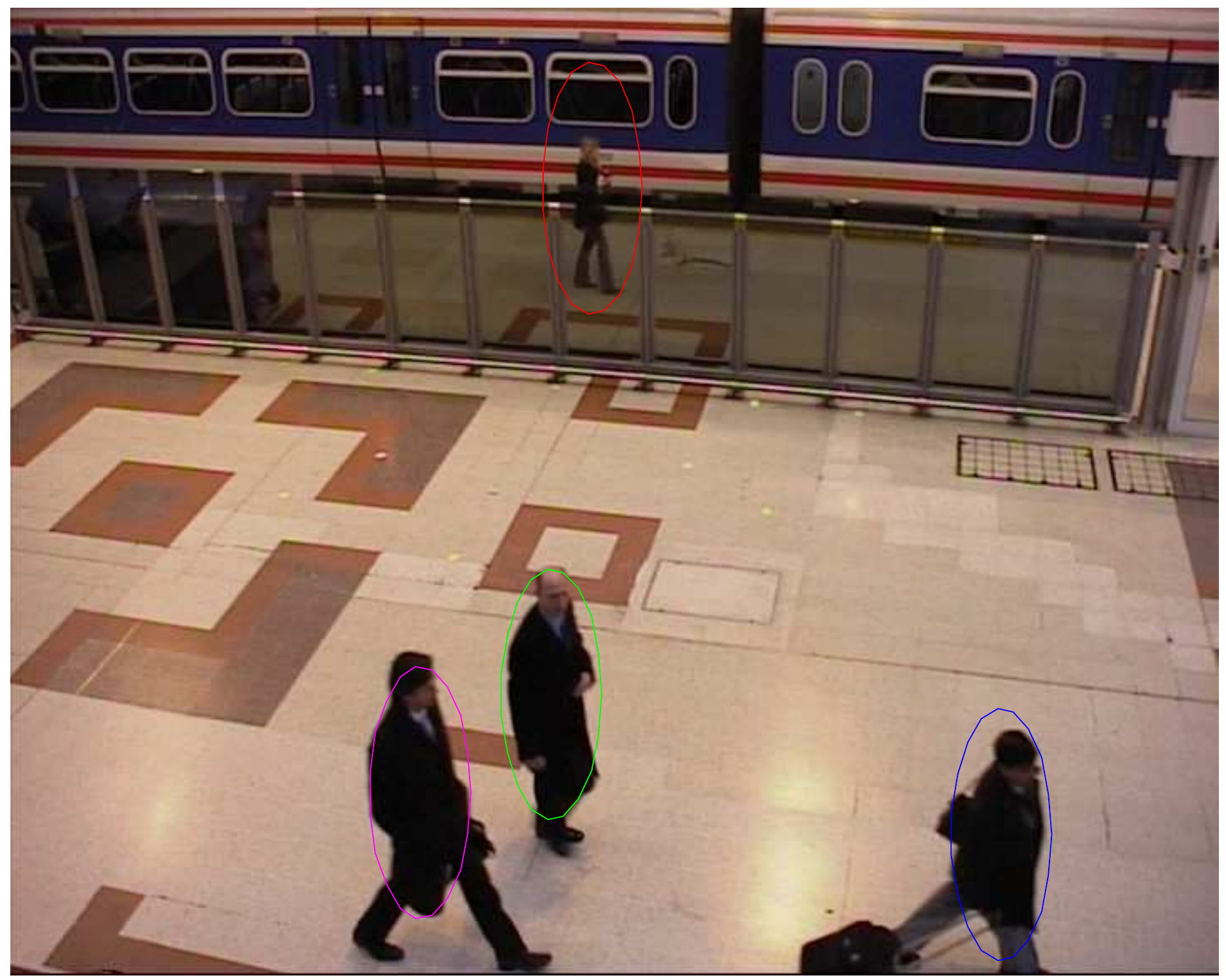}\\

\mbox{(a) Frame 85} & \mbox{(b) Frame 273}&\mbox{(c) Frame 292} & \mbox{(d) Frame 310}&\mbox{(e) Frame 327}
\end{array}$

$\begin{array}{c@{\hspace{0.35in}}c@{\hspace{0.35in}}c@{\hspace{0.35in}}c@{\hspace{0.35in}}c}
\includegraphics[width=1.0in]{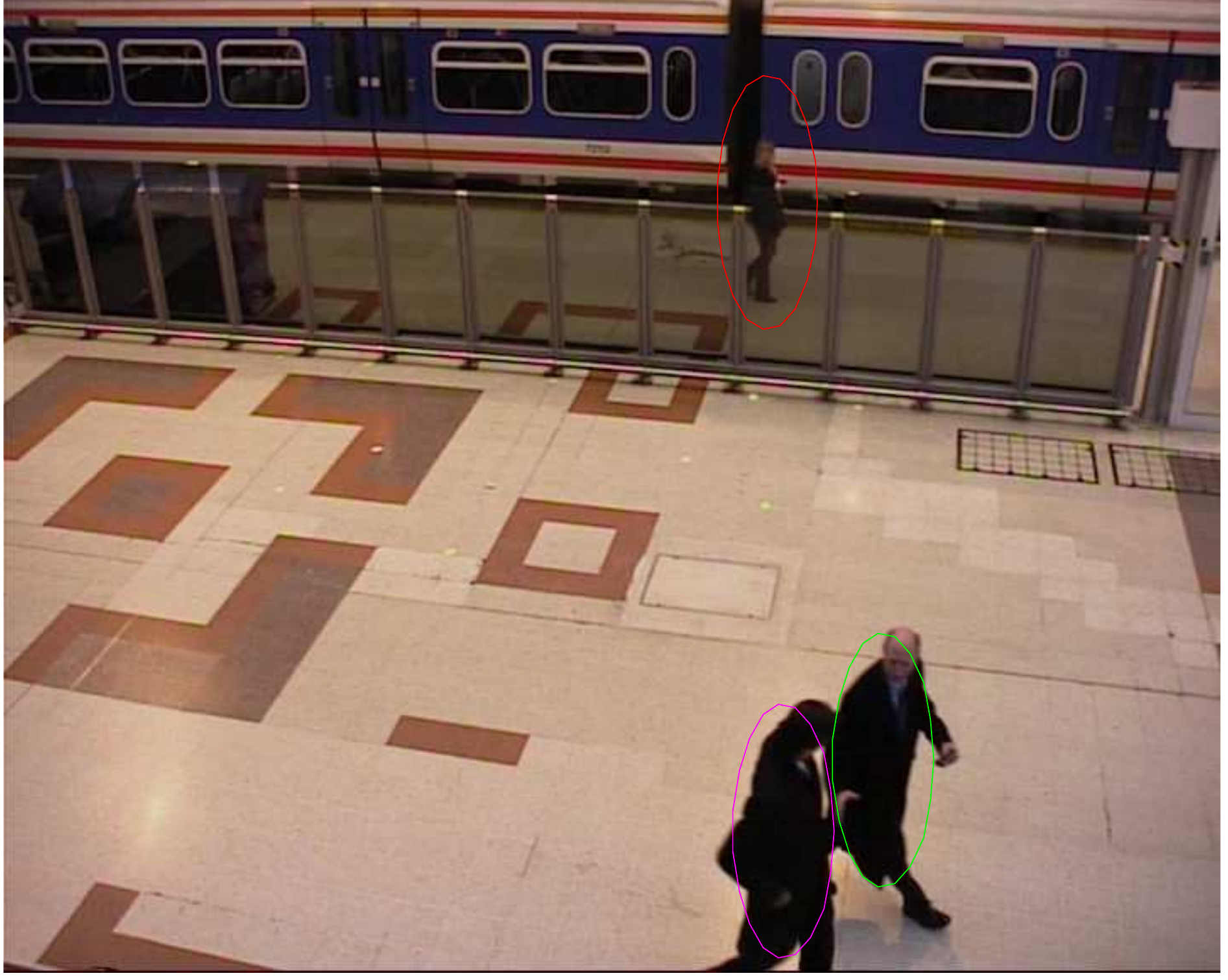}&
\includegraphics[width=1.0in]{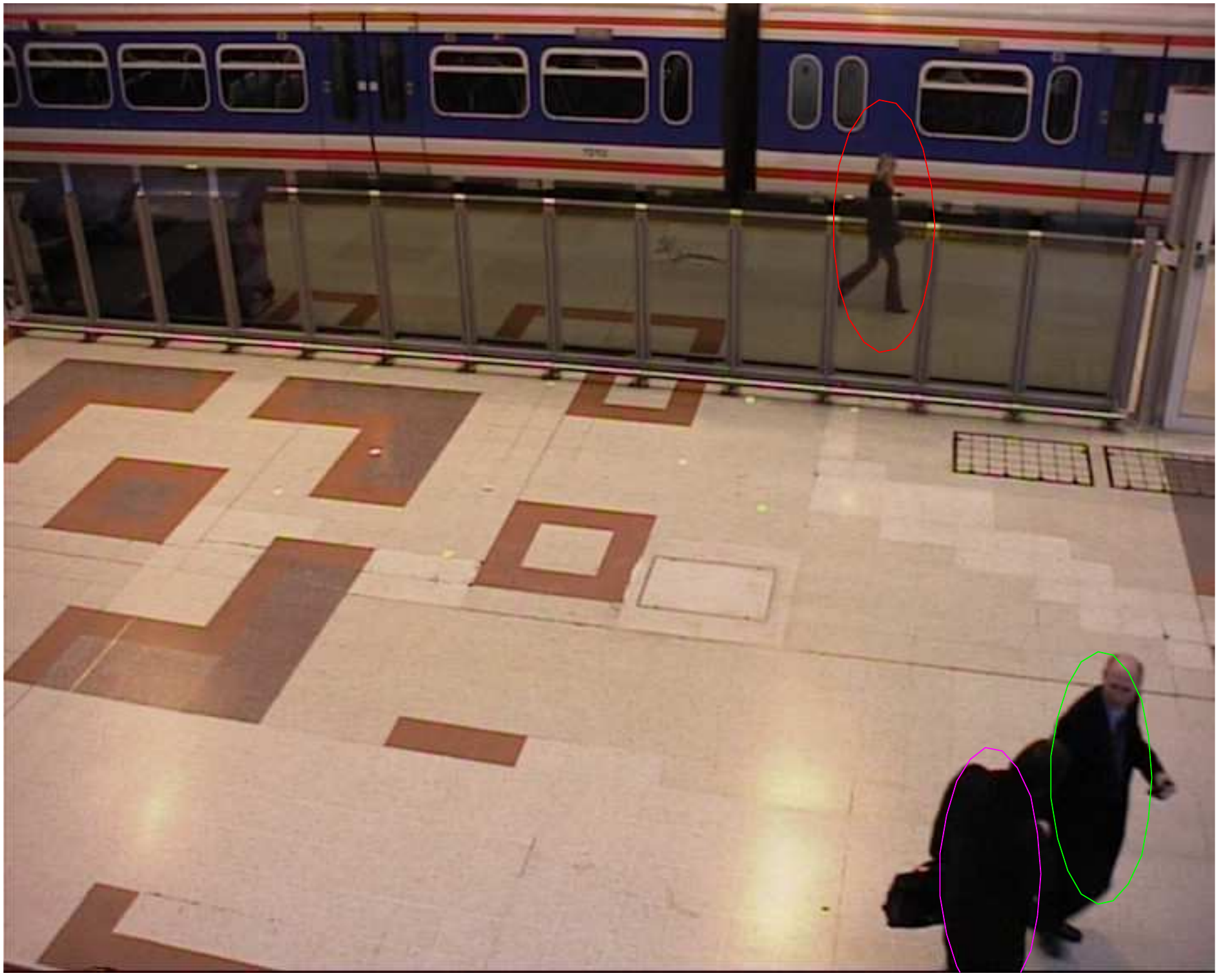}&
\includegraphics[width=1.0in]{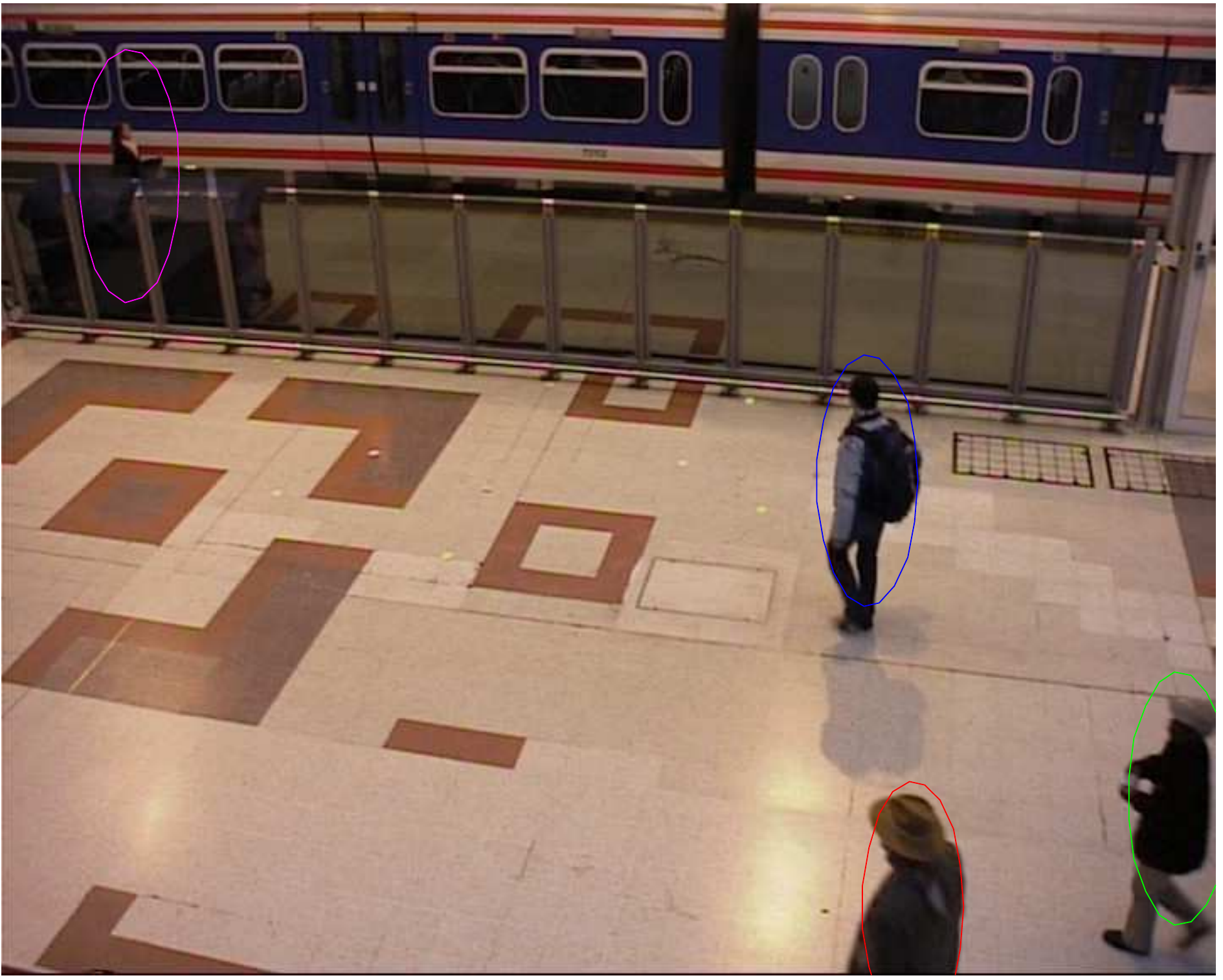}&
\includegraphics[width=1.0in]{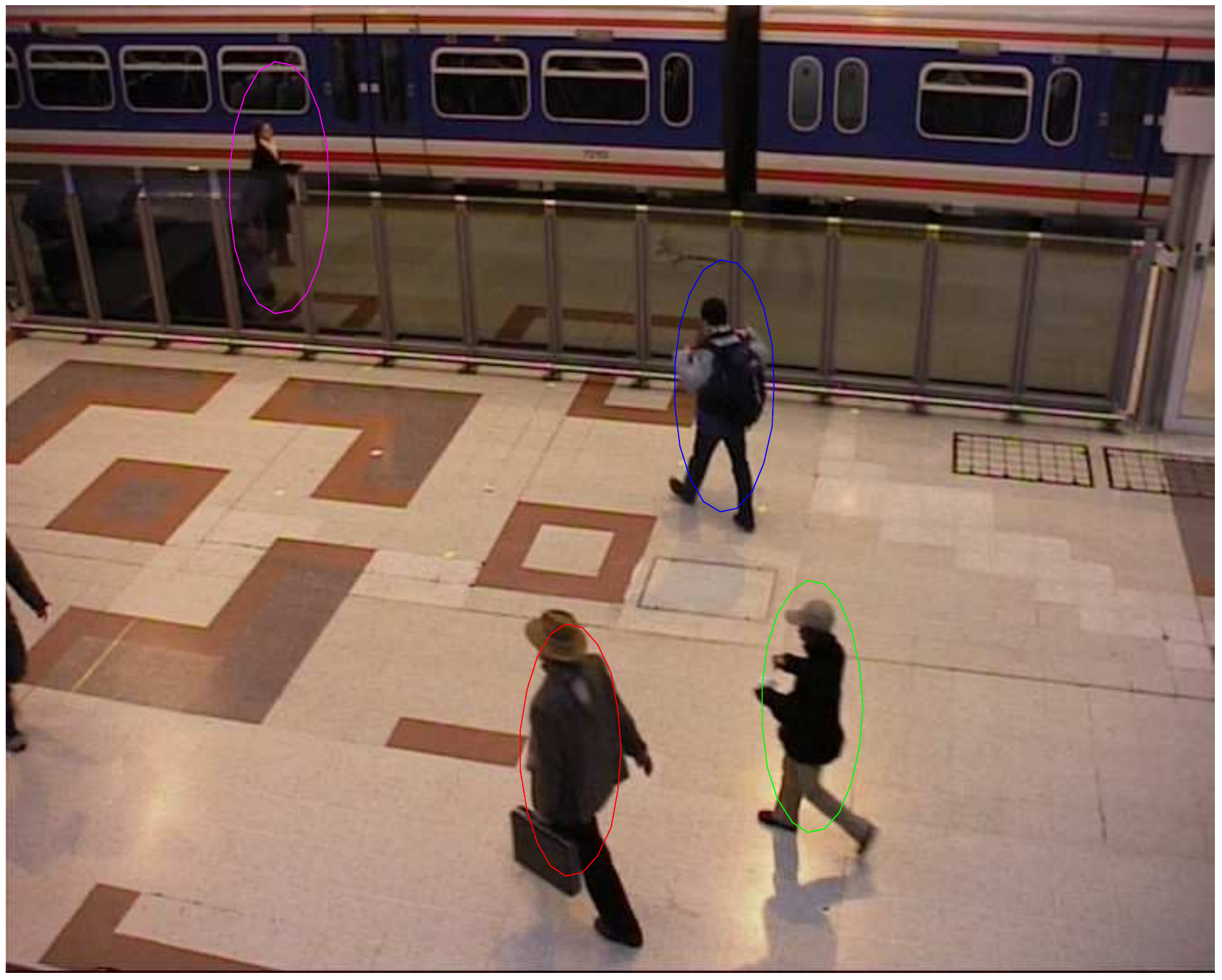}&
\includegraphics[width=1.0in]{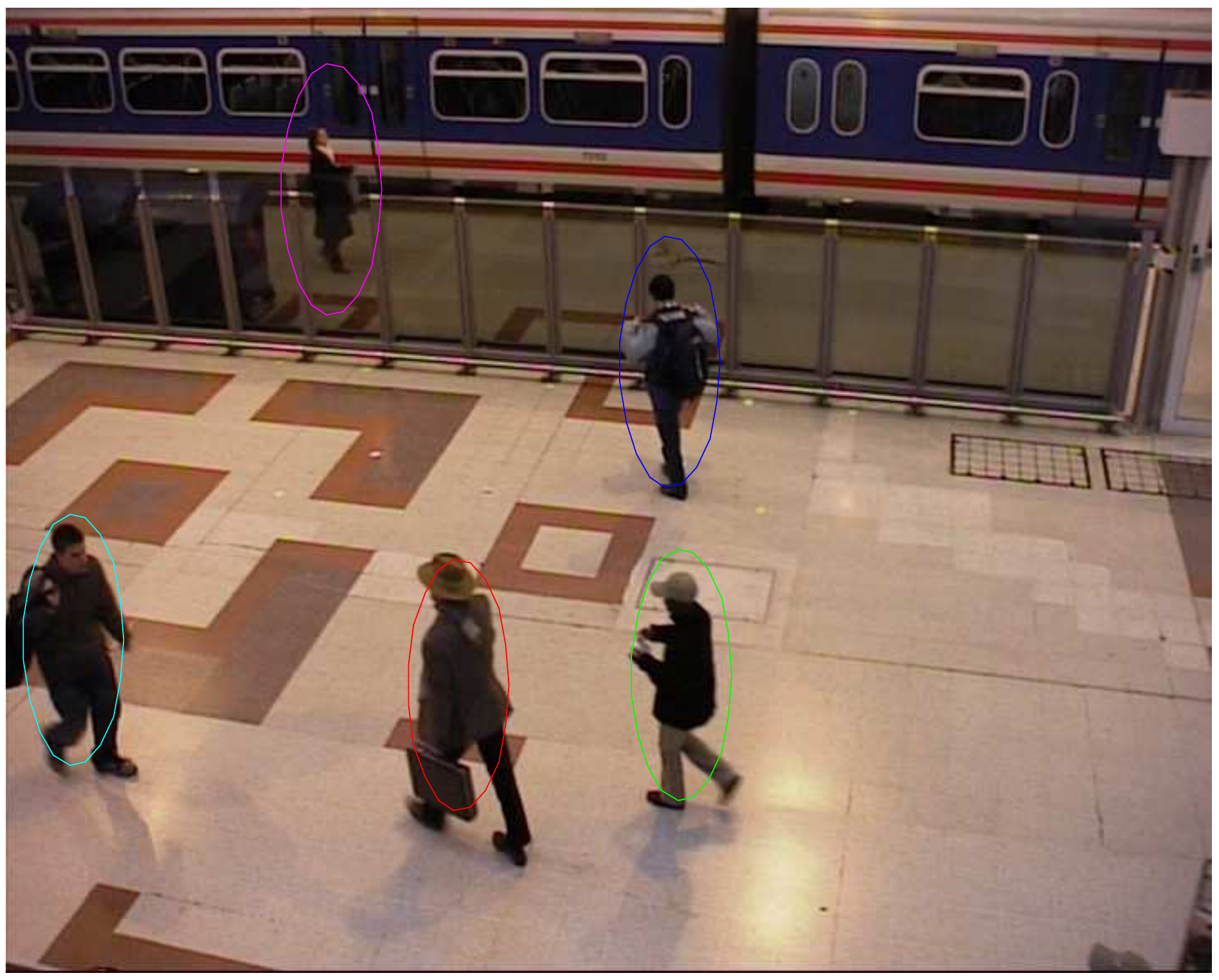}\\
\mbox{(f) Frame 359}&\mbox{(g) Frame 379} & \mbox{(h) Frame 996}&\mbox{(i) Frame 1026} & \mbox{(j) Frame 1038}
\end{array}$

\caption{Tracking results for certain frames of sequence ``S1-T1-C'' of the PETS2006 dataset: the proposed tracking algorithm can successfully track a variable number of targets while handling complex occlusions.}
\label{figorg3}
\end{figure*}

The tracking results for a few selected frames from AV16.3, CAVIAR
and PETS2006 datasets are given in Figs. \ref{fig6}, \ref{figorg2}
and \ref{figorg3} respectively. Blue, red, green, magenta and cyan
ellipses represent first, second, third, fourth and fifth targets,
respectively. Fig. \ref{fig6} shows that the algorithm has
successfully initialized new targets in frames $225$ and $278$. In
frame $278$ it can be seen that the algorithm can cope with the
occlusions between target $1$ and target $2$. Frame $288$ shows that
the tracker keeps tracking all targets even when they are very close
to each other. Frames $320$ and $326$ show that the algorithm has
successfully handled the occlusion between targets $2$ and $3$. In
frame $375$ we can see that target $2$ has left the room and the
algorithm has automatically removed its tracker, which is started
again when the target has returned back in frame $420$. Success in
dealing with occlusions between targets $3$ and $1$ can be seen in
frames $389$ and $405$.

Results in Fig. \ref{figorg2} demonstrate that new target
appearing in frame $334$ is initialized. In frame $476$ it can be
seen that the tracker has solved the occlusion between target three
and four. Similarly, frames $524$ and $572$ show that the tracker
copes with occlusion between target three and five. A similar
behavior is also observed in Fig. \ref{figorg3}.

\subsubsection{CLEAR MOT matrices}
The tracking performance of the proposed algorithm is measured based
on two different performance measures: CLEAR multiple object
tracking (MOT) matrices \cite{MOT} and OSPAMT \cite{OSPAMT}. The
performance is compared with the methods proposed in \cite{tinne},
\cite{zia} and \cite{Czyz}.

The CLEAR MOT measure includes the multiple object tracking
precision (MOTP) matrix, miss detection rate, false positive rate
and mismatch rate.

Let ${\bf G}_k = [({\bf g}_k^1)^T, \cdots , ({\bf
g}_k^i)^T,\cdots,({\bf g}_k^N)^T]^T$ be the ground truth at time
$k\in[k_s,k_e]$, where $k_s$ and $k_e$ are respectively the starting
and ending points of the observation interval. Each ground truth
vector component ${\bf g}_k^i = ({\bf   {p}}_k^i, {I}_i)$ contains
the actual position and identity of the target $i$. Similarly, ${\bf
O}_k = [({\bf o}_k^1)^T, \cdots , ({\bf o}_k^i)^T,\cdots,({\bf
o}_k^N)^T]^T$ represents the output of the tracking algorithm at
time $k$, where each ${\bf o}_k^i = ({\hat{\bf {p}}}_k^i,
\hat{{I}}_i)$ represents the estimated location vector and identity
variable of target $i$. At every time $k$ the error is defined as:
\begin{itemize}
\item Missed detections corresponding to the number of missed targets, calculated based on the difference between the ground truth and the estimates from the developed technique.
\item Mismatch corresponding to the number of targets which have given a wrong identity.
\item False positives corresponding to the estimated target locations which are not associated with any of the targets.
\end{itemize}
The threshold is set at $45cm$, which is the distance between the estimated position of a target and its ground truth beyond which it is considered as a missed detection.

If we assume that ${m}_k$, ${mme}_k$, ${fp}_k$ and ${gt}_k$ are respectively the total number of missed detections, mismatches, false positives and ground truths at time $k$, the errors are calculated as \cite{MOT}
\begin{equation}
\overline {m}=  \frac{\sum_{k}{m}_k}{\sum_{k}{gt}_k}, \overline{mme} = \frac{\sum_{k}{mme}_k}{\sum_{k}{gt}_k}, \overline{fp} = \frac{\sum_{k}{fp}_k}{\sum_{k}{gt}_k}.
\end{equation}
The precision is calculated as \cite{MOT}
\begin{equation}
MOTP = \frac{\sum_{k,i}{d}_{i,k}}{\sum_{k,i}{c}_k},
\end{equation}
where ${d}_{i,k}$ is the distance between the estimated location and the ground truth location of target $i$ and ${c}_k$ is the total number of matches found at time step $k$.
Table \ref{tab7} presents the performance result of the proposed algorithm and the tracking algorithms proposed in \cite{tinne}, \cite{zia} and \cite{Czyz}. The performance results are obtained by using $45$ sequences from AV16.3, CAVIAR and PETS2006 datasets. All three datasets have different environments and backgrounds.
\begin{table}
\centering
\caption{Comparison of performance of the proposed tracking algorithm with the approach of \cite{tinne}, \cite{zia} and \cite{Czyz}}\label{tab7}
 \scalebox{0.9}
 {
\begin{tabular}{|c|c|c|c|}
\hline

Method &   $\overline{m}(\%)$ & $\overline{mme}(\%)$ &  $\overline{fp}(\%)$  \\
\hline
Tracking algorithm \cite{zia}   &     6.41  &   9.32   & 3.92   \\
\cline{1-4}
  Tracking algorithm \cite{Czyz}   &    9.38  &   5.15   & 5.08   \\
\cline{1-4}
  Tracking algorithm \cite{tinne}   &     5.41  &   4.57   & 1.95  \\
\cline{1-4}
Proposed algorithm                  &    5.04 &   1.31   &  0.33    \\
\hline
\end{tabular}
}
\end{table}
The results show that the proposed algorithm has significantly improved performance compared with other techniques \cite{tinne}, \cite{zia} and \cite{Czyz}. 
Table II shows that there is a significant reduction in missed
detections, mismatches and false positives. The reduction in the
missed detections is mainly due to the proposed clustering based
approach along with the new social force model based estimation
technique. A simple particle filter based algorithm without any
interaction model proposed by \cite{Czyz} shows the worst results
mainly due to the highest number of mismatches. The algorithm
proposed in \cite{zia} shows better results because of a better
interaction model. However, the number of missed detections,
mismatches and false positives are higher than that of \cite{tinne}.
This is because \cite{zia} does not propose a solution when two
targets partially occlude each other or appear again after full
occlusion. The technique proposed in \cite{tinne} gives improved
results due to the clustering and the JPDAF however fails when
targets reappear after occlusion. This is due to the fact that only
location information is used for data association and a simple
constant velocity model is used for state prediction. Our proposed
algorithm improves the tracking accuracy by reducing the number of
missed detections, mismatches and false positives. Reduction in the
number of missed detections is mainly due to the proposed particle
filter based force model for particles predictions, while the
proposed data association technique reduces the chances of
mismatches. This improvement is thanks to the utilization of both
feature information of targets along with their locations together
with the location of clusters.

There is a significant $3.26\%$ reduction in the wrong
identifications which has improved the overall accuracy of the
tracker. Average precision results for the video sequences from
AV16.3, CAVIAR and PETS2006 data sets are shown in
Table~\ref{tab-prec}.

\begin{figure}
\centering
$\begin{array}{c@{\hspace{0.32in}}c}
\includegraphics[width=3.0in]{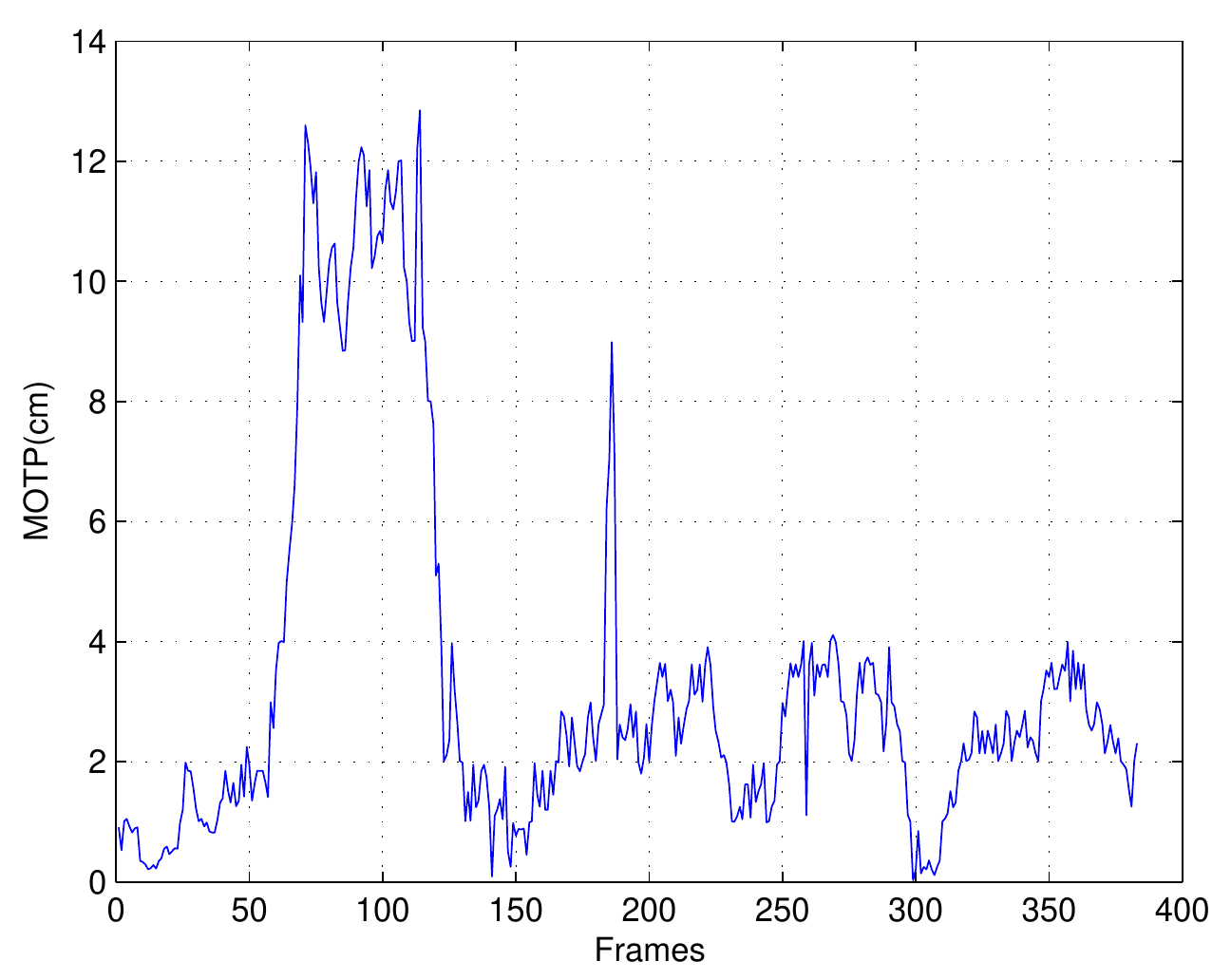} \\
\small{\mbox{(a)}}
\end{array}$
$\begin{array}{c@{\hspace{0.32in}}c}
\includegraphics[width=3.0in]{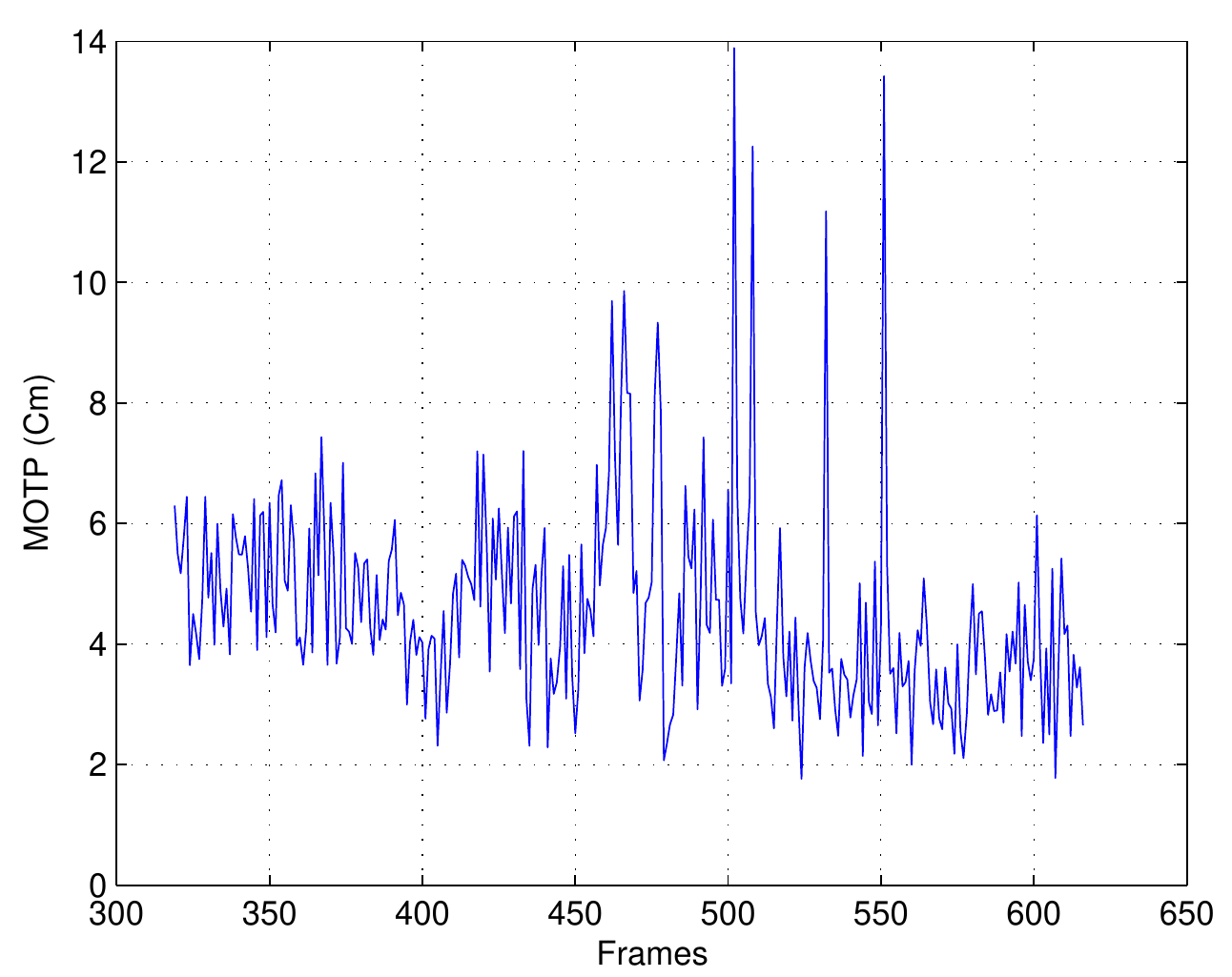} \\
\small \small{\mbox{(b)}}
\end{array}$
$\begin{array}{c@{\hspace{0.32in}}c}
\includegraphics[width=3.0in]{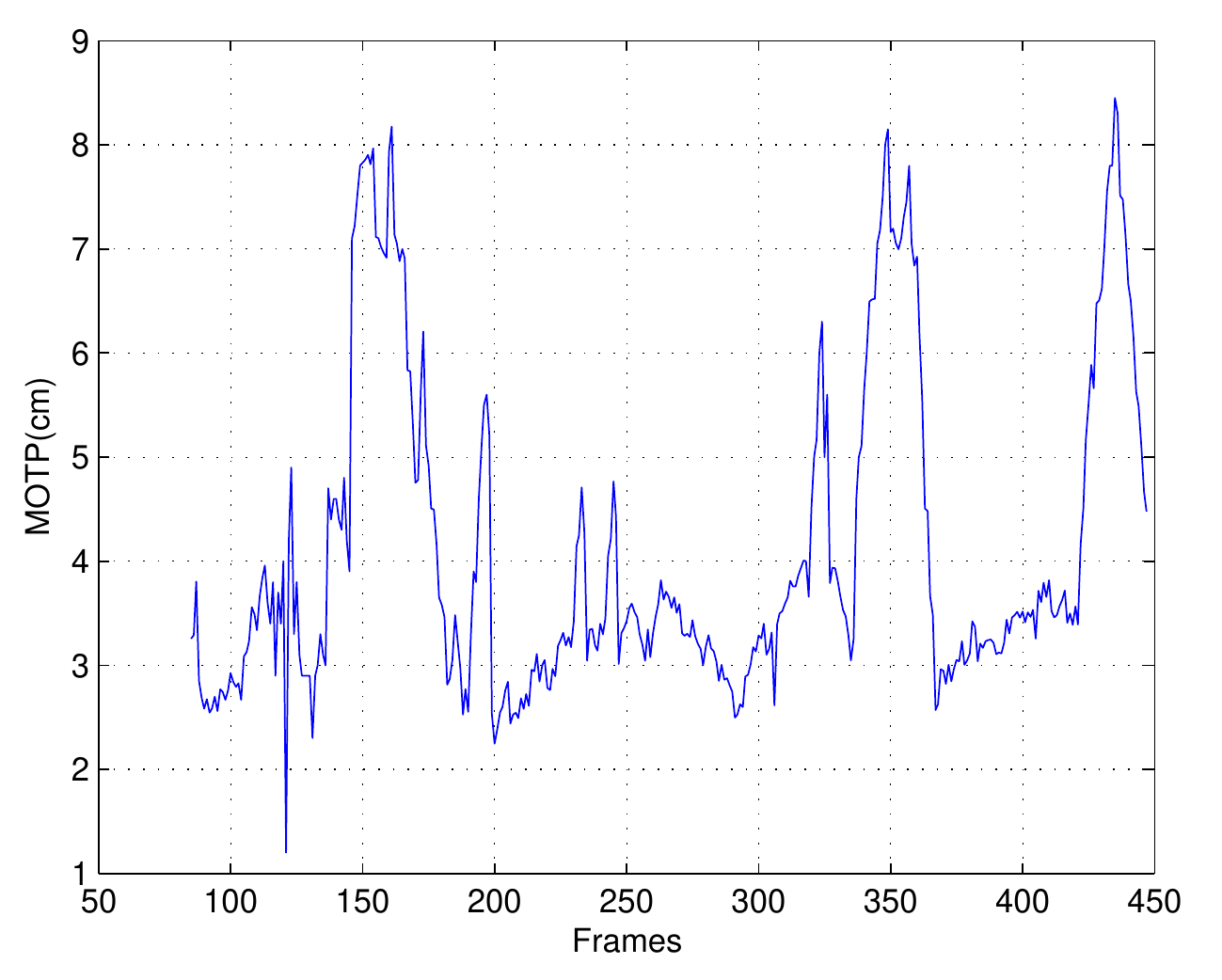}\\
\small \small{\mbox{(c)}}
\end{array}$
\caption{Precision plot against the video frames of sequences: (a) ``seq45-3p-1111\_cam3\_divx\_audio'' of the AV16.3 dataset (b)``ThreePastShop2cor'' of the CAVIAR dataset (c) ``S1-T1-C'' of the PETS2006 dataset}
\label{motp}
\end{figure}

\begin{table}
\centering
\caption{Comparison of precision of the proposed tracking algorithm with the approach of \cite{tinne}, \cite{zia} and \cite{Czyz}}\label{tab-prec}
 \scalebox{0.9}
 \small
 {
\begin{tabular}{|c|c|c|}
\hline

Method &  Dataset & MOTP(cm)   \\
\hline
                                &   AV16.3   &  9.07     \\
\cline{2-3}
Tracking algorithm \cite{zia}   &   CAVIAR   &  11.15     \\
\cline{2-3}
                                &   PETS2006  &  11.02     \\
\cline{1-3}
                                &   AV16.3   &  7.12   \\
\cline{2-3}
Tracking algorithm \cite{Czyz}  &   CAVIAR   &  9.45   \\
\cline{2-3}
                                &   PETS2006  &  8.99   \\
\cline{1-3}
                                &   AV16.3  &  5.85    \\
\cline{2-3}
Tracking algorithm \cite{tinne} &   CAVIAR  &  7.13    \\
\cline{2-3}
                                &   PETS2006  &  7.01    \\
\cline{1-3}
                                &  AV16.3     & 3.97.     \\
\cline{2-3}
Proposed algorithm              &  CAVIAR    &  5.52     \\
\cline{2-3}
                                &  PETS2006    &  5.13     \\
\hline
\end{tabular}
}
\end{table}
Precision plots for two video sequences against the video frame are shown in Fig. \ref{motp}. Results show that the MOTP remains less than $4$cm for most of the frames. It increases to $12$cm for the frames where occlusions occur but it again drops when targets emerge from occlusion.

\subsubsection{OSPAMT}
\begin{figure}[!htb]
\centering
\includegraphics[width=2.8in]{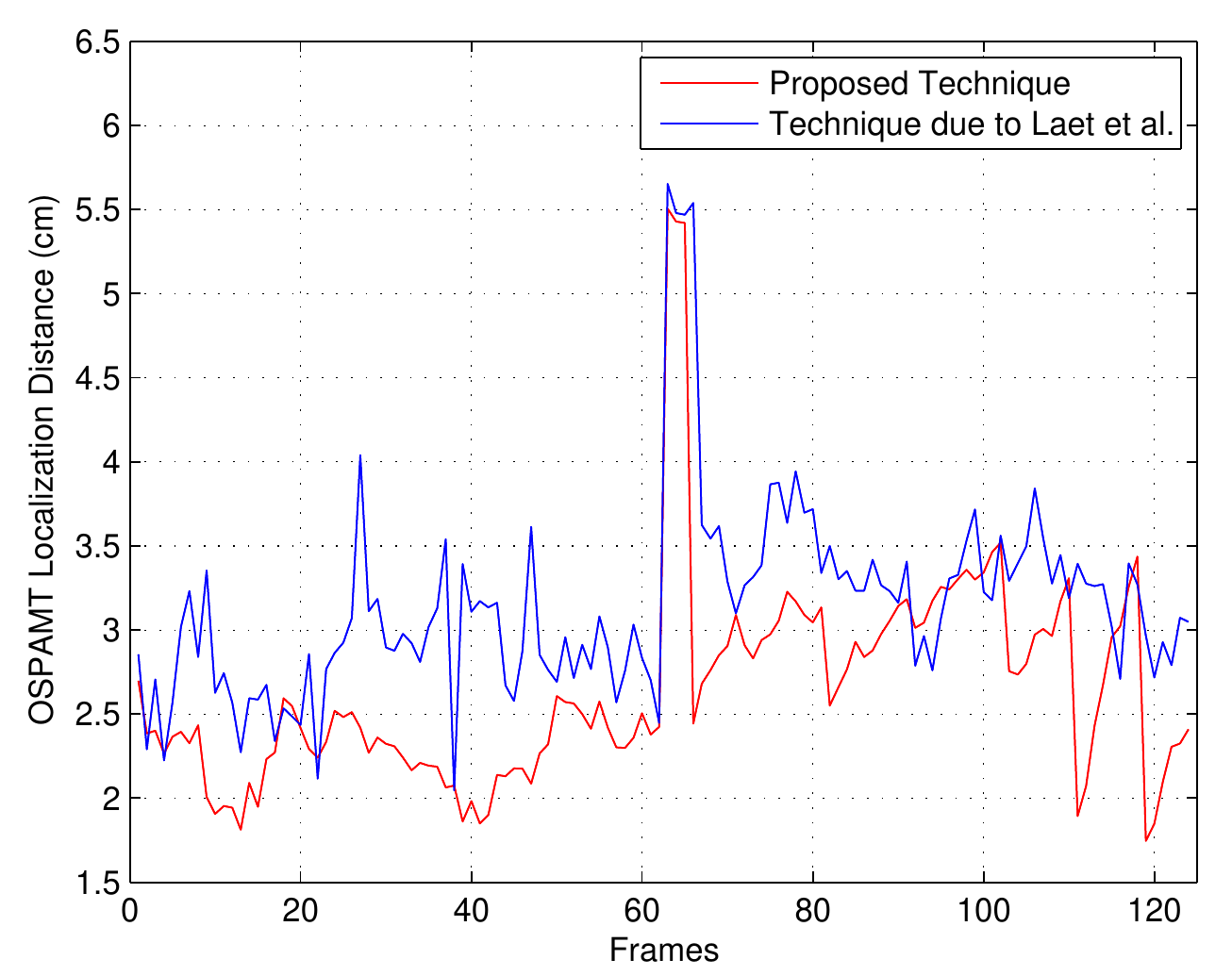}

\caption{OSPAMT localization distance plot against the video frames of sequence ``ThreePastShop2cor'' of the CAVIAR dataset}
\label{ospamt-loc}
\end{figure}

\begin{figure}[!htb]
\centering
\includegraphics[width=2.8in]{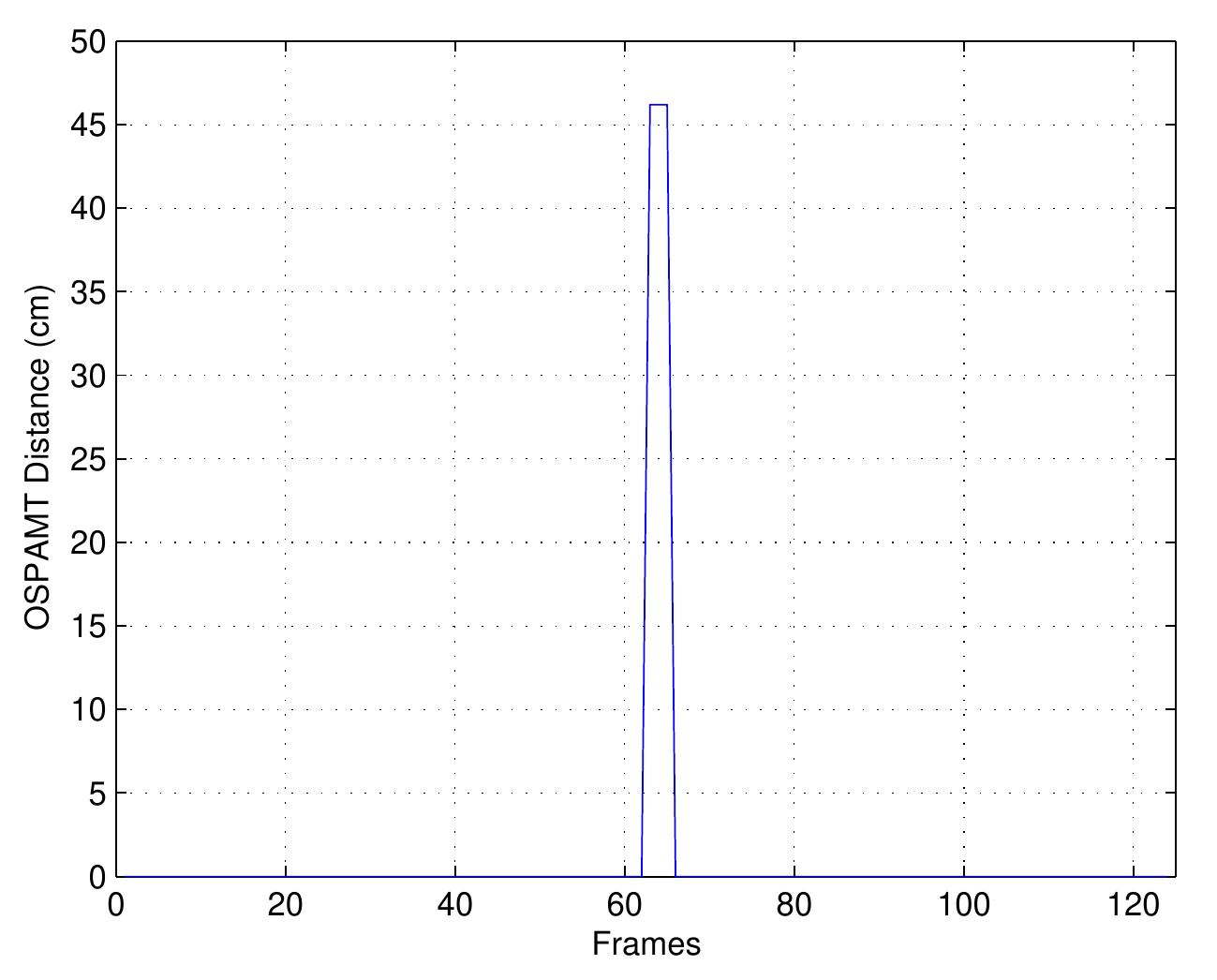}

\caption{OSPAMT distance plot against the video frames of sequence ``ThreePastShop2cor'' of the CAVIAR dataset.}
\label{osmamt-card}
\end{figure}

A new performance metric, optimal subpattern assignment metric for
multiple tracks (OSPAMT) \cite{OSPAMT}  is also used to evaluate the
performance of the proposed technique and approach of \cite{tinne}. The OSPAMT metric calculates
the localization distance $\mathcal{D}_k^{Loc}(\omega,\omega')$
between a set of true tracks $\omega$ and a set of estimated tracks
$\omega'$ at time step $k$ as follows
\begin{equation}
\mathcal{D}^{Loc}_k(\omega,\omega') = \biggr[\frac{1}{n_t}\sum_{i =
1}^{|\omega|}d_k(\tau_i^{\omega},\omega')\biggr]^{1/p},
\end{equation}
where $n_t = \max\{n_t^\omega, n_t^{\omega'}\}$ and $n_t^\omega$ and $n_t^{\omega'}$ are the number of targets in true and estimated set of tracks, respectively, at time step $k$. The distance $d_t(\tau_i^{\omega},\omega')$ is between the $i^{th}$ true track and the set of estimated tracks $\omega'$, complete explanation of which can be found in Section IV of \cite{OSPAMT}.
The distance $\mathcal{D}^{Card}_k(\omega,\omega')$ at time index $k$ is calculated as
\begin{equation}
\label{osp1} \mathcal{D}^{Card}_k(\omega,\omega') =
\biggr[\frac{1}{n_t}\mathcal{S}_k\biggr]^{1/p},
\end{equation}
where $\mathcal{S}_t$ is defined as follows
\begin{equation}
\label{osp2}
\begin{split}
\mathcal{S}_k = \sum_{i = 1}^{|\omega|}|\tau_i^{\omega}(k)|\max\{\bar n_{k,i}^\lambda - 1,0\}(\Delta^p + c^p) +\\
c^p(n_k - \sum_{i = 1}^{|\omega|}\bar
n_{k,i}^\lambda|\tau_i^{\omega}(k)|),
\end{split}
\end{equation}
where $\bar n_{k,i}^\lambda$  is the number of targets at time $k$ in $\omega'$ assigned to target $i$ in $\omega$, $\lambda$ represents an assignment between tracks in $\omega'$ and the tracks in $\omega$, $\Delta$ is the assignment parameter, $c$ is the cutoff parameter and $p = 2$. Details about all these parameters and the OSPAMT metric can be found in \cite{OSPAMT}. For results in Figures \ref{ospamt-loc} and \ref{osmamt-card} we have used $c = 80$ and $\Delta = 10$.  Note that $\mathcal{D}^{Card}_k(\omega,\omega')$ is basically a distance between the objects.
Figures \ref{ospamt-loc} and \ref{osmamt-card} show the OSPAMT distances, (\ref{osp1}) and (\ref{osp2}), respectively, plotted against the video frame index. Figure \ref{ospamt-loc} shows low localization errors compared to the technique proposed in \cite{tinne}. Figure \ref{osmamt-card} indicates a peak dynamic error when a new target enters the field of view which is between frame numbers $63$ and $65$.

%
%
%

\subsection{Computational complexity}
A variable measurement size technique reduces significantly the
computational complexity. The measurement size is increased or
decreased on the basis of the distance between the targets.
Downsampling of foreground pixels is performed when the distance
between targets becomes $80$cm or less. The decrease in the
measurement size reduces the time and number of iterations for
reaching clustering convergence which results in improved
computational complexity. This is shown in Table \ref{tab6} with the
help of a few selected frames from sequence
``seq45-3p-1111\_cam3\_divx\_audio'' of the AV16.3 dataset.
\begin{table}
\centering
\caption{Comparison of convergence iterations for different measurement sizes }\label{tab6}
\small
 \scalebox{0.9}
 {
\begin{tabular}{|c|c|c|c|c|c|}
\hline

Frame & Minimum  & \multicolumn{2}{l|}{Measurement Size} & \multicolumn{2}{l|}{Convergence Iterations} \\
\cline{3-6}
No. &   Distance &   Original & Reduced & Original  &  Reduced  \\
\hline
210    &   227.33    &  11439  &   1255   & 193  &   15     \\
\cline{1-6}
218    &    201.01    &  14496 &   1614   &  389  &   20      \\
\cline{1-6}
221       &  186.85   & 15887   &   1780    &   282  &    25    \\
\cline{1-6}
225       & 159.64  &  17792  &   1966   &  279   &   29    \\
\cline{1-6}
228       & 137.60  &  17819  &   1964   &  334  &   40    \\
\hline
\end{tabular}
}
\end{table}

Table \ref{tab6} also demonstrates that the proposed technique reduces the number of iterations needed for achieving convergence. The reduction in the number of iterations for convergence improves the run-time of the tracking algorithm. The average run-time (calculated using $45$ video sequences from three datasets) of the proposed algorithm due to the reduction in number of iterations is $0.587$ seconds per frame, as compared to the run-time without measurement reduction which is $2.611$ seconds per frame. The run-time of the approach of \cite{tinne} is $1.417$ seconds per frame. This run-time comparison is made by implementing the algorithms on MATLAB (version R2012a) with a $3.2$GHz I5 processor.

\subsection{Summary}
Successful background subtraction is achieved with the help of the
codebook background subtraction technique and the results are shown
in Figs. \ref{fig1}-\ref{figbs3}. Grouping
foreground pixels into clusters with the help of the proposed
variational Bayesian technique is  presented in Figs. \ref{fig3}-\ref{figcls3}. Tracking results are shown in Figs.
\ref{fig6}-\ref{figorg3}. The variational
Bayesian clustering improves the overall tracking results especially
during close inter-target occlusions. The JPDAF based technique
proposed in \cite{tinne} does not always assign the appropriate
clusters to the targets and hence results in tracking failures while
solving the complex occlusions.

The graphical results shown in Figs. \ref{var-tar-av} and \ref{var-tar-cav} indicate that
estimating the number of targets on the basis of the number of
clusters does not produce accurate results because the number of
clusters does not always vary with the number of targets. Figs.
\ref{fig6}, \ref{figorg2} and \ref{figorg3} show that accurate
target number estimation results are achieved by the proposed
technique thanks to exploiting the sizes and locations of clusters,
and the estimated state of the targets at the previous time step.

The results in Tables \ref{tab7} and \ref{tab-prec} confirm that the
overall performance of the proposed tracking algorithm results in an
improvement as compared with the recently proposed techniques
\cite{tinne}, \cite{zia} and \cite{Czyz}.

\section{Discussion}
\label{sec-dis}

The quality of the clustering results influences the proposed data
association technique. We initialize the clusters on the basis of
the estimated locations of targets and by taking into account
merging and splitting of targets. The motion of people is predicted
with the social force model. This yields accurate clustering results
in scenarios where one cluster contains regions of multiple targets.
Therefore, the proposed data association technique performs well
even during close interactions of targets. It associates multiple clusters to targets with the correct proportion of probability. These association results help in achieving highly accurate tracking of multiple targets even during their close interactions and partial occlusions. The extensive evaluation of the proposed technique on $45$ sequences from three very different datasets validates the tracking accuracy 
of the proposed algorithm in such scenarios.

A further improvement in accuracy of cluster-to-target association
can be achieved by calculating joint associations between multiple
clusters. A several-to-one or several-to-several correspondence
strategy can be adopted. However, for these strategies the
association problem is combinatorial and complexity may become
intractable because it will grow geometrically. The
cluster-to-target association can also be improved by considering
the regions of multiple targets in a cluster as unique identities
and associating those regions to respective targets
\cite{jpdaf-alternate}. 



\section{Conclusions}
\label{sec4}
A learned variational Bayesian clustering and social force model
algorithm has been presented for multi-target tracking by using
multiple measurements originating from a target. An improved data
association technique is developed that exploits the clustering
information to solve complex inter-target occlusions. The algorithm
accurately and robustly handles a variable number of targets. Its is
compared extensively with recent techniques~\cite{zia}, \cite{tinne}
and
\cite{Czyz}. 


\vspace{3pt} \textbf{Acknowledgment:}
We are grateful to the support from the UK EPSRC via grant
EP/K021516/1, Bayesian Tracking and Reasoning over Time (BTaRoT). We
greatly appreciate the constructive comments of the Associate Editor
and anonymous reviewers helping us to improve this paper.

\ifCLASSOPTIONcaptionsoff
\fi \vspace{-0.2cm}
\bibliographystyle{IEEEtran}
\bibliography{./include/books,./include/journals,./include/biblo,./include/ref,./include/bib,./include/book}
\section*{Appendices}
\vspace{-0.6cm}
\subsection{Derivation of equation (\ref{eq24})}
\vspace{-0.3cm}
The presented derivation is for time index $k$ and for simplicity we
omit the index $k$ from the equations.
%
According to (\ref{eq18b}) we can write
\begin{equation}\label{aa2}
\ln \mathcal{Q}^*({\bf B}) = E_{{\bf C}, \boldsymbol{\mu},
\boldsymbol{\Sigma}}[\ln p({\tilde{\bf y}}, {\bf B}, {\bf
C},\boldsymbol{\mu}, \boldsymbol{\Sigma})]+Const.
\end{equation}\\
\vspace{-0.2cm}
and using the decomposition of equation (\ref{eq19}) we have
\begin{equation}\label{aa3}
\begin{split}
\ln \mathcal{Q}^*({\bf B}) = E_{{\bf C}, \boldsymbol{\mu},
\boldsymbol{\Sigma}} [\ln p({\tilde{\bf y}}|{\bf B},
{\boldsymbol{\mu}},{\bf \Sigma})+ \ln p({\bf B}|{\bf C})+
\\\ln p({\bf C})+
\ln p({\boldsymbol{\mu}}|{\bf \Sigma})+\ln p({\bf \Sigma})]
+Const.
\end{split}
\end{equation}\\
The optimal distribution over ${\bf B}$ is calculated by considering
different forms of $q({\bf B})$ and by keeping the remaining terms
constant. Therefore, quantities not depending on ${\bf B}$ will
merge into the constant and we can write equation (\ref{aa3}) as
\begin{equation}\label{aa4}
\begin{split}
\ln \mathcal{Q}^*({\bf B}) = E_{{\bf C}, \boldsymbol{\mu},
\boldsymbol{\Sigma}} [\ln p({\tilde{\bf y}}|{\bf B},
{\boldsymbol{\mu}},{\bf \Sigma})+ \ln p({\bf B}|{\bf C})]
+Const.\\
= E_{ \boldsymbol{\mu}, \boldsymbol{\Sigma}} [\ln p({\tilde{\bf y}}|{\bf
B}, {\boldsymbol{\mu}},{\bf \Sigma})]+ E_{{\bf C}}[\ln p({\bf
B}|{\bf C})] +Const.
\end{split}
\end{equation}\\
By using equations (\ref{eq10}) and (\ref{eq12}) we can write
\begin{equation}\label{aa6}
\begin{split}
\ln \mathcal{Q}^*({\bf B}) = E_{ \boldsymbol{\mu}, \boldsymbol{\Sigma}}
\biggr[\ln \prod_{j=1}^M\prod_{q=1}^{\kappa}\mathcal{N}{({\tilde{\bf y}}^j|{\boldsymbol{\mu}}^q,{\bf \Sigma}^q)}^{b^{j,q}}\biggr]+~~~~~~~~\\
E_{{\bf C}}\biggr[\ln \prod_{j=1}^M\prod_{q=1}^{\kappa}{(C^{q})}^{b^{j,q}}\biggr]
+Const.\\
 =\sum_{j=1}^M\sum_{q=1}^{\kappa}
{b^{j,q}}E_{ \boldsymbol{\mu}, \boldsymbol{\Sigma}}[\ln\mathcal{N}{({\tilde{\bf y}}^j|{\boldsymbol{\mu}}^q,{\bf \Sigma}^q)}]+~~~~~~~~\\
 \sum_{j=1}^M\sum_{q=1}^{\kappa}{{b^{j,q}E_{{\bf C}}\ln(C^{q})}}
+Const.\\
= \sum_{j=1}^M\sum_{q=1}^{\kappa}
{b^{j,q}}E_{ \boldsymbol{\mu}, \boldsymbol{\Sigma}}
\biggr[\ln\biggr\{\frac{1}{(2\pi)^{D/2}|\boldsymbol{\Sigma}^q|^{1/2}}~~~~~~~~~\\
\exp\biggr( -\frac{1}{2}({\tilde{\bf y}}^j - {\boldsymbol{\mu}}^q)^T
(\boldsymbol{\Sigma}^q)^{-1}({\tilde{\bf y}}^j - {\boldsymbol{\mu}}^q) \biggr)
\biggr\}\biggr]+\\
 \sum_{j=1}^M\sum_{q=1}^{\kappa}{{b^{j,q}E_{{\bf C}}\ln(C^{q})}}
+Const.\\
= \biggr[\sum_{j=1}^M\sum_{q=1}^{\kappa}
{b^{j,q}}E_{ \boldsymbol{\mu}, \boldsymbol{\Sigma}}
\biggr[-\frac{D}{2}\ln{(2\pi)}-\frac{1}{2}\ln|\boldsymbol{\Sigma}^q|~~~~~\\
-\frac{1}{2}({\tilde{\bf y}}^j - {\boldsymbol{\mu}}^q)^T
(\boldsymbol{\Sigma}^q)^{-1}({\tilde{\bf y}}^j - {\boldsymbol{\mu}}^q)
\biggr]+\\
{b^{j,q}E_{{\bf C}}\ln(C^{q})}\biggr]
+Const.\\
= \sum_{j=1}^M\sum_{q=1}^{\kappa}
{b^{j,q}}\biggr[
-\frac{D}{2}\ln{(2\pi)}-\frac{1}{2}E_{\boldsymbol{\Sigma}}[\ln|\boldsymbol{\Sigma}^q|]~~~~~~~\\
-\frac{1}{2}E_{ \boldsymbol{\mu}, \boldsymbol{\Sigma}}[({\tilde{\bf y}}^j - {\boldsymbol{\mu}}^q)^T(\boldsymbol{\Sigma}^q)^{-1}({\tilde{\bf y}}^j - {\boldsymbol{\mu}}^q)]+\\
E_{{\bf C}}\ln(C^{q})\biggr] +Const.
\end{split}
\end{equation}\\
Considering that
\begin{equation}\label{aa11}
\begin{split}
\ln {{\rho}}^{j,q} = -\frac{D}{2}\ln{(2\pi)}-\frac{1}{2}E_{\boldsymbol{\Sigma}}[\ln|\boldsymbol{\Sigma}^q|]~~~~~~~~~~~~~~~~~~~~~~~~~~~\\
-\frac{1}{2}E_{ \boldsymbol{\mu}, \boldsymbol{\Sigma}}[({\tilde{\bf
y}}^j -
{\boldsymbol{\mu}}^q)^T(\boldsymbol{\Sigma}^q)^{-1}({\tilde{\bf
y}}^j - {\boldsymbol{\mu}}^q)]+ E_{{\bf C}}\ln(C^{q})
\end{split}
\end{equation}\\
we get
\begin{equation}\label{aa12}
\begin{split}
\ln \mathcal{Q}^*({\bf B}) =
\sum_{j=1}^M\sum_{q=1}^{\kappa}
{b^{j,q}}\ln {{\rho}}^{j,q}
+Const.\\
=\prod_{j=1}^M\prod_{q=1}^{\kappa}
({{\rho}}^{j,q})^{b^{j,q}} +Const.
\end{split}
\end{equation}\\
where according to \cite{bishop}
\begin{equation}\label{aa11-b}
\begin{split}
E_{\boldsymbol{\Sigma}}[\ln|\boldsymbol{\Sigma}^q|] = \sum_{\tilde{i}}^{D}{\tilde{\psi}\biggr(\frac{\upsilon+1-\tilde{i}}{2}}\biggr)+
D\ln2+ln|{\bf W}_q| \\
E_{ \boldsymbol{\mu}, \boldsymbol{\Sigma}}[({\tilde{\bf y}}^j
- {\boldsymbol{\mu}}^q)^T(\boldsymbol{\Sigma}^q)^{-1}({\tilde{\bf y}}^j -
{\boldsymbol{\mu}}^q)] = D\beta_q^{-1}+
\upsilon_q({\tilde{\bf y}}^j
- {\bf m}_q)^T\\
{\bf W}_q
({\tilde{\bf y}}^j
- {\bf m}_q)~~~~~~~~~~~~~\\
E_{{\bf C}}\ln(C^{q}) =\tilde{\psi}(\alpha^q)-\tilde{\psi}(\tilde{\alpha})~~~~~~~~~~~~~~~~~~~~~~~~~~~~~~~
\end{split}
\end{equation}
where $\tilde{\alpha} = \sum_{q=1}^{\kappa}\alpha^q$ and
$\tilde{\psi}(\cdot)$ is the digamma function. We can write
\begin{equation}\label{aa14}
 \mathcal{Q}^*({\bf B}) \propto
\prod_{j=1}^M\prod_{q=1}^{\kappa}
({{\rho}}^{j,q})^{b^{j,q}}
\end{equation}
Requiring that this distribution be normalized,
\begin{equation}\label{aa15}
{r}^{j,q} =
\frac{{\boldsymbol{\rho}}^{j,q}}{\sum_{q=1}^\kappa{\boldsymbol{\rho}}^{j,q}
},
\end{equation}\\
\subsection{Derivation of equation (\ref{eq25})}
Using again equation (\ref{eq18})
we can write
\begin{equation}\label{aa18}
\ln \mathcal{Q}^*({\bf C}, {\boldsymbol{\mu}}, \boldsymbol{\Sigma})
= E_{{\bf B}} [\ln p({\tilde{\bf y}}, {\bf B}, {\bf
C},\boldsymbol{\mu}, \boldsymbol{\Sigma})]+Const.
\end{equation}\
According to the decomposition of equation (\ref{eq19}) we have
\begin{equation}\label{aa19}
\begin{split}
\ln \mathcal{Q}^*({\bf C}, {\boldsymbol{\mu}}, \boldsymbol{\Sigma}) =
E_{{\bf B}}
[\ln p({\tilde{\bf y}}|{\bf B}, {\boldsymbol{\mu}},{\bf \Sigma})+ \ln p({\bf B}|{\bf C})\\
+ \ln p({\bf C})+\ln p({\boldsymbol{\mu}}|{\bf \Sigma})+\ln p({\bf
\Sigma})] +Const.
\end{split}
\end{equation}
\begin{equation}\label{aa20}
\begin{split}
\ln \mathcal{Q}^*({\bf C}, {\boldsymbol{\mu}}, \boldsymbol{\Sigma}) =
\ln p({\bf C})+ E_{{\bf B}}[\ln p({\bf B}|{\bf C})]+ \ln p({\boldsymbol{\mu}}|{\bf \Sigma})+ \\
\ln p({\bf \Sigma})+E_{{\bf B}}[\ln p({\tilde{\bf y}}|{\bf B},
{\boldsymbol{\mu}},{\bf \Sigma})] +Const.
\end{split}
\end{equation}
which can also be written as
\begin{equation}\label{aa21}
\begin{split}
\ln \mathcal{Q}^*({\bf C}, {\boldsymbol{\mu}}, \boldsymbol{\Sigma}) =
\ln p({\bf C})+ E_{{\bf B}}[\ln p({\bf B}|{\bf C})]+ \ln p({\boldsymbol{\mu}},{\bf \Sigma})  \\
+E_{{\bf B}}[\ln p({\tilde{\bf y}}|{\bf B}, {\boldsymbol{\mu}},{\bf
\Sigma})] +Const.
\end{split}
\end{equation}
By using equation (\ref{eq12}), we can write the above equation as
\begin{equation}\label{aa22}
\begin{split}
\ln \mathcal{Q}^*({\bf C}, {\boldsymbol{\mu}}, \boldsymbol{\Sigma}) =
\ln p({\bf C})+ E_{{\bf B}}[\ln p({\bf B}|{\bf C})]+ \ln p({\boldsymbol{\mu}},{\bf \Sigma})   \\
+E_{{\bf B}}\biggr[\ln
\prod_{j=1}^M\prod_{q=1}^{\kappa}\mathcal{N}{({\tilde{\bf
y}}^j|{\boldsymbol{\mu}}^q,{\bf \Sigma}^q)}^{b^{j,q}}\biggr] +Const.
\end{split}
\end{equation}\\
Also
\begin{equation}\label{aa23}
\begin{split}
\ln \mathcal{Q}^*({\bf C}, {\boldsymbol{\mu}}, \boldsymbol{\Sigma})
= \ln p({\bf C})+ E_{{\bf B}}[\ln p({\bf B}|{\bf C})]+
\sum_{q=1}^{\kappa} \ln p({\boldsymbol{\mu}}^q,{\bf \Sigma}^q)+ \\
\sum_{j=1}^M\sum_{q=1}^{\kappa}E_{{\bf
B}}[{b^{j,q}}\ln\mathcal{N}{({\tilde{\bf y}}^j|{\boldsymbol{\mu}}^q,{\bf
\Sigma}^q)}] +Const.
\end{split}
\end{equation}
We know that the right-hand side of this expression decomposes into
a sum of terms involving only ${\bf C}$ and the terms only involving
$\boldsymbol\mu$ and ${\boldsymbol\Sigma}$ which implies that
$p({\bf C}, \boldsymbol \mu, \boldsymbol\Sigma)$ factorizes to give
$p({\bf C})p(\boldsymbol\mu,\boldsymbol\Sigma)$.
\begin{equation}\label{aa24}
p({\boldsymbol{\mu}},{\bf \Sigma}, {\bf C}) = p( {\bf C})p({\boldsymbol{\mu}},{\bf \Sigma})
\end{equation}\\
By keeping only those terms of (\ref{aa23}) which depend on ${\bf
C}$ we get
\begin{equation}\label{aa25}
\ln \mathcal{Q}^*({\bf C}) =
\ln p({\bf C})+ E_{{\bf B}}[\ln p({\bf B}|{\bf C})] +Const.
\end{equation}
From equation (\ref{eq20}), we know
\begin{equation}\label{aa26}
p({\bf C}) = Dir({\bf C}|\alpha_\circ) = \prod_{q=1}^{\kappa} ({\bf
C}^q)^{(\alpha_{\circ}-1)}
\end{equation}
and therefore
\begin{equation}\label{aa27}
\ln p({\bf C})= {(\alpha_{\circ}-1)}\sum_{q=1}^{\kappa} \ln {\bf
C}^q.
\end{equation}
According to equation (\ref{eq22})
\begin{equation}\label{aa28}
p({\bf B}|{\bf C}) = \prod_{j=1}^M\prod_{q=1}^{\kappa}{({\bf
C}^q)}^{b^{j,q}}.
\end{equation}
Using equation (\ref{aa27}) and (\ref{aa28}) in (\ref{aa25}) we get
\begin{equation}\label{aa29}
\begin{split}
\ln \mathcal{Q}^*({\bf C})= {(\alpha_{\circ}-1)}\sum_{q=1}^{\kappa} \ln {\bf C}^q +
E_{{\bf B}}\biggr[\sum_{j=1}^M\sum_{q=1}^{\kappa}{{b^{j,q}}\ln({\bf C}^q)}\biggr] \\
+ Const.
\end{split}
\end{equation}
\begin{equation}\label{aa30}
\begin{split}
\ln \mathcal{Q}^*({\bf C})\!=\!\!
\!{(\alpha_{\circ}-1)}\sum_{q=1}^{\kappa} \ln {\bf C}^q \! + \!
\sum_{j=1}^M\sum_{q=1}^{\kappa}{E_{\bf B}[{b^{j,q}}\ln({\bf
C}^q)}]\!\! + \!Const.
\end{split}
\end{equation}
For the discrete distribution $\mathcal{Q}^*({\bf B})$ we have the
standard result (a similar representation is shown in equation
(10.50) of \cite{bishop}), and $E_{\bf B}[b^{j,q}] = r^{j,q}$.
Therefore
\begin{equation}\label{aa32}
\ln \mathcal{Q}^*({\bf C})= {(\alpha_{\circ}-1)}\sum_{q=1}^{\kappa}
\ln {\bf C}^q + \sum_{j=1}^M\sum_{q=1}^{\kappa}{r^{j,q}}\ln({\bf
C}^q) + Const.
\end{equation}
by taking the exponential on both sides
\begin{equation}\label{aa33}
 \mathcal{Q}^*({\bf C})= \prod_{q=1}^{\kappa} ({\bf C}^q)^{(\alpha_{\circ}-1)}
\prod_{j=1}^M\prod_{q=1}^{\kappa}({\bf C}^q)^{r^{j,q}} + Const.
\end{equation}
\begin{equation}\label{aa34}
 \mathcal{Q}^*({\bf C})= \prod_{q=1}^{\kappa} ({\bf C}^q)^{(\alpha_{\circ}-1)}
\prod_{q=1}^{\kappa}({\bf C}^q)^{\sum_{j=1}^M{r^{j,q}}} + Const.
\end{equation}
\begin{equation}\label{aa35}
 \mathcal{Q}^*({\bf C})= \prod_{q=1}^{\kappa} ({\bf C}^q)^{(\alpha_{\circ}-1}
({\bf C}^q)^{\sum_{j=1}^M{r^{j,q}}} + Const.
\end{equation}
Suppose ${\sum_{j=1}^M{r^{j,q}}} = N^q$, then
 \begin{equation}\label{aa37}
 \mathcal{Q}^*({\bf C})= \prod_{q=1}^{\kappa} ({\bf C}^q)^{(\alpha_{\circ}-1)}
({\bf C}^q)^{N^q} + Const.
\end{equation}
 \begin{equation}\label{aa38}
 \mathcal{Q}^*({\bf C})= \prod_{q=1}^{\kappa} ({\bf C}^q)^{(\alpha_{\circ}-1)+{N^q}}
 + Const.
\end{equation}
by comparing it with equation (\ref{aa26}), $ \mathcal{Q}^*({\bf C})
= Dir({\bf C}|\alpha),$
where $\alpha = {(\alpha_{\circ}-1)+{N^q}}$.
\subsection{Derivation of equation (\ref{eq28})}
Consider again the factorization of $p({\bf C}, \boldsymbol\mu, \boldsymbol\Sigma)$ to give $p({\bf C}), p(\boldsymbol\mu,\boldsymbol\Sigma)$. By keeping only those terms of equation (\ref{aa23}) which depend on ${\boldsymbol \mu}$ and ${\boldsymbol \Sigma}$ we get
\begin{equation}\label{aa41}
\begin{split}
\ln \mathcal{Q}^*({\boldsymbol{\mu}}, \boldsymbol{\Sigma}) =
 \sum_{q=1}^{\kappa} \ln p({\boldsymbol{\mu}}^q,{\bf \Sigma}^q)+ ~~~~~~~~~~~~~~~~~~~~~~~~~~ \\
 \sum_{j=1}^M\sum_{q=1}^{\kappa}E_{{\bf B}}[{b^{j,q}}\ln\mathcal{N}{({\tilde{\bf y}}^j|{\boldsymbol{\mu}}^q,{\bf \Sigma}^q)}]
+Const.
\end{split}
\end{equation}\\
According to equation (\ref{eq21})
\begin{equation}\label{aa42}
p({\boldsymbol{\mu}},{\bf \Sigma}) = \prod_{q=1}^{\kappa}\mathcal{N}({\boldsymbol{\mu}}^q|{\bf m}_\circ,\beta_\circ^{-1}{\bf \Sigma}^q)\mathcal{W}({\bf \Sigma}^q|{\bf W}_\circ,\upsilon_\circ)
\end{equation}
therefore
\begin{equation}\label{aa43}
\begin{split}
\ln \mathcal{Q}^*({\boldsymbol{\mu}}, \boldsymbol{\Sigma}) =
 \sum_{q=1}^{\kappa} \ln \mathcal{N}({\boldsymbol{\mu}}^q|{\bf m}_\circ,\beta_\circ^{-1}{\bf \Sigma}^q) +
\ln \mathcal{W}({\bf \Sigma}^q|{\bf W}_\circ,\upsilon_\circ)\\
+ \sum_{j=1}^M\sum_{q=1}^{\kappa}E_{{\bf
B}}[{b^{j,q}}\ln\mathcal{N}{({\tilde{\bf y}}^j|{\boldsymbol{\mu}}^q,{\bf
\Sigma}^q)}] +Const.
\end{split}
\end{equation}\\
whereas the Wishart distribution can be defined as
\begin{equation}\label{aa44}
\mathcal{W}({\bf \Sigma}^q|{\bf W}_\circ,\upsilon_\circ) = \frac{({\bf \Sigma}^q)^{(\upsilon_\circ - D - 1 )}  \\
\exp (\frac{-1}{2}T_r({\bf W}_\circ ({\bf
\Sigma}^q)^{-1}))}{2^{\upsilon_\circ D / 2}|{\bf
W}_\circ|^{\upsilon_\circ / 2} T_D(\upsilon_\circ/2)}
\end{equation}
also
\begin{equation}\label{aa45}
\begin{split}
\mathcal{N}({\boldsymbol{\mu}}^q|{\bf m}_\circ,\beta_\circ^{-1}{\bf \Sigma}^q) =
\biggr[\ln\biggr\{\frac{1}{(2\pi)^{D/2}(\beta_\circ^{-1}|\boldsymbol{\Sigma}^q|)^{1/2}} \\
\exp\biggr( -\frac{1}{2}({\boldsymbol{\mu}}^q - {\bf m}_\circ)^T
\beta_\circ\boldsymbol{\Sigma}_q^{-1}({\boldsymbol{\mu}}^q - {\bf
m}_\circ) \biggr) \biggr\}\biggr]
\end{split}
\end{equation}
Using these equations in (\ref{aa43}), keeping only the terms which
contain $\boldsymbol \mu$ and $\boldsymbol \Sigma$ we get the final
expression for (\ref{eq28})
where  $\beta_q, {\bf m}_q, {\bf W}_q$ and $\upsilon_q$ are defined
in (\ref{eq29})-(\ref{eq32}). 
\begin{biography}[{\includegraphics[width=1.5in,height=1.2in,clip,keepaspectratio]{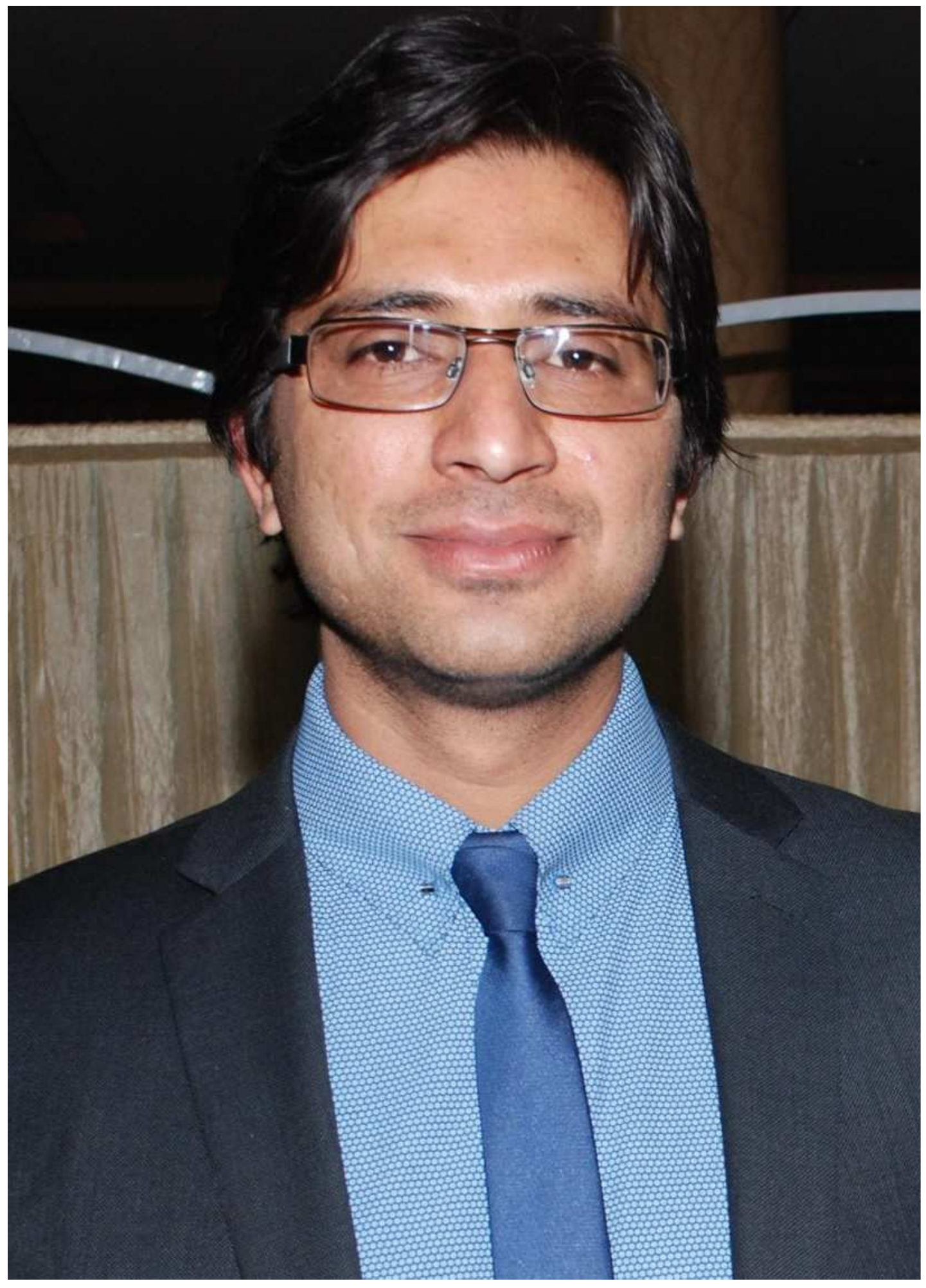}}]{Ata-ur-Rehman}
received the B.Eng. degree
in electronic engineering from Air University,
Islamabad, Pakistan in 2006. He then joined the
University of Engineering and Technology, Lahore,
Pakistan, as a Lecturer/Lab Engineer in 2007. He moved to
Loughborough University, U.K., in 2009 where he
received his M.Sc. degree with distinction in Digital
Communication Systems in 2010. He received his Ph.D
degree in Signal Processing from Loughborough University, UK in 2014.
Since December 2013 he is a PDRA at the Department of ACSE, The University of Sheffield.
His main area of research is multi-target
tracking.
\end{biography}
\begin{biography}[{\includegraphics[width=1.1in,height=1.35in,clip,keepaspectratio]{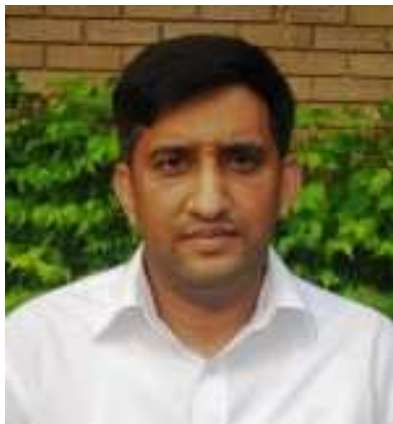}}]{Mohsen Naqvi} (S'06-M'09-SM'14) received  Ph.D. degree in Signal Processing from Loughborough University, UK in 2009. Where he was a PDRA on the EPSRC UK funded projects and REF Lecturer from July 2009 to July 2015.

Dr Naqvi is a Lecturer  in Signal and Information Processing at the School of Electrical and Electronic Engineering, Newcastle University, UK and Fellow of the Higher Education Academy. His research interests include multimodal processing for human behavior analysis, multi-target tracking and source separation, all for machine learning.
\end{biography}
\begin{biography}[{\includegraphics[width=1in,height=1.25in,clip,keepaspectratio]{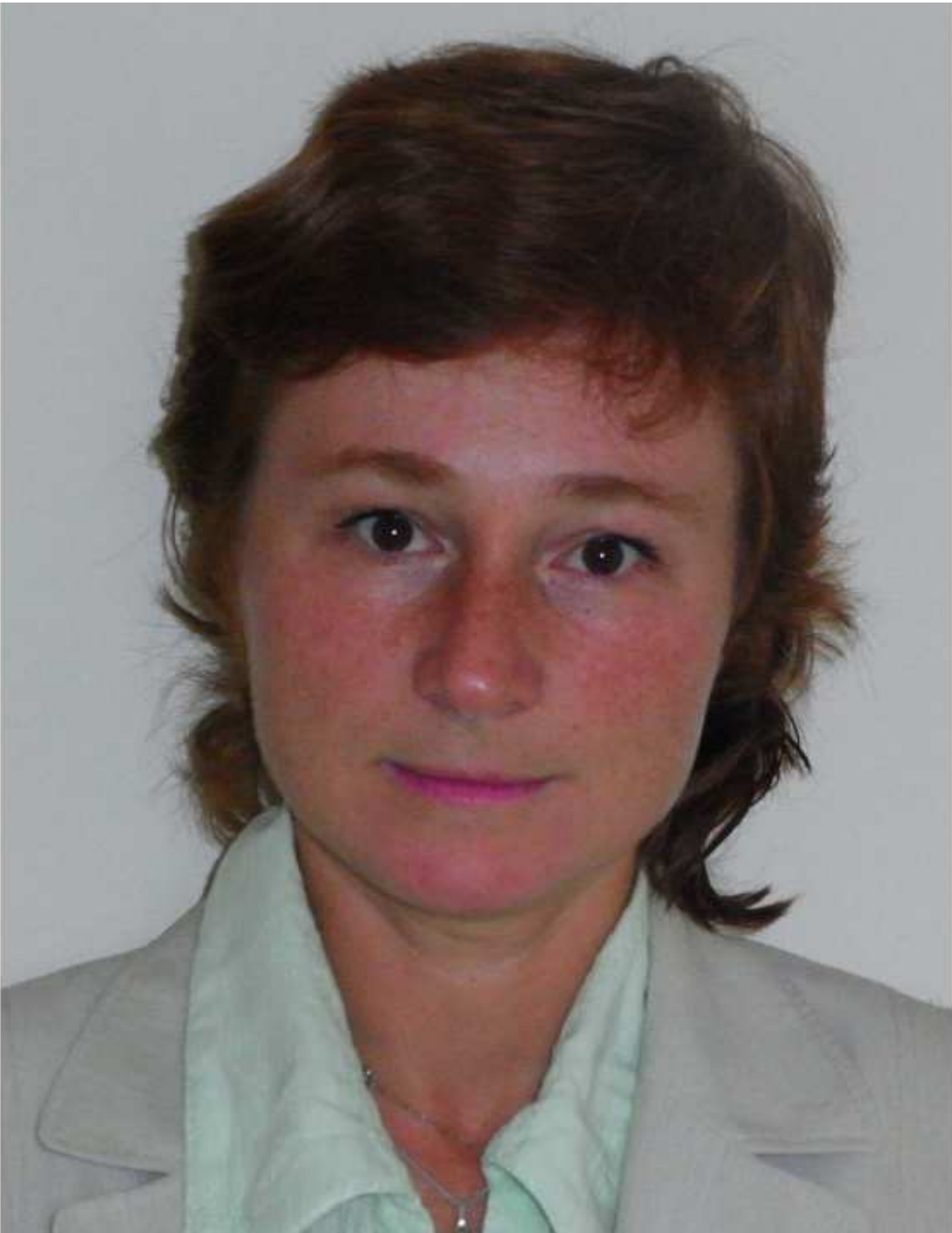}}]{Lyudmila~Mihaylova} (M'98, SM'2008) is Associate Professor (Reader in Advanced Signal Processing and Control) at the Department of Automatic Control and Systems Engineering at the University of Sheffield, United Kingdom. Her research is in the areas of machine learning and autonomous systems with various applications  such as navigation, surveillance and sensor network systems. Dr Mihaylova is an Associate Editor of the IEEE Transactions on Aerospace and Electronic Systems and of the Elsevier Signal Processing Journal.
\end{biography}
\begin{biography}[{\includegraphics[width=1in,height=1.25in,clip,keepaspectratio]{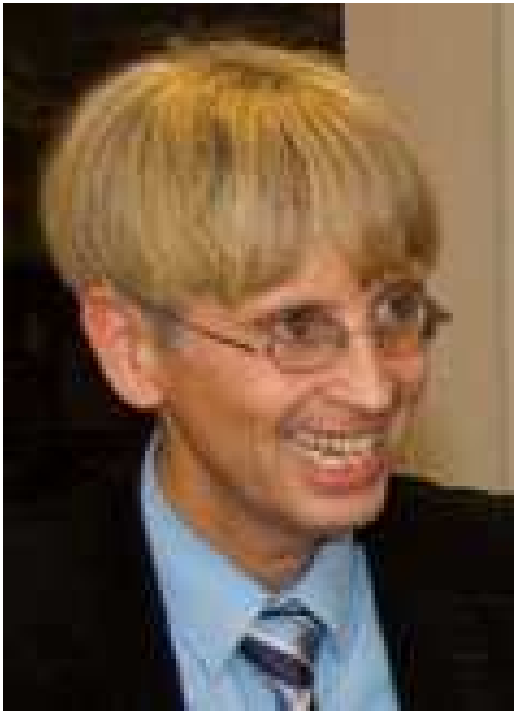}}]{Jonathon A. Chambers}(S'83-M'90-SM'98-F'11)
received the Ph.D. and D.Sc. degrees in signal processing from the Imperial College of Science, Technology and Medicine (Imperial College London), London, U.K., in 1990 and 2014, respectively.  He is a Professor of signal and information processing within the School of Electrical and Electronic Engineering, Newcastle University, U.K., and a Fellow of the Royal Academy of Engineering, U.K.. He has also served as an Associate Editor for the IEEE TRANSACTIONS ON SIGNAL PROCESSING over the periods 1997-1999, 2004-2007, and as a Senior Area Editor 2011-15.
\end{biography}

\end{document}